\documentclass{article}
\usepackage[preprint]{colm2025_conference} 
\usepackage{xcolor}
\usepackage{hyperref}
\usepackage[margin=1.25in]{geometry}
\usepackage{graphicx}
\usepackage{amsmath}
\usepackage{amsfonts}
\usepackage{amssymb}
\usepackage{booktabs}
\usepackage{adjustbox}
\usepackage{cellspace}
\usepackage{wrapfig}
\usepackage{amsthm}
\usepackage[ruled,linesnumbered,noresetcount]{algorithm2e}
\usepackage{mathtools}
\usepackage{csquotes}
\usepackage{array}
\usepackage{url,eucal}
\usepackage{tabularx}
\usepackage{makecell}

\theoremstyle{definition}
\newtheorem{definition}{Definition}[section]

\title{A Definition and Roadmap for World Models}

\author{Physical Intelligence Team, Shanghai AI Laboratory}

\begin{document}
\maketitle

\begin{abstract}
World models---internal simulators that learn the structure and dynamics of an environment---have become one of the most actively debated concepts in AI.
From model-based reinforcement learning and video generation to embodied robotics and ultimately, physical AI,
researchers across AI subfields are building systems that they call ``world models'',
yet there is no consensus on what a world model fundamentally is, what it should predict, or how it should be built.
This perspective article provides a scientific definition of world models, discussions of their key technical aspects, and a staged roadmap for developing effective world models.
\end{abstract}

\tableofcontents

\section{Introduction}

The term ``world model'' has accumulated a diverse range of contextual usages
that conceals as
much as it reveals.
As early as in 1943, 
\cite{craik1943nature}
proposed that biological organisms survive and thrive by holding
``working models''
of the physical world within their minds.
Rather than relying exclusively on slow, trial-and-error physical interactions,
these models enable the mind to simulate hypothetical scenarios, predict the outcomes of candidate actions,
and precompute optimal strategies---this algorithmic instantiation represents the formal realization of what we today term the ``world action model".
Craik's hypothesis underpins some important AI architectures such as reinforcement learning agents and generative models
\citep{sutton2018rl}.
The concept has
been invoked across computer vision, robotics, and generative modeling,
often to describe very different systems. The most influential recent
attempt at disambiguation is Fei-Fei's functional taxonomy, which
partitions world models into three categories: \textit{renderers} (which
generate pixels), \textit{simulators} (which predict state transitions), and
\textit{planners} (which propose
actions)~\citep{worldlabs2026taxonomy}.

This function taxonomy performs a useful clarification, but its fundamental
limitation is that it classifies \textit{decoding} (a.k.a.\ projections) of a representation,
not the representation itself. A single unified 
internal model can be
decoded into RGB pixels, 
state vectors, or candidate 
action proposals
in accordance with the query interface invoked. 
The taxonomy specifies what a world model outputs but
does not define what the internaml model is or how it should be constructed. The
upstream question---how to build the shared internal representation that
supports all three decodings---still remains a question for functional taxonomy.

A competing view, most prominently articulated by
\cite{lecun2022path}, holds that generative reconstruction is never the
right objective: joint-embedding architectures that predict in latent
space are sufficient, and pixel-level reconstruction wastes
representational capacity on photometric irrelevancies. 
While we agree that precise pixel-level reconstruction should not be the ultimate goal of a world model, reconstruction quality is a good way to trace the information a compressed representation fails to preserve. Ultimately, the only way to verify how important reconstruction is for world modeling is through testing.

Related to compressed presentation, it is now broadly accepted that large language models (LLMs) excel at compressing human {\it  linguistic}  knowledge. 
Similarly, a world model instantiates a unified compression mechanism whose core objective is to maximally compress the joint distribution of {\it real-world physical sensory observations and agent actions.
}
Lossy compression creates an unavoidable representational tradeoff across a continuous spectrum: latent representations may lean toward compact high-level semantic abstraction, or prioritize fine-grained physical dynamics and perceptual fidelity, with all practical world models occupying an intermediate balance between these two extremes.
This compression-centered
view may better capture the architectural decisions, data constraints, and
empirical bottlenecks that characterize current work. 
\paragraph{Data Determines the Ceiling} 
We begin from a premise that is both simple and consequential: the ceiling of
any intelligent system, in terms of its capacity to generalize in the
physical world, is set by the diversity of physical experience represented in
its training data. At a fixed architecture and compute budget, it is data
diversity that determines this ceiling, not model structure or training
hyperparameters. Architecture and compute affect how efficiently the ceiling
is approached, but they cannot raise
it~\citep{hoffmann2022training}.

The only source of physical data that currently scales to the required
diversity is the open internet: images, videos, and text at hundreds of
billions of examples. Proprietary sources---dexterous manipulation datasets,
first-person embodied recordings, and exoskeleton-mounted sensors---are
structurally limited in both diversity and scale. No hardware deployment, in
the foreseeable future, will produce internet-scale quantities of
manipulation data. To achieve something like a ``world model'' that
generalizes to the physical world, we must therefore extract maximal physical
understanding from the data that we already have.

\begin{figure}[h]
    \centering
    \includegraphics[width=0.85\linewidth]{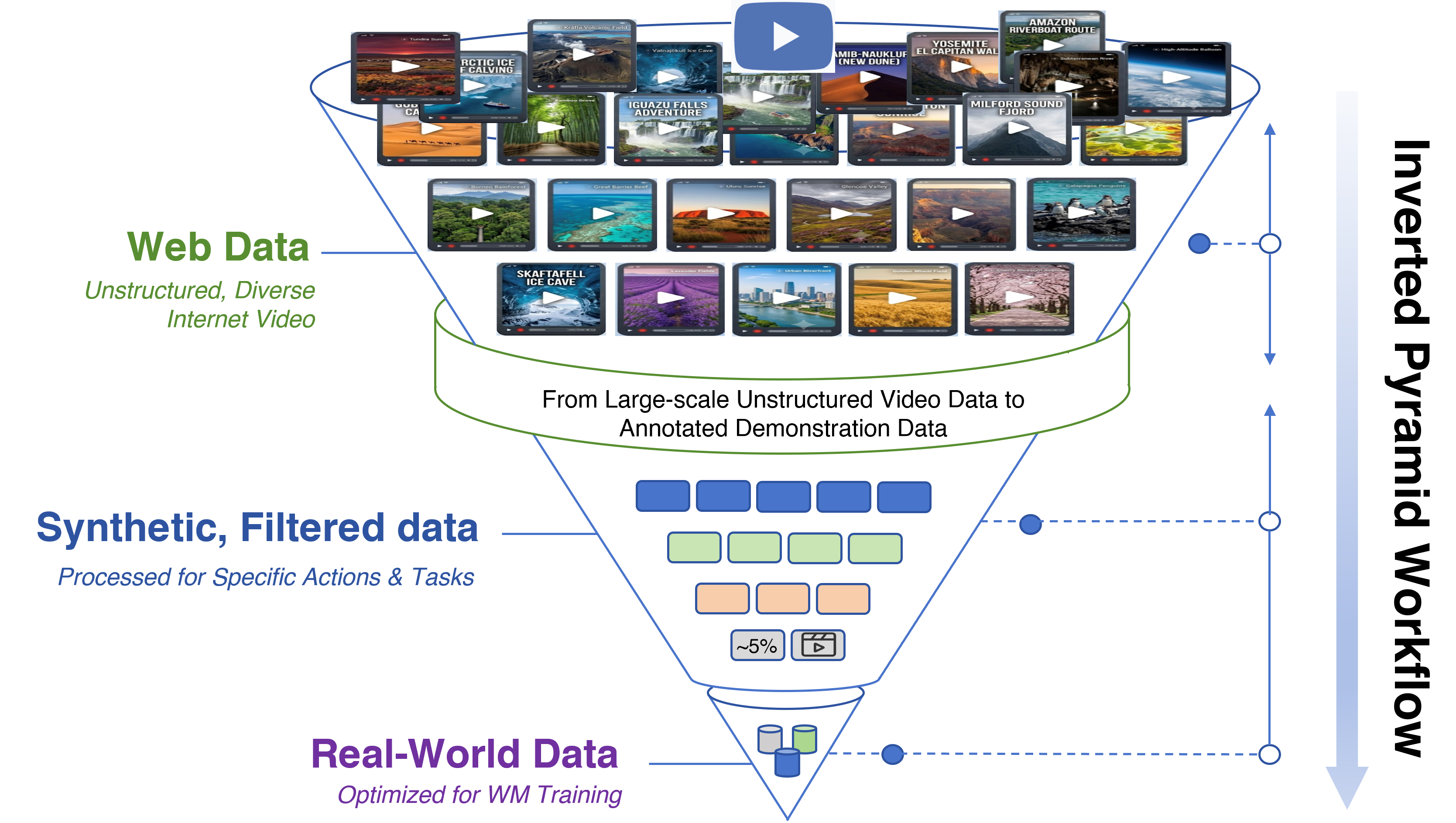}
    \caption{This diagram illustrates a data pyramid inversion funnel pipeline for robot and physical world model training, which shows
    vast unstructured public media down to compact, task-optimized robotic training data. 
    At the widest top tier sits Web Data Sources: unstructured, diverse public video, representing billions of raw pixel streams of human physical interaction. 
    The funnel narrows into the middle layer, Synthetic and Filtered Dataset, a drastically pruned subset 
    refined and synthesized for targeted robotic actions and specific manipulation tasks. 
    The narrowest bottom tip yields Real-World Task Data: a small, highly curated dataset fully optimized for end-to-end robot training, compact enough for efficient embodied model fine-tuning.
}
    \label{fig:data1}
\end{figure}

Internet video, in particular, serves as an unprecedented, large-scale natural corpus that encodes a staggering breadth of structured physical world knowledge entirely implicitly. Embedded within the raw pixel streams of everyday video are the foundational priors that govern our physical reality: object permanence (the invariant persistence of objects outside the visual field), rigidity and non-rigidity constraints that distinguish solid bodies from deformable materials, kinematic limits on object motion and interaction, the temporal dynamics of lighting and shadow, the causal logic of occlusion and disocclusion, the hierarchical causal chaining of events, and the structured priors of human-like action and interaction.
{
\it
The central challenge is that all of this physical structure remains \textit{latent} and inaccessible within raw pixel space.
}
We introduce the \textbf{Inverted Pyramid Workflow} in Figure \ref{fig:data1}, which addresses the core challenge laid out above. While raw internet video contains latent physical priors buried in unlabeled pixels, proprietary robotic hardware cannot generate data of comparable scale and diversity on its own. By starting with the internet's unmatched breadth of real-world physical footage as the foundation, the pipeline first unlocks that latent, implicit physical knowledge through automated filtering and annotation, raising the generalization ceiling of the world model via unmatched data diversity. The successive funneling acts as a precision distillation step: it strips irrelevant visual content, synthesizes standardized action representations, and retains only the physically meaningful motion and interaction signals that robots need. The final, compact, task-aligned real-world dataset then lets models efficiently converge on embodied manipulation tasks without sacrificing the broad physical priors extracted from billions of web videos. Architecture and compute only govern how quickly the model learns, but this inversion pipeline expands the foundational physical experience diversity of the training corpus---lifting the hard upper limit on how well the world model can generalize to unseen real-world objects, motions, and causal scenarios that limited, small-scale proprietary robot datasets cannot capture.

\paragraph{World Model as a Compression Mechanism}
High-dimensional video observations are dominated by task-irrelevant photometric variance: changes in viewpoint, illumination, texture, sensor noise, and background clutter that obscure the underlying physical laws.
Raw pixels carry no explicit representation of permanence, causality, or constraint; they encode only superficial appearance, not the generative physical structure that produces the visual signal.
Extracting these implicit physical priors therefore requires a purpose-built {\it compression mechanism}: one capable of distilling high-dimensional, noisy video observations into a compact, powerful, semantically structured representation. The core function of this compression is not mere dimensionality reduction, but targeted information preservation: it must retain precisely the structured causal and physical information required for downstream physical reasoning and control, while systematically discarding the irrelevant photometric nuisance variance that dominates raw pixel data.
{\it This, we argue, is the defining, foundational task of the world model—and at its core, this is strictly a compression problem, not a generation or simulation problem. }
Generation and simulation are downstream capabilities that emerge from a good compressed representation, but they are not the objective itself; the core goal of a world model is to solve the information-theoretic problem of distilling implicit physical knowledge from high-dimensional sensory data into a usable, compact form.

 \section{What Is a World Model?}

\subsection{Definition and Major Properties}

\begin{figure}
\centering
\includegraphics[width=\linewidth]{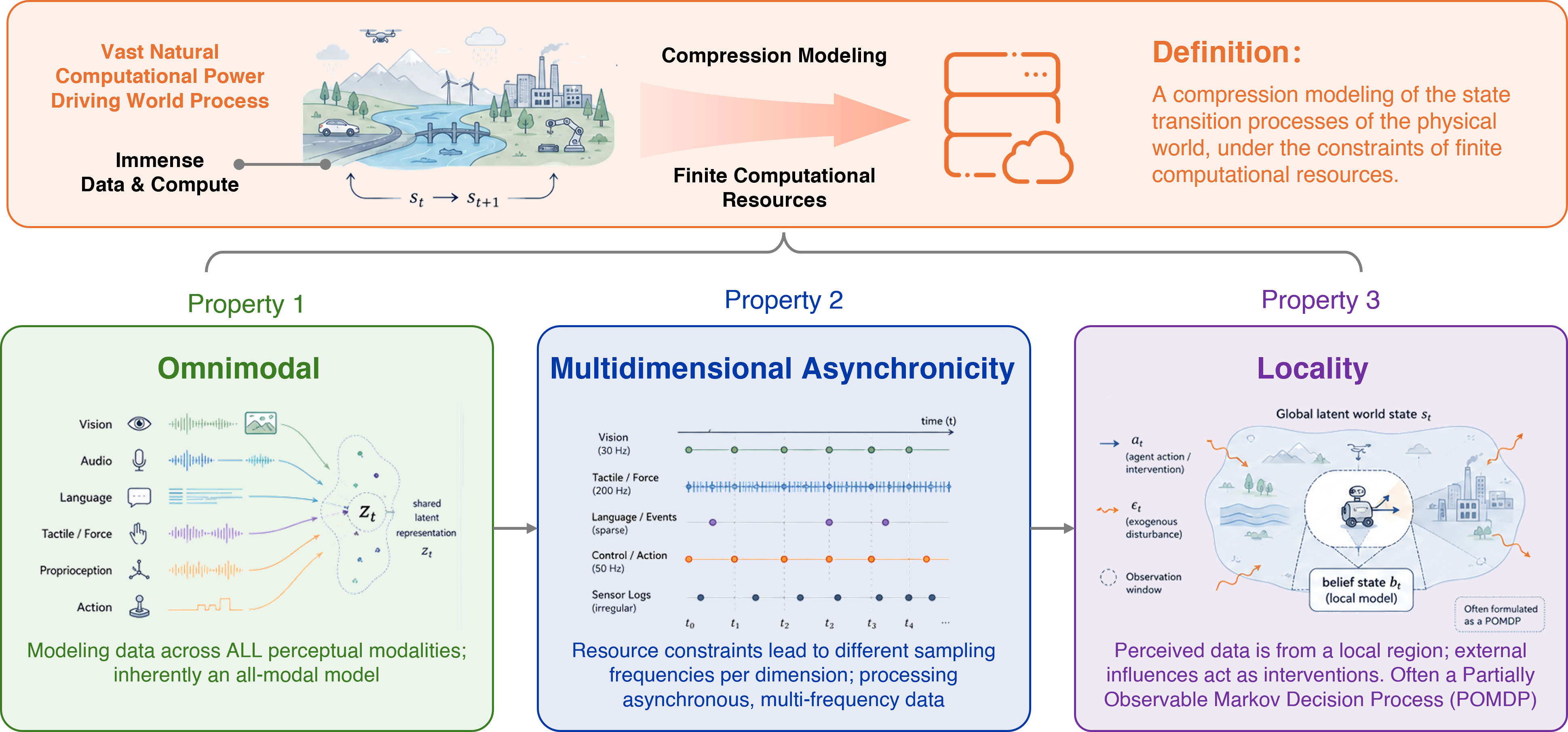}
\caption{An illustration of the nature of the world model and its major properties.}\label{fig:def-and-propoerties}
\end{figure}

In this section, we explicitly define the concept of a world model, specifically from the perspective of modeling the physical world—and outline its fundamental properties.

While the state transition processes of all matter and waves in the universe exist objectively, the natural computational power driving these physical processes is unfathomably immense. Because human-engineered systems cannot replicate this infinite complexity, we establish the following formal definition.

\begin{definition}[World Model]
    \label{def:worldmodel}
    \textit{A world model is a compression modeling of the state transition processes of the physical world, constructed under the constraints of finite computational resources.}\footnote{Note that, based on the definition, the term ``world model'' used in this article carries the same meaning with the term ``physical world model'' and they are used interchangably, but this does mean that ``world model'' is the same as ``physical world model'' in other literature, such as interactive/3D geometry consistent video generation models.}
\end{definition}

As illustrated by Figure~\ref{fig:def-and-propoerties}, operating under these inherent computational constraints, a universal physical world model inherently exhibits three major properties:
\begin{itemize}
    \item \textbf{Omnimodal workscope.} A world model must possess the capacity to model data across all perceptual modalities. It is not limited to text or vision but is fundamentally an all-modal foundational model capable of unified latent representation.
    \item \textbf{Multidimensional Asynchronicity.} Due to finite computational and sensing resources, the data perceived across the physical world's many dimensions are sampled at different frequencies. Consequently, a world model must be capable of processing multi-dimensional, asynchronous (multi-frequency) sequence data.
    \item \textbf{Locality.} Because an agent's perception is bounded by its resources, the data it gathers is often restricted to a local region. External regions constantly exert influence on this local area, acting as interventions. Therefore, modeling the world from a localized perspective is often formulated as a Partially Observable Markov Decision Process (POMDP).
\end{itemize}
To understand the necessity of this localized, omnimodal approximation, we must inspect how current models attempt to answer questions about reality. The world models that dominate the AI field today—whether latent-space predictive representations, reinforcement learning models for game agents, or video-prediction models for embodied simulation—are unified by a shared, generative framing: given current observations, what will happen next? A true physical world model redefines this entire paradigm. The fundamental question shifts from ``what will happen?'' to ``what is happening, why does it happen, and what will happen?'' This requires epistemic reasoning\footnote{
Epistemic reasoning refers to the logical process for evaluating states of knowledge, belief, and justified conviction,
rather than focusing on the physical state of the world. 
It analyzes how agents build, update, and appraise their internal representations from partial observations, supporting evidence, and meta-knowledge of what other agents know.
} 
to understand the operational structure of a physical system. It discovers the system's native operational ontology: identifying discrete operating regimes, establishing the boundaries between healthy and anomalous behaviors, and mapping the causal trajectory of how a system arrived at its current state.
The significance of this approach lies in the drastic differences between the physical and digital worlds. Digital data, once created, remains legible indefinitely; a language model trained years ago can still answer questions about quantum mechanics today. In contrast, physical systems are inherently non-stationary and constantly changing. Critically, the most important events, e.g.,  safety-critical failures and rare fault modes, may never appear in a static training corpus. This is why static world models inevitably struggle to solve physical world problems. We require an architecture capable of taking a universal foundational representation and adapting it locally to the unique physics of individual applications, autonomously and in real time. Such a dynamics system should be of \textbf{self-consistency}, i.e., when new modal or channel of data comes in, the world model can still work and evolve without the risk of collapse.
Ultimately, a physical world model is not merely a simulation engine. It represents a fundamentally different form of intelligence: one that deeply understands the physical world, its constant evolution, and its underlying operational rules under the constraints of localized observation and finite computation.

\subsection{The Agent-Environment Loop}
\begin{figure}[t]
    \centering
    \includegraphics[width=0.8\linewidth]{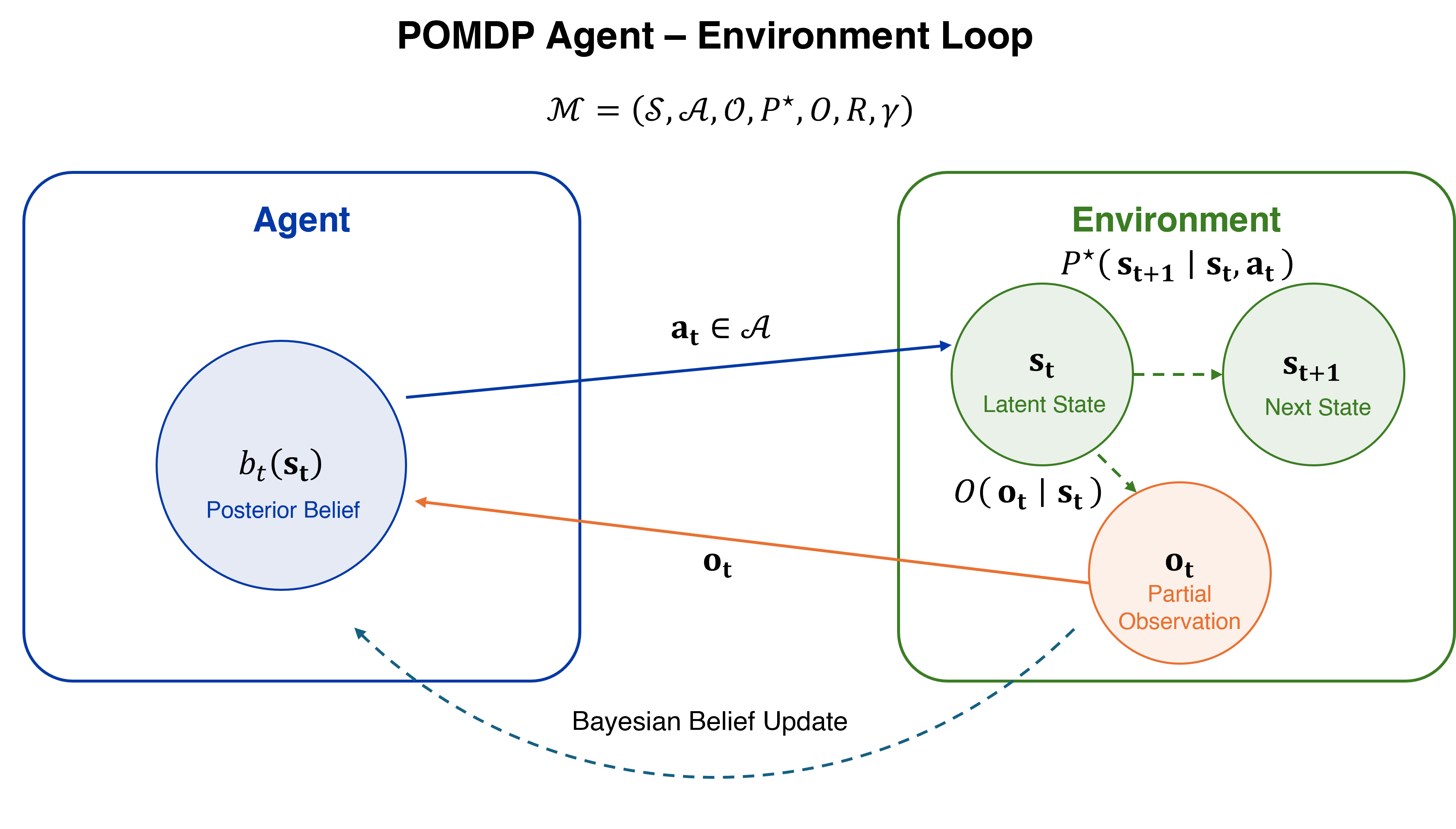}
    \caption{A schematic of the POMDP agent--environment loop using the notation of this subsection. The agent maintains a posterior belief $b_t(\mathbf{s}_t)$, selects an action $\mathbf{a}_t$, the environment evolves under the true transition kernel $P^\star$, produces a partial observation $\mathbf{o}_t$ through the observation model $O$, and the belief is updated by Bayesian filtering.}
    \label{fig:pomdp-agent-environment-loop}
\end{figure}

{
A scientifically useful definition of a world model should begin from the sequential loop between an agent and its environment rather than from any particular model family. A natural formalization is the partially 
observable Markov decision process (POMDP), $\mathcal{M}=(\mathcal{S},\mathcal{A},\mathcal{O}, P^\star,O,R,\gamma)$, in which the environment occupies a latent state $\mathbf{s}_t\in\mathcal{S}$, the agent selects an 
action $\mathbf{a}_t\in\mathcal{A}$, the state evolves according to a transition kernel $P^\star(\mathbf{s}_{t+1}\mid \mathbf{s}_t,\mathbf{a}_t)$, and the agent receives only a partial observation 
$\mathbf{o}_t\in\mathcal{O}$ drawn from an observation model $O(\mathbf{o}_t\mid \mathbf{s}_t)$. Because the true state is generally hidden, competent behavior cannot rely on raw observations alone. It requires an 
internal belief over possible states and a mechanism for updating that belief as new observations arrive. In this sense, a world model is not simply a predictor of pixels, tokens, or trajectories. It is an internal 
model of latent state, observation formation, and action-conditioned dynamics that allows the agent to infer what is currently the case, imagine what could happen next, and evaluate intervention under uncertainty. Figure~\ref{fig:pomdp-agent-environment-loop} summarizes this agent--environment loop and its core variables.

This framing also clarifies the historical lineage of the term. Craik's early account of mental models treated intelligence as the construction of an internal surrogate that can be run ahead of reality before overt 
action is taken \citep{craik1943nature}. Sutton's Dyna architecture translated that intuition into modern reinforcement learning by explicitly coupling real experience, model learning, planning, and acting in a single 
loop \citep{sutton1990dyna,sutton1991dyna}. Recent critiques sharpen the same point: the goal of a world model is not to reproduce the world exhaustively, nor merely to generate visually plausible futures, but to 
simulate the actionable possibilities of the real world for purposeful reasoning and acting \citep{xing2025critiques}. The qualifier actionable is essential. A useful world model must predict not only how observations 
co-vary, but how the environment changes under intervention, which hidden variables matter for control, and which uncertainties remain unresolved. Under this view, the formal objective is to learn a sufficiently 
faithful internal simulator of decision-relevant state, transition, and observation structure so that perception, counterfactual imagination, and policy selection can all be carried out within a common closed loop.
}

\subsection{Understanding vs.\ Predicting: Two Views of the World Model}
{The definition above defines a world model in the physical setting as a finite-resource approximation to the state-transition processes of the physical world and emphasizes three major properties: omnimodality, multidimensional asynchronicity, and locality. Within this definition, recent literature shows that the term ``world model'' is used in two partially overlapping but philosophically distinct senses \citep{ding2024understanding}. In the first, a world model is primarily an instrument for understanding: it compresses sensory data into stable internal representations that expose the entities, relations, and mechanisms needed to interpret the present situation. Prediction remains important, but mainly as a training signal or consistency constraint that pressures the representation to capture the right latent structure. This is the spirit in which early deep world-model work emphasized compact internal state, memory, and latent dynamics as the substrate on which downstream control can be built \citep{ha2018worldmodelsorig}. In the second, a world model is primarily an instrument for prediction: it is judged by its ability to roll the world forward, generate candidate futures, and support foresight for planning and decision-making. LeCun's autonomous-intelligence program places such predictive world models at the center of reasoning and action \citep{dawid2023autonomous}, and large-scale video generators such as Sora make this predictive interpretation especially salient at the level of observable futures \citep{openai2024sora}.

The distinction matters because it changes what we ask a model to represent, optimize, and prove. An understanding-oriented model typically favors latent abstraction, object- or relation-centric structure, belief estimation, and invariance to nuisance detail; its training objectives often emphasize representation learning, state estimation, or compact predictive sufficiency, and its evaluation should ask whether the learned state is informative for control, explanation, and causal intervention. A prediction-oriented model can tolerate weaker interpretability if it supports accurate and controllable forward simulation; therefore, it often favors next-state or next-observation generation, long-horizon rollout stability, action conditioning, and externally judged consistency. The two views also imply different failure modes. A model may infer a compact decision-relevant state and still produce visually poor renderings. In contrast, a video generator may produce strikingly realistic futures yet remain unreliable as a control-relevant simulator if it does not preserve hidden state, causal structure, or intervention semantics \citep{xing2025critiques}. The strongest systems will ultimately need both properties: internal representations that isolate the right latent variables and predictive machinery that propagates them through counterfactual futures. The disagreement is therefore best understood not as a choice between two incompatible definitions but as a statement about which capability is treated as primary and which is treated as derivative in a given architecture.

Our view is that, for physical world models, understanding should be primary and prediction should be in service of it. A model that cannot identify latent operating state, causal structure, and intervention-relevant uncertainty may generate plausible futures yet still fail as a basis for scientific reasoning or embodied control. Conversely, predictive rollout remains indispensable, but chiefly as a mechanism for testing and refining internal understanding through counterfactual simulation. In this sense, we regard the most useful world models not as mere future generators, but as systems that turn partial observation into actionable understanding and then use prediction to support decision-making.
}

\subsection{A Functional Taxonomy: Renderers, Simulators, Planners}

{
The distinction between understanding-oriented and prediction-oriented world models clarifies what capabilities are emphasized, but it does not yet specify how such capabilities are organized within an acting system. A functional taxonomy therefore provides a useful next step: instead of classifying world models by their philosophical emphasis, it decomposes their roles in the agent--environment loop.

Fei-fei's functional taxonomy can be reformulated as three complementary computational roles within an agent–environment system \citep{worldlabs2026taxonomy}. First, a renderer is an observation-generating model: it outputs perceptual consequences—most commonly images or videos—conditioned on a scene description, interaction history, or prospective action sequence. Its function is therefore to approximate a distribution over visually plausible observations rather than to recover the complete underlying state of the environment. Second, a simulator is a dynamics-bearing model that represents how latent or explicitly structured world states evolve under interventions. It must preserve geometry, physical constraints, object persistence, and action-conditioned temporal regularities sufficiently well to support reliable forward prediction. Third, a planner is a decision model that evaluates counterfactual futures and selects actions or action sequences according to a task objective. Rather than merely predicting what may happen, it determines which controllable future should be pursued.

These roles map naturally onto the closed loop of a partially observable Markov decision process (POMDP) \citep{kaelbling1998planning}. Because the true environmental state is not directly accessible, the agent maintains a belief distribution over possible states. The simulator approximates the transition model, propagating candidate states under contemplated actions; the renderer approximates the observation model, projecting those states into the sensory evidence that the agent may subsequently receive; and the planner evaluates the resulting belief-contingent futures to choose an action. After the selected action is executed, the environment produces a new observation, which revises the agent’s belief and initiates the next cycle. The three functions are therefore neither isolated modules nor mutually exclusive model classes. A planner queries the simulator to compare action consequences and may invoke the renderer to anticipate task-relevant perceptual outcomes. Conversely, the planner’s interventions determine which aspects of the environment become observable, thereby generating evidence that can refine both perceptual prediction and dynamics estimation. Their interaction constitutes a recurrent process of inference, imagination, intervention, and correction.

\begin{figure}
    \centering
    \includegraphics[width=0.96\linewidth]{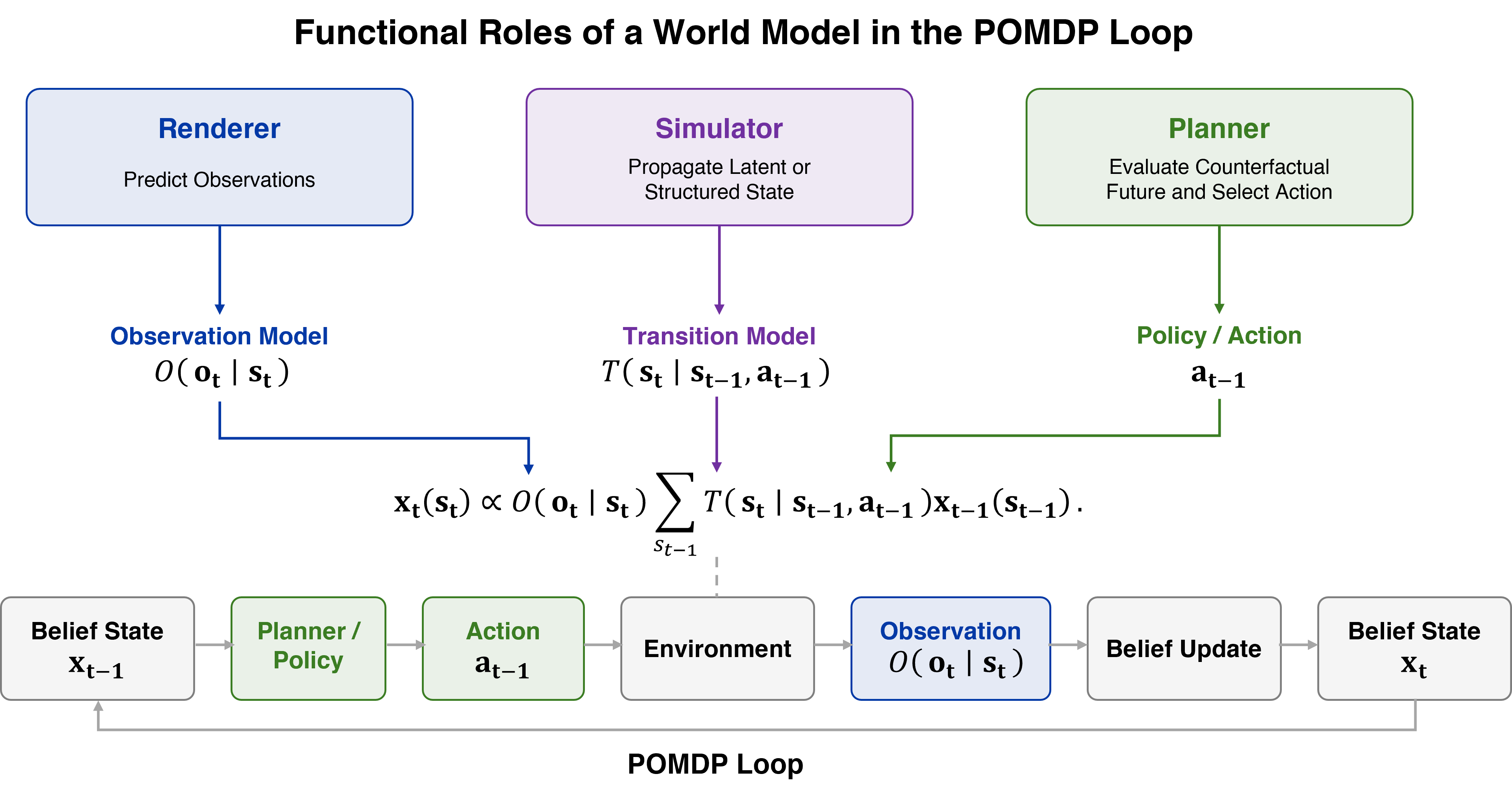}
    \caption{Functional roles of world models in the POMDP loop.}
    \label{fig:functional-roles}
\end{figure}

Bayesian inference provides a normative account of how this loop should manage uncertainty. Before receiving the current observation, the agent forms a predictive prior by propagating its previous posterior through the transition dynamics. The new observation then updates this prior according to

\begin{equation}
\mathbf{x}_t(\mathbf{s_t})\propto O(\mathbf{o}_t\mid\mathbf{s}_t)\sum_{s_{t-1}}T(\mathbf{s_t}\mid \mathbf{s_{t-1}},\mathbf{a_{t-1}})\mathbf{x}_{t-1}(\mathbf{s_{t-1}}).
\end{equation}

Here, $\mathbf{x}_t$ denotes the posterior belief, $T$ the state-transition model, and $O$ the observation likelihood. The resulting posterior constitutes the appropriate informational basis for decision-making because it integrates accumulated knowledge with current evidence while explicitly preserving uncertainty. This posterior subsequently becomes the source of the next predictive prior, making sequential Bayesian filtering the inferential backbone of the POMDP loop. Importantly, rational planning should optimize expected utility over the full belief distribution rather than only over the most likely trajectory. Actions may therefore possess both instrumental value and epistemic value: an agent may act to achieve an external objective, to reduce uncertainty through informative observations, or to balance the two. Priors encode reusable regularities learned across environments, posteriors adapt those regularities to the current situation, and counterfactual simulation connects both to policy selection. A unified world model should thus be understood not merely as a generator of possible futures, but as a Bayesian decision system that continuously converts uncertain predictions into evidence-conditioned, goal-directed behavior.
}

\subsection{A Two-Dimensional Taxonomy: Functions and Architectures}
{Fei-Fei's renderer–simulator–planner taxonomy and our proposed taxonomy operate at different levels of abstraction. Fei-Fei's framework classifies a world model by its functional role in the a\-g\-en\-t–en\-vi\-ron\-ment loop: a renderer produces predicted observations, a simulator propagates world states under physical or dynamical constraints, and a planner selects actions with respect to goals. Our framework instead classifies a model by its representational substrate and predictive mechanism: observation-level generative models predict directly in perceptual spaces such as pixels or video; latent-space dynamics models encode observations into compact states and predict their evolution in representation space; and 3D or otherwise structured world representations explicitly model geometry, topology, objects, relations, or physical attributes. The first taxonomy therefore asks what a system does, whereas the second asks how the relevant world knowledge is represented and computed. They are complementary rather than competing, and a model should generally be assigned one or more labels along both axes.

Figure 1 makes this many-to-many mapping explicit. Sora and Seedance are placed in the observation-level renderer region because their externally exposed prediction targets are visually plausible videos rather than explicitly structured and independently queryable world states\citep{openai2024sora,gao2025seedance}. Genie 3 remains observation-level in its representation and output, but extends upward toward the 3D world representation\citep{genie3}: it generates the environment frame by frame in response to navigation commands, agent actions, and promptable world events, thereby supporting interactive rollouts, counterfactual exploration, and agent research and training. 
The JEPA family is positioned in the latent-space simulator region because its defining prediction target is an abstract future representation rather than reconstructed pixel \citep{assran2023ijepa,bardes2024vjepa}. VLA-JEPA extends this latent predictive mechanism toward the planner region by coupling action-relevant latent transitions to continuous action generation\citep{sun2026vlajepa}. Dreamer is also placed in the latent-space column, but follows a distinct generative state-space paradigm \citep{hafner2019dreamer}: it learns compact action-conditioned dynamics from observations and uses imagined latent trajectories to train its actor and critic for long-horizon behavior. At the system level, it therefore combines a latent simulator with an imagination-trained action policy. More generally, action-conditioned latent world models can extend from simulation toward planning when their predictions are coupled with an explicit planning or control objective, as demonstrated by V-JEPA 2-AC for image-goal robotic planning \citep{assran2025vjepa2}.
Marble lies predominantly in the 3D/structured representation region \citep{worldlabs2025marble,kerbl20233dgaussians} . Its Gaussian-splat representation supports high-fidelity rendering and novel-view synthesis, whereas its collider meshes provide simulation-ready geometry that downstream physics engines can operate on. It therefore bridges rendering and simulation, although the collision representation itself should not be conflated with a complete learned model of physical dynamics. 
Cosmos 3 is represented as a cross-cutting capability profile rather than a single point in the taxonomy \citep{nvidia2026cosmos3}. Within a shared omnimodal architecture, its audiovisual generation mode performs a renderer function, its action-conditioned forward-dynamics mode performs a simulator function, and its policy mode jointly predicts actions and their expected visual consequences. Its location in Figure 1 therefore reflects the fact that different input–output configurations and post-training procedures can expose different functional roles from a shared model backbone. Taken together, these systems suggest that increasingly capable world models are more appropriately characterized by their operational configuration and capability profile than by assigning an indivisible model name to a single category.

For the same reason, we do not treat the currently prominent World Action Model (WAM) as a fourth implementation category. As illustrated by the horizontal WAM region in Figure 1, WAM is a cross-architectural functional paradigm for embodied decision-making rather than a representational substrate parallel to the three architecture columns. Its defining commitment is the coupling of predictive state modeling with action generation. At the system level, a WAM models the relationship between anticipated future states and actions, even when this relationship is implemented through a cascaded factorization rather than a fully unified network.

A WAM may reason through explicit future video, as in visual planning pipelines; through compact latent future representations or trajectories; or through structured spatial representations such as optical flow, 3D point flows, RGB-D trajectories, or other 3D representations. It can therefore be instantiated using observation-level, latent-space, or structured world models. On Li’s functional axis, WAMs are primarily planner-facing, but they normally span both planner and simulator functions because future-state prediction is constitutive of, rather than merely auxiliary to, action generation. Treating WAM as a functional family centered on predictive action generation avoids conflating output function with implementation substrate and permits meaningful comparison between cascaded systems, which first predict future-state representations and then derive actions, and joint systems, which co-model state and action trajectories.}

\begin{figure}
    \centering
    \includegraphics[width=0.96\linewidth]{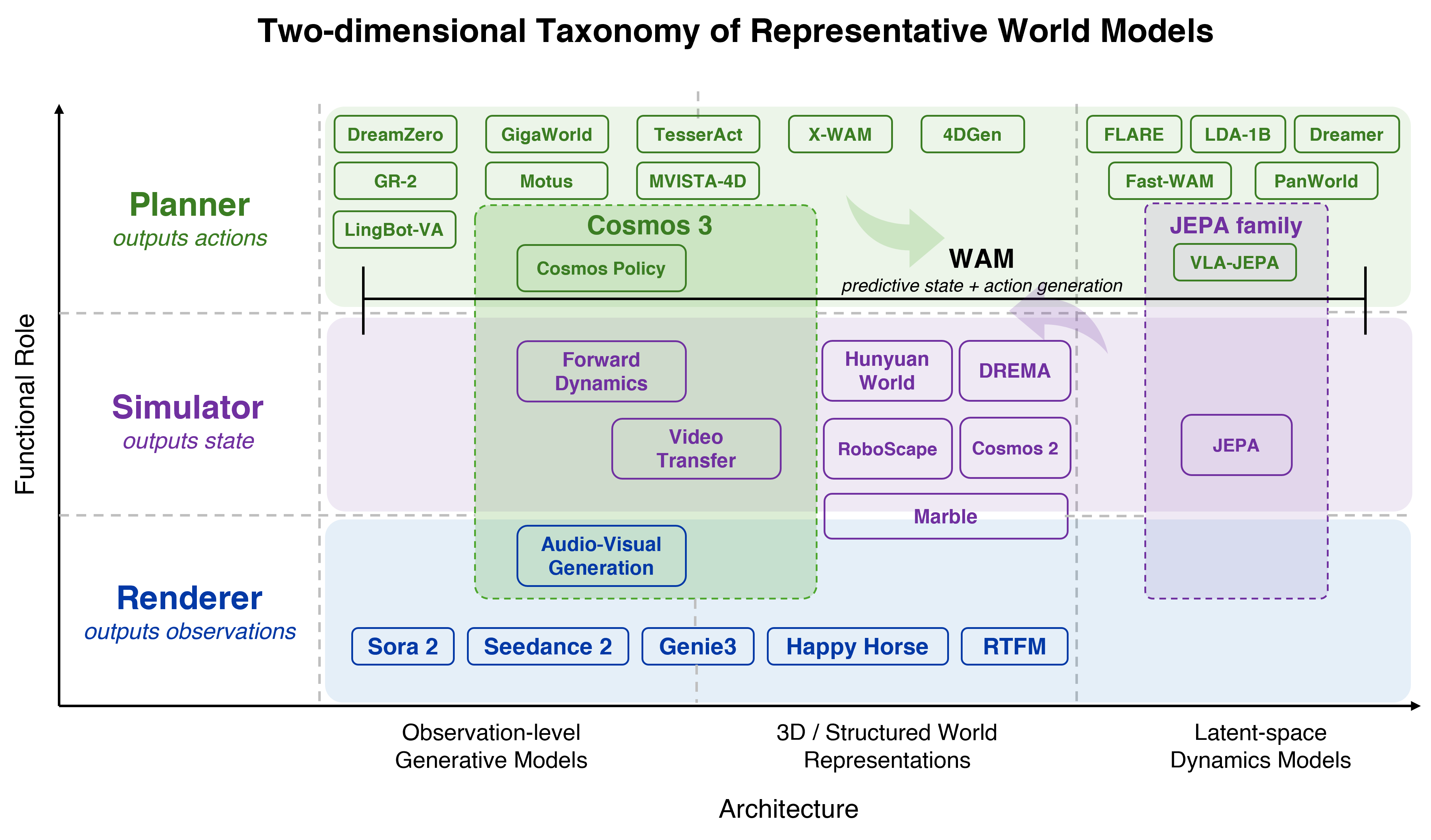}
    \caption{A Two-Dimensional Taxonomy: Functions and Architectures.}
    \label{fig:taxonomy}
\end{figure}

\subsection{Extending World Models Beyond Physical Environments}
{The preceding discussion is still largely confined to everyday macroscopic environments, including scenes, robots, vehicles, and manipulable objects. A broader research agenda should ask whether analogous world models can be constructed for scientific domains. In chemistry, depending on the task, the state may include molecular species and structures, reaction progress, and relevant kinetic or thermodynamic variables; observations may include spectra, chromatograms, or measured yields; and actions may correspond to experimentally controlled interventions. In biology, relevant states may span molecular networks, cellular phenotypes, tissues, and organism-level processes, potentially through hierarchical or multiscale representations, while interventions may include genetic perturbations, drug administration, and other controlled manipulations. In astronomy, latent astrophysical states are observed indirectly through instruments, while planning may involve selecting targets, instruments, and observation schedules on the basis of models of astrophysical dynamics and instrument-dependent observations.

Existing scientific models already illustrate parts of this formulation. In meteorology, GraphCast learns an autoregressive transition model over global atmospheric states and generates medium-range forecasts by repeatedly propagating hundreds of atmospheric variables forward in time \citep{lam2023graphcast}. Within the present taxonomy, it can be interpreted as a structured scientific simulator, although it is not an agentic world model because it does not select interventions or close the loop between prediction and action. At molecular scales, MDGen learns a generative model of molecular-dynamics trajectories and can be conditioned for forward simulation, transition-path generation, and trajectory upsampling \citep {jing2024mdgen}. It similarly exemplifies learned scientific dynamics modeling, but does not by itself constitute a complete experimental planning system. These cases demonstrate that individual components of world modeling—state representation, temporal propagation, uncertainty-aware generation, and conditional prediction—are already emerging across scientific domains, even when they are not explicitly described as world models.Across these domains, a scientific world model should not merely generate plausible measurements. It should capture mechanistic or otherwise scientifically meaningful structure, respect known physical constraints and invariants where applicable, quantify uncertainty, support counterfactual intervention, and help select informative or utility-maximizing experiments under practical constraints. A complete agentic scientific world model would further connect these capabilities to an experimental loop in which hypotheses or interventions are proposed, outcomes are predicted, experiments are executed, and the internal state of the model is updated from new evidence.

An even broader, and substantially more speculative, extension concerns bounded social systems \citep{zhou2025socialworldmodels}. We restrict this notion to systems with clearly defined participants, interaction rules, observation channels, intervention spaces, and time horizons, such as small teams, small organizations, or narrowly specified institutional processes. Such systems may be modeled as partially observable, multi-agent dynamical processes. Candidate latent variables may include incentives, beliefs, organizational structures, resource constraints, and interaction networks; observations may include communications, transactions, and performance traces; and interventions may include task allocation, organizational redesign, or negotiated actions. Within these boundaries, social world models could support scenario analysis for team coordination and organizational operations.

Before proceeding, we emphasize that the primary focus of this paper is on world models of the physical world rather than social systems. The discussion of social domains here is intended only as a conceptual extension to clarify scope and potential limitations, and should not be interpreted as the central object of study.

This formulation should not be extended without qualification to nations, populations, or society-wide prediction. Large-scale social systems are open, reflexive, non-stationary, and normatively contested: predictions and interventions can alter the behavior being modeled, agents may strategically respond to the model, and different groups may pursue incompatible objectives. Moreover, variables such as beliefs, preferences, and intentions are only partially observable and may depend on contestable theoretical assumptions. Historical predictive accuracy therefore does not guarantee reliable intervention effects under changing institutions or public awareness of the prediction.
The ethical and governance risks are substantial. Inferring latent states from communications and behavioral traces may enable intrusive surveillance, while historically biased data may reproduce discrimination or unequal risk allocation. Model outputs may also become performative when used to allocate opportunities, enforce policies, or shape behavior, thereby creating self-reinforcing feedback loops \citep{perdomo2020performative}. Social world models should therefore represent uncertainty explicitly, document their causal and normative assumptions, and be subjected to privacy protection, bias auditing, distribution-shift testing, and human oversight. They should be treated as bounded decision-support tools rather than social oracles, and should not be used for unconstrained population-scale forecasting, mass profiling, automated governance, or high-stakes decisions about individuals and social groups.

To avoid terminological dilution, the designation “world model” should be reserved for systems that maintain an internal representation of state, model temporal and, where relevant, intervention-conditioned transitions, and generate testable predictions. Under these restrictions, the two-di\-men\-sion\-al framework remains applicable to bounded and operationally defined social domains. An unconstrained model of an entire nation or society should not be regarded as a world model merely by virtue of its scale or predictive ambition.}

\section{Architectural Paradigms}
Following the representation-centered view of world models, this section classifies existing systems by how they represent and propagate world state. We distinguish observation-level models, which predict directly in pixels, voxels, or video tokens; latent-space models, which learn compact predictive states; and 3D-enhanced or object-centric models, which introduce explicit spatial, geometric, or entity-level structure. We then discuss a cross-paradigm unification trend that integrates generation, understanding, interaction, and multimodal grounding within shared architectures. The main trade-offs and representative systems are summarized in Table~\ref{tab:wm_architecture_comparison}.
\subsection{Observation-Level Generative World Models}

Observation-level generative world models treat world modeling as high-dimensional observation synthesis, directly predicting future pixels, voxels, or video tokens. Recent large-scale video generators, including Sora-style models~\citep{openai2024sora}, Wan~\citep{wan2025wan}, Happy Horse~\citep{happyhorse2026}, Seedance series~\citep{gao2025seedance,seedance2026seedance2}, Movie Gen~\citep{polyak2024moviegen}, HunyuanVideo~\citep{kong2024hunyuanvideo}, CogVideoX~\citep{yang2024cogvideox}, and Kling series~\citep{kling2025omni}, are not all world models in the strict decision-centric sense, but they represent an observation-level route to learning physical priors from large-scale video. More explicitly interactive systems, including the Genie series~\citep{bruce2024genie,parkerholder2024genie2,genie3}, GameNGen~\citep{valevski2025gamengen}, DIAMOND~\citep{alonso2024diamond}, and Oasis~\citep{decart2024oasis}, move from passive video continuation toward controllable environment generation. In physical and driving scenarios, Cosmos~\citep{nvidia2025cosmos}, GAIA-1~\citep{hu2023gaia1}, DriveDreamer~\citep{wang2023drivedreamer}, Drive-WM~\citep{wang2023drivewm}, and the Waymo World Model~\citep{waymo2026worldmodel} further adapt observation-level generation to trajectory-, action-, or scene-conditioned simulation.

Their central advantage is scale: internet video provides an enormous weakly supervised corpus for learning appearance, motion, object interaction, and scene dynamics. As a result, these models often achieve high visual fidelity and broad coverage of real-world phenomena. However, visual plausibility is not equivalent to physical correctness. Pixel-space models may generate locally convincing but globally inconsistent rollouts, especially over long horizons, where errors in object permanence, contact dynamics, causality, or scene consistency can accumulate. They are also computationally expensive, because prediction happens in a high-dimensional observation space. Systems such as Diffusion Forcing~\citep{chen2024diffusionforcing}, VideoPoet~\citep{kondratyuk2024videopoet}, Wan~\citep{wan2025wan}, and Cosmos~\citep{nvidia2025cosmos} therefore explore how the perceptual strength of large generative video models can be extended toward temporally coherent, sequence-level prediction. Thus, the central criterion for evaluating observation-level world models is not merely perceptual fidelity, but whether their generated trajectories remain physically consistent, causally coherent, and controllable under action, instruction, trajectory, or other conditioning signals.

\begin{table*}[t]
\centering
\setlength{\tabcolsep}{3pt}
\renewcommand{\arraystretch}{1.08}
\caption{Advantages and limitations of major world-model architectural paradigms.}
\label{tab:wm_architecture_comparison}
\vspace{.2cm}
\resizebox{\linewidth}{!}{
\begin{tabular}{@{} c | c | c | c @{}}
\hline
\textbf{Paradigm} & \textbf{Advantages} & \textbf{Limitations} & \textbf{Representative systems} \\
\hline
\makecell{Observation-level} &
\makecell{High visual fidelity;\\
intuitive outputs;\\
scales with internet video} &
\makecell{Expensive generation;\\
physical inconsistency;\\
long-horizon rollout drift} &
\makecell{\citep{openai2024sora}\\
\citep{bruce2024genie}\\
\citep{wan2025wan}\\
\citep{gao2025seedance}\\
\citep{nvidia2025cosmos}\\
\citep{chen2024diffusionforcing}} \\
\hline
\makecell{Latent space} &
\makecell{Efficient imagination planning;\\
compact latent states;\\
long-horizon scaling} &
\makecell{Loss of critical details;\\
harder to inspect visually;\\
objective mismatch} &
\makecell{\citep{ha2018worldmodels}\\
\citep{hafner2019planet}\\
\citep{hafner2019dreamer}\\
\citep{hafner2021dreamerv2}\\
\citep{hafner2023dreamerv3}\\
\citep{hansen2023tdmpc2}\\
\citep{schrittwieser2020muzero}\\
\citep{assran2023ijepa}\\
\citep{bardes2024vjepa}}\\
\hline
\makecell{3D/object-centric} &
\makecell{Better spatial consistency;\\
object-level reasoning;\\
viewpoint/occlusion handling} &
\makecell{Needs structured supervision;\\
struggles with dynamic scenes} &
\makecell{\citep{zheng2024occworld}\\
\citep{zhang2025robooccworld}\\
\citep{mildenhall2020nerf}\\
\citep{kerbl20233dgaussians}\\
\citep{worldlabs2025marble}\\
\citep{wu2023slotformer}\\
\citep{huang2025enerverse}} \\
\hline
\makecell{Cross-paradigm\\unification} &
\makecell{Combines generation, \\understanding,
interaction,\\ and multimodal grounding} &
\makecell{Early-stage;\\
modality alignment issues;\\
action fidelity;\\
open evaluation} &
\makecell{\citep{liu2025lwm}\\
\citep{ge2024worldgpt}\\
\citep{zhou2025hermes}\\
\citep{ye2026dreamzero}\\
\citep{zhu2025uwm}\\
\citep{nvidia2026cosmos3}} \\
\hline
\end{tabular}}
\end{table*}

\subsection{Latent-Space World Models}
{The transition from pixel-space to latent-space world models is not merely an efficiency improvement, but reflects a fundamental shift from modeling visual appearance to modeling task-relevant state evolution. Early latent world models such as PlaNet~\citep{hafner2019planet} and the Dreamer family~\citep{hafner2019dreamer, hafner2021dreamerv2, hafner2023dreamerv3} employ latent dynamics models, such as recurrent state-space models (RSSM), to encode high-dimensional observations into compact states and perform imagination-based planning efficiently. By rolling out compact latent states rather than full-resolution pixels, these approaches scale more naturally to long horizons, better handle partial observability, and allow agents to evaluate more hypothetical futures within a fixed compute budget.}

{Beyond efficiency, latent modeling changes the optimization target of world modeling. Instead of spending capacity on reconstructing every RGB detail, latent-space models can focus on slower-varying and decision-relevant factors that determine future controllability, such as geometry, object relationships, affordances, contact dynamics, and temporal evolution. Related model-based control methods such as TD-MPC and TD-MPC2~\citep{hansen2022tdmpc, hansen2023tdmpc2} further show that planning can be performed directly in learned task-oriented latent spaces, enabling efficient trajectory optimization without requiring a full pixel decoder. Similarly, contrastive and reconstruction-free world models~\citep{kipf2019cswm, okada2022dreamingv2, poudel2023recore} suggest that avoiding pixel-level reconstruction can improve robustness by reducing overfitting to high-frequency appearance nuisance factors, including texture, lighting, and small photometric variations. More recent approaches, including JEPA-style predictive representation learning~\citep{assran2023ijepa, bardes2024vjepa, assran2025vjepa2}, further challenge the necessity of full visual reconstruction by predicting representations rather than pixels. More broadly, embodied learning systems such as LDA-1B~\citep{lyu2026lda1b}, and Motus~\citep{bi2025motus} suggest that effective policies can often be learned from compact task-oriented latent representations without photorealistic reconstruction of the full visual scene. In this sense, latent world models may not only be faster, but also more aligned with control: by preserving low-frequency, causally relevant structure while suppressing superficial visual details, they can support more robust and accurate decision making.}

{However, the main risk is that excessive abstraction may discard information that is visually small but decision-critical, such as gripper-object contact, thin obstacles, subtle object pose changes, or deformation. Thus, the central challenge is not simply to compress observations, but to learn latents that discard appearance-level noise while preserving the factors that determine future controllability. This also shifts evaluation: latent world models should not be judged only by reconstruction or prediction error, but by whether their rollouts preserve the information needed for downstream planning and control.}

\subsection{3D-Enhanced and Object-Centric World Models}

3D-enhanced and object-centric world models push world modeling away from pure frame prediction and toward structured scene understanding. A conventional video model may generate visually plausible future frames, but its internal state is often hard to interpret and may not correspond to the actual 3D layout of the scene. For embodied AI, this is limiting: the model should also support reasoning about free space, object locations, occlusion, viewpoint changes, object permanence, and physical plausibility. For this reason, recent work has explored 3D occupancy, BEV representations, NeRF-like radiance fields, 3D Gaussian Splatting, and object slots as more structured state spaces for world models~\citep{mildenhall2020nerf,kerbl20233dgaussians,locatello2020slotattention,wu2023slotformer,zheng2024occworld,zhang2024bevworld,zhang2025robooccworld}.

One direction is occupancy- and BEV-based world modeling. These representations expose the spatial structure of a scene more directly than pixels, making them useful for reasoning about free space, obstacles, semantic regions, and future scene evolution. In autonomous driving, systems such as OccWorld, OccSora, Drive-OccWorld, and BEVWorld learn future scene dynamics in occupancy or BEV-like spaces~\citep{zheng2024occworld,wang2024occsora,yang2024driveoccworld,zhang2024bevworld}. In robotics, RoboOccWorld extends this structured-state view to indoor embodied prediction by forecasting 3D semantic occupancy from spatial observation history and future camera poses~\citep{zhang2025robooccworld}. The key advantage of this family is that predictions become spatially queryable rather than only visually plausible. Another family uses continuous or point-based 3D scene representations. NeRF and 3D Gaussian Splatting provide geometry-aware states that support novel-view rendering and stronger viewpoint consistency~\citep{mildenhall2020nerf,kerbl20233dgaussians}. Recent systems such as Marble, RenderWorld, GaussianWorld, and GWM extend this idea toward persistent 3D world generation, driving-world modeling, streaming occupancy prediction, and robotic manipulation~\citep{worldlabs2025marble,yan2024renderworld,zuo2025gaussianworld,lu2025gwm}. Compared with pure video generation, these models better preserve spatial structure across views and time, although they still face challenges in dynamic scenes and long-horizon interaction. Object-centric world models address a complementary problem: even a dense 3D state may not isolate the entities that matter. Methods such as OP3, C-SWM, Slot Attention, SAVi, and SlotFormer learn object- or slot-level representations and dynamics, supporting more compositional prediction and reasoning~\citep{veerapaneni2019op3,kipf2019contrastive,locatello2020slotattention,kipf2022savi,wu2023slotformer}. Recent robotic systems such as FOCUS and Object-Centric World Model for Language-Guided Manipulation further connect object-level abstraction to manipulation and instruction-conditioned prediction~\citep{ferraro2023focus,jeong2025objectcentricwm}. This abstraction is useful because many embodied tasks are naturally defined around persistent entities and their relations, rather than global image patterns.

The most promising direction is to combine these representations rather than treat them as competing choices. Occupancy supports free-space and collision reasoning, NeRF and 3D Gaussian representations support view-consistent rendering, and object slots support compositional reasoning. Hybrid embodied systems such as EnerVerse and EnerVerse-AC already point in this direction by connecting multi-view prediction, memory, 4D scene representation, instruction or action conditioning, and embodied future generation~\citep{huang2025enerverse,jiang2025enerverseac}. The broader lesson is that future embodied world models may need to combine spatial geometry, temporal memory, object structure, and interaction grounding within a shared physical representation.

At the same time, adding 3D structure does not automatically solve world modeling. A model may reconstruct space accurately but still fail to capture affordances, causality, object permanence, or task-relevant semantics. Object-centric models are compact and interpretable, but reliable object discovery remains difficult in cluttered scenes and under occlusion. NeRF or 3D Gaussian models can render detailed views, but scaling them to dynamic scenes, long-horizon prediction, and closed-loop interaction remains challenging. Therefore, the key question is not whether a world model looks more 3D, but whether its state forms a useful and coherent representation of the physical world. A strong embodied world model should be geometrically consistent, temporally stable, object-aware, physically plausible, and accessible to downstream reasoning, planning, or control modules.

\subsection{The Unification Trend: Omnimodal World Models}
\begin{figure}[t]
    \centering
    \includegraphics[width=\linewidth]{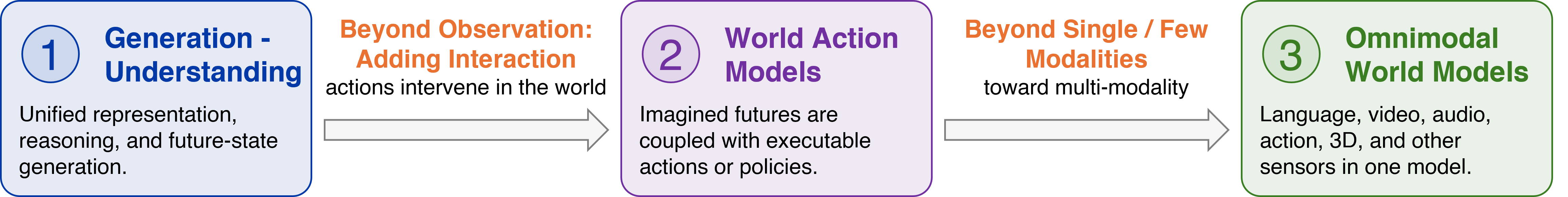}
    \caption{Unification trend of world-model architectures. Generation--understanding unified models first move beyond passive observation by incorporating interaction and action grounding, giving rise to World Action Models (WAMs). The paradigm then further moves beyond single- or few-modality modeling toward omnimodal models that integrate language, video, audio, action, 3D structure, and other sensor signals within a shared physical representation.}
    \label{fig:omnimodal_world_model_trend}
\end{figure}
{A growing consensus in world-model research is that future world models should not be pure generators, but generation-understanding unified models. A model of the world should be able not only to synthesize future observations, but also to understand objects, spatial relations, temporal dynamics, causality, and physical plausibility. This direction is more explicitly reflected in systems such as LWM~\citep{liu2025lwm}, WorldGPT~\citep{ge2024worldgpt}, HERMES~\citep{zhou2025hermes}, HERMES++~\citep{zhou2026hermespp}, GaussianDWM~\citep{deng2026gaussiandwm}, and UniDrive-WM~\citep{xiong2026unidrivewm}, which attempt to connect scene or video understanding with future-state generation within a unified framework. Adjacent multimodal foundation models such as Chameleon~\citep{chameleon2025}, Janus/Janus-Pr~\citep{wu2024janus, chen2025januspro}, and Emu series~\citep{wang2024emu3, cui2025emu35} further support this trend by showing that understanding and generation can be handled within shared token-based or autoregressive architectures. Together, these models suggest that world modeling is moving beyond visually appealing synthesis toward architectures that jointly represent, reason about, and generate dynamic worlds.}

{However, the physical world is not only observed; it is also acted upon. From a partially observable decision-making perspective, an embodied agent only receives partial observations of the underlying world state, while its actions intervene on the latent state transition and shape future observations, as illustrated in Fig.~\ref{fig:omnimodal_world_model_trend}. 
Therefore, generation-understanding unification is not sufficient by itself. A world model for physical AI should not only represent and generate future states, but also connect those imagined futures to executable actions. This motivates the emergence of World Action Models (WAMs)~\citep{wang2026wam_survey}, which integrate world modeling with action or policy modeling within a shared generative framework. Rather than treating dynamics prediction and policy learning as separate modules, WAMs aim to learn their coupling directly from video-action data. For example, DreamZero~\citep{ye2026dreamzero} shows that jointly modeling video and action can turn a world model into a zero-shot policy; LingBot-VA~\citep{li2026lingbotva} unifies visual dynamics prediction and action inference within an autoregressive video-action framework; and $\tau$0-WM~\citep{zhou2026tau0wm} integrates action generation, video prediction, and future-state evaluation for robotic manipulation. Related systems such as UWM~\citep{zhu2025uwm}, Cosmos Policy~\citep{kim2026cosmospolicy}, GigaWorld-Policy~\citep{ye2026gigaworldpolicy}, Fast-WAM~\citep{yuan2026fastwam}, and Flash-WAM~\citep{akbari2026flashwam} further explore different ways of coupling video prediction with action generation, including joint diffusion, action-centered prediction, and faster inference without explicit test-time video imagination. Together, these models mark a shift from passive world simulation toward generative models that can both imagine physical futures and produce actions grounded in those futures.}

Besides, world models are expanding from single- or few-modality settings toward omnimodal physical modeling, integrating language, image, video, audio, action, 3D structure, and sensor data within a shared representation of the physical world, as shown in Fig.~\ref{fig:omnimodal_world_model_trend}. NVIDIA Cosmos 3~\citep{nvidia2026cosmos3} can be viewed as a clear exemplar of this broader attempt: it explicitly aims to unify language, image, video, audio, and action within a Mixture-of-Transformers architecture, moving beyond vision-centric generation toward a general-purpose physical AI backend. Other recent systems provide important building blocks for this direction, even if they are not yet complete omnimodal world models. Seedance 2.0~\citep{seedance2026seedance2} explores unified multimodal audio-video generation from text, image, audio, and video inputs; Wan/Wan2.1~\citep{wan2025wan} provides scalable open video-generation backbones; Tencent's HunyuanVideo and HunyuanWorld series~\citep{kong2024hunyuanvideo,hunyuanworld2025} extend video generation toward explorable and interactive 3D world construction; Kling-Omni~\citep{kling2025omni} integrates multimodal instruction following, video generation, editing, and reasoning; and Qwen-Omni~\citep{qwen2025omni} shows how language, vision, audio, and video can be processed within a unified omnimodal foundation model. Together, these works suggest that generation, understanding, interaction, spatial reasoning, and multimodal physical prediction are beginning to converge into shared model backbones. In this sense, the unification trend is driven by physical AI's need for a single model that can understand the present, imagine plausible futures, interact with the world, and integrate multimodal physical signals.

\section{Training and Learning Paradigms} 

\begin{figure}[t]
    \centering
    \includegraphics[width=\linewidth]{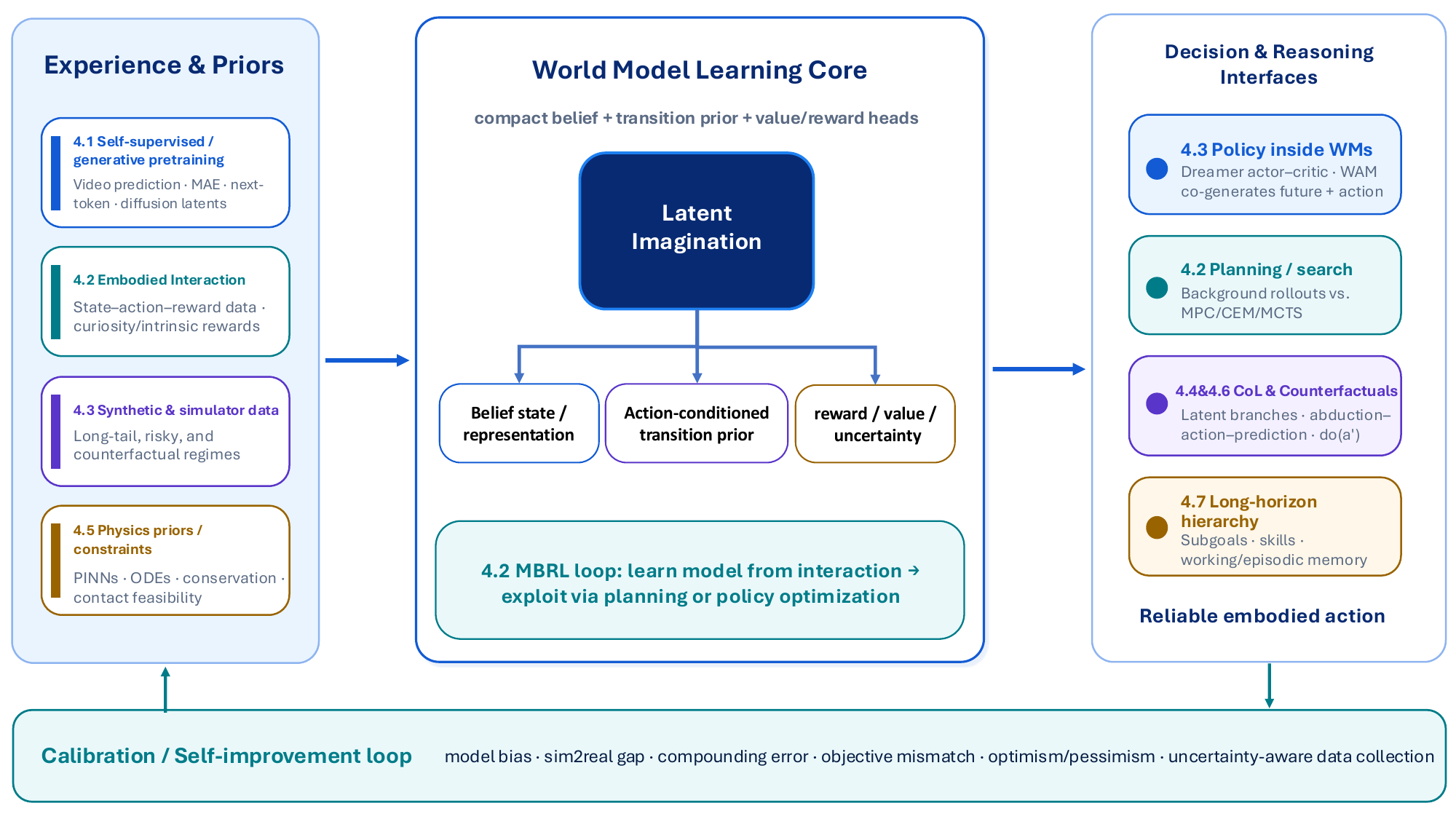}
    \caption{An overview of major training and learning paradigms for world models. }
    \label{fig:training-paradigm}
\end{figure}

\subsection{Self-Supervised and Generative Pretraining}

Self-supervised and generative pretraining has become the natural entry point for world models because it turns unlabelled experience into dense supervision before task rewards or dense expert labels are available. Let $\mathbf{x}_{1:T}$ denote pixel-level observations or video frames, $\mathbf{a}_{1:T}$ actions, $\mathbf{c}$ optional conditions, and $u_{1:T}$ discrete tokens. The common objective is not reconstruction for its own sake, but the acquisition of a compact belief state and a transition prior by predicting withheld, future, or next elements of experience.

Video prediction is the most direct formulation:
\begin{equation}
    p_{\theta}\!\left(
    \mathbf{x}_{t+1:t+K}
    \mid
    \mathbf{x}_{\le t},
    \mathbf{a}_{t:t+K-1},
    \mathbf{c}
    \right).
\end{equation}
It grounds representation learning in time: persistence, occlusion, contact, motion, and scene change become supervision at every frame. Yet action-free prediction learns observational dynamics, whereas action-conditioned prediction begins to approximate interventional dynamics---the distinction that matters for planning. Pixel prediction is also a fragile proxy: photorealism is neither necessary nor sufficient for controllable physical understanding. Modern systems therefore increasingly predict stochastic, tokenized, or diffusion latents, shifting the goal from rendering plausible futures to preserving action-relevant dynamics~\citep{finn2016unsupervised,ha2018worldmodels}.

Masked autoencoding provides a complementary route. MAE-style objectives reconstruct masked patches from visible context, while video variants exploit temporal redundancy through tube masking and very high mask ratios~\citep{he2022mae,tong2022videomae}. The useful effect is compression: the encoder cannot simply copy pixels, and must infer structure from context. But reconstruction is only a proxy. For world models, the best masked objectives are those whose latent spaces remain predictive under intervention; this motivates JEPA-style representation prediction, which discards nuisance appearance while preserving semantic and dynamical invariants useful for planning~\citep{bardes2024vjepa,assran2025vjepa2}.

Next-token prediction supplies the most scalable generative interface. Once video, actions, language, maps, proprioception, and rewards are tokenized or embedded, pretraining can minimize
\begin{equation}
\mathcal{L}_{\mathrm{NTP}}
=
-\sum_{t=1}^{T}
\log p_\theta\!\left(
u_t \mid u_{<t}, \mathbf{a}_{\le t}, \mathbf{c}
\right).
\end{equation}
This aligns world modeling with Transformer scaling and enables multimodal prompting. GAIA-1 casts autonomous-driving world modeling as discrete-token sequence prediction, while Genie combines a spatiotemporal video tokenizer, an autoregressive dynamics model, and a latent action model trained from unlabelled videos~\citep{hu2023gaia1,bruce2024genie}. The hidden bottleneck is tokenization: tokens define the model's ontology, and geometry, contact, or controllability discarded by the tokenizer cannot be recovered by scale alone.

Scaling laws turn these objectives into a systems problem. A useful abstraction is
\begin{equation}
\mathcal{L}(N,D)
\approx
\mathcal{L}_{\infty}
+
A N^{-\alpha}
+
B D^{-\beta},
\qquad
C \approx \kappa N D,
\end{equation}
where $N$ is model size, $D$ training data, and $C$ compute. Language-model results show predictable power-law improvements with scale, while compute-optimal training requires balancing parameters and tokens~\citep{kaplan2020scaling,hoffmann2022training}. For world models, however, $D$ should be read not merely as token count, but as coverage over states, actions, embodiments, viewpoints, contacts, rare events, and horizons. Early studies of embodied pretraining find analogous scaling behavior, but with coefficients shaped by tokenizer, task, and architecture~\citep{pearce2024scaling}. Thus, likelihood is insufficient: scalable world models must also be judged by closed-loop consistency, controllability, long-horizon stability, and robustness to rare regimes.

Synthetic data is therefore a practical scaling lever. Platforms such as Omniverse Replicator and Isaac Sim support physically based synthetic-data generation, while recent pipelines such as Cosmos-Drive-Dreams and GR00T-Dreams use world foundation models to generate controllable driving scenarios or synthetic robot trajectories~\citep{nvidia2022replicator,nvidia2025cosmos,nvidia2025grootdreams}. Their strongest role is not to replace reality, but to amplify it: real data anchors the distribution, whereas simulation and generative SDG expose long-tail, dangerous, and counterfactual regimes. The central problem is calibration---how to mix real, simulated, and generated data so that broader coverage improves physical generalization rather than teaching simulator artifacts. In this view, pretraining does not teach a model to render the world; it teaches the variables through which an agent can change it.

\subsection{Model-Based Reinforcement Learning}
\label{sec:mbrl}
Model-based reinforcement learning (MBRL)~\citep{moerland2022modelbasedreinforcementlearningsurvey} is a learning paradigm in which world models were first given a decision-centric definition: a world model is an internal model used to improve a policy by predicting the consequences of actions well enough.
The typical pipeline of MBRL proceeds in two interleaved stages: 1) an agent interacts with the environment and fits a dynamics model, i.e., a learned transition model $\widehat P_\theta$, often accompanied by learned value functions; 2) the agent exploits this model as a cheap, differentiable, and resettable surrogate of reality, either by planning action sequences directly inside it or by training a policy on imagined rollouts, thereby amortizing the cost of real interaction.

In formulation, MBRL is defined as a tuple $\mathcal{M}_{\mathrm{MDP}} = (\mathcal{S}, \mathcal{A}, P^\star, R, \gamma, \rho_0)$,
where $\mathcal{S}$ the state space, $\mathcal{A}$ the action space, $P^\star(\mathbf{s}_{t+1} \mid \mathbf{s}_t, \mathbf{a}_t)$ denotes the true transition dynamics, $R: \mathcal{S} \times \mathcal{A} \to \mathbb{R}$ the reward function, $\gamma \in [0,1)$ the discount factor, and $\rho_0$ the initial state distribution.
For the learning objective, the agent seeks a policy $\pi_\phi: \mathcal{S} \to \Delta(\mathcal{A})$ that maximizes the expected cumulative return as
\begin{equation}
    J(\pi_\phi)
    =
    \mathbb{E}_{\pi_\phi,P^\star}
    \left[
    \sum_{t=0}^{H-1}\gamma^t r_t
    \right],
    \qquad
    r_t = R(\mathbf{s}_t,\mathbf{a}_t).
\end{equation}
The key distinction is the learning of an \textbf{approximate transition function} $\widehat P_\theta$, parameterized by $\theta$, to approximate the oracle $P^\star$:
\begin{equation}
    \widehat P_\theta:
    \mathcal{S}\times\mathcal{A}\to\Delta(\mathcal{S}),
    \qquad
    \widehat{\mathbf{s}}_{t+1}
    \sim
    \widehat P_\theta(\cdot\mid\mathbf{s}_t,\mathbf{a}_t),
\end{equation}
where $\widehat P_\theta$ is trained by minimizing the negative log-likelihood over a replay buffer $\mathcal{D} = \{(\mathbf{s}_t, \mathbf{a}_t, \mathbf{s}_{t+1})\}$ as
\begin{equation}
\label{eq:learn_dynamic_mbrl}
    \mathcal{L}_{\mathrm{dyn}}(\theta)
    =
    -\mathbb{E}_{(\mathbf{s},\mathbf{a},\mathbf{s}')\sim\mathcal{D}}
    \left[
    \log \widehat P_\theta(\mathbf{s}'\mid\mathbf{s},\mathbf{a})
    \right].
\end{equation}
By learning $\widehat P_\theta$, the agent can perform policy
optimization entirely within the model, substantially improving 
\textbf{sample efficiency} compared to model-free methods.

Existing MBRL research focuses on two directions: \textit{model learning} and \textit{model exploitation}.
For model learning, early works such as PILCO~\citep{deisenroth2011pilco} employed Gaussian Processes to capture epistemic uncertainty in the dynamics, while subsequent approaches adopted DNNs for greater expressiveness---including deterministic  networks~\citep{nagabandi2018mbmf}, probabilistic ensembles~\citep{chua2018pets, janner2019mbpo}, and latent-space world models that jointly learn compact state  representations and dynamics~\citep{hafner2019dreamer, hafner2020dreamerv2,  hafner2023dreamerv3}.
For model exploitation, methods diverge in how they 
leverage the learned model: Dyna-style approaches~\citep{sutton1990dyna, 
janner2019mbpo} use the model as a generator to augment the 
replay buffer for off-policy policy optimization; planning-based methods such as PETS~\citep{chua2018pets} and MPPI~\citep{williams2017mppi} perform 
\textit{online trajectory optimization} via model predictive control (MPC); 
while model-based policy gradient methods~\citep{heess2015svg, clavera2020mbmpo} backpropagate gradients \textit{through model rollouts} to directly optimize the policy.
Together, these two lines of work highlight the central challenge  of MBRL: learning a sufficiently accurate model while efficiently exploiting it for downstream policy optimization. The reasoning and decision-making procedures described next should therefore be understood as concrete forms of model exploitation: they differ mainly in whether the model is used during background training, decision-time planning, or causal diagnosis.

\paragraph{Reasoning and Decision-Making as Model Exploitation.}
Possessing a world model is not the same as using one well. The literature divides model usage into two regimes that \citet{sutton2018rl} distinguish as \emph{background planning} and \emph{decision-time planning}. In background planning, the model serves as a data generator: imagined trajectories are treated as additional experience for policy improvement, as in Dyna-style updates \citep{sutton1991dyna}, the short model-based rollouts of MBPO that are deliberately truncated to bound model error \citep{janner2019mbpo}, and Dreamer's actor--critic learning in latent imagination \citep{hafner2019dreamer}. The computational cost is paid during training, and at deployment the agent acts reactively through its amortized policy. In forward search, by contrast, the model is interrogated at decision time: given the current state, the agent simulates candidate futures and selects the action whose imagined consequences score best. Decision-time planning adapts instantly to changed goals and exploits additional compute at inference, but inherits the model's errors afresh at every step. The two regimes are complementary rather than competing---modern systems such as MuZero use a learned policy to propose actions and search to refine them---and the trade-off between amortized reaction and deliberative search is increasingly read as an analogue of the System~1/System~2 distinction in cognition \citep{kahneman2011thinking}.
 
Forward search itself spans two methodological traditions. In continuous control, planning is posed as trajectory optimization over action sequences, typically embedded in model-predictive control (MPC): the agent optimizes an $H$-step action sequence against the model, executes only the first action, and replans at the next step, so that frequent feedback corrects for model drift. The inner optimization is often performed with the cross-entropy method (CEM), a derivative-free sampler that iteratively refits a sampling distribution to the elite fraction of imagined trajectories, as in PETS \citep{chua2018pets}, with refinements such as model-predictive path integral control \citep{williams2017mppi} and hybrid schemes that warm-start sampling with a learned policy and value function, as in TD-MPC \citep{hansen2022tdmpc}. In discrete domains, the dominant tool is Monte Carlo tree search, which incrementally builds a search tree guided by learned policy priors and value estimates \citep{silver2016alphago, silver2017alphagozero}; MuZero demonstrated that the search can run entirely inside a learned implicit model, and Stochastic MuZero and Sampled MuZero extend the recipe to stochastic dynamics and large or continuous action spaces \citep{antonoglou2022stochasticmuzero, hubert2021sampledmuzero}.
 
Three failure modes recur whenever an imperfect model is pressed into service for decision-making, and they motivate much of the near-term research agenda. The first is \emph{compounding error}: autoregressive rollouts feed the model its own predictions, so one-step inaccuracies accumulate and the imagined state distribution drifts from anything seen in training, with error growing superlinearly in horizon \citep{talvitie2017selfcorrecting, lambert2022compounding}. Practical systems respond by keeping rollouts short \citep{janner2019mbpo}, replanning frequently under MPC, weighting imagined returns with $\lambda$-style multi-horizon estimates, or penalizing trajectories by ensemble disagreement---but principled long-horizon rollout stability remains open, and we treat it as a first-class bottleneck for video-generation-based world models as well. The second is \emph{objective mismatch}: the model is trained to maximize predictive likelihood, yet it is evaluated by the return of the policy or plan it supports, and the two objectives are not aligned---a model that spends capacity on visually salient but decision-irrelevant detail can be a worse planning substrate than a less accurate but decision-relevant one \citep{lambert2020objectivemismatch}. Value-equivalent and decision-aware model learning \citep{grimm2020valueequivalence, farahmand2017vaml} respond by training the model directly on the quantities the planner consumes, which is precisely the design philosophy MuZero embodies. The third is the \emph{optimism--pessimism bias}: a planner is an adversarial consumer of its own model, actively seeking action sequences with high predicted return, and will therefore systematically discover and exploit the model's optimistic errors---an effect sometimes described as model exploitation or, in the offline setting, as the planner ``hallucinating'' value in regions without data support. In online learning, calibrated optimism in the face of uncertainty is a feature, driving exploration \citep{curi2020hucrl}; in offline or safety-critical settings it is a liability, and conservative methods such as MOPO and MOReL subtract an uncertainty penalty from imagined rewards or terminate rollouts that leave the data manifold \citep{yu2020mopo, kidambi2020morel}. Choosing where a system should sit on this optimism--pessimism spectrum is not a hyperparameter detail but a statement about how much the agent trusts its model of the world---and, as we argue in the following sections, the question generalizes verbatim from MBRL to foundation-scale world models deployed in physical environments.

Two representative paradigms instantiate these forms of model exploitation. Muzero~\citep{schrittwieser2020muzero}, unlike conventional MBRL methods that explicitly approximate the transition function $\widehat P_\theta$ over physical states, it learns a value-equivalent latent dynamics model and performs Monte Carlo tree search entirely within the latent space, achieving superhuman performance in games without requiring explicit environment priors.
Besides, Dreamer family~\citep{hafner2019dreamer, hafner2020dreamerv2, hafner2023dreamerv3}, adopts a RSSM (recurrent state-space model) to imagine latent rollouts and optimizes a policy via actor-critic learning, demonstrating high sample efficiency in visual continuous control.
While both paradigms have significantly adanced MBRL, their world models remain task-specific and limit to controllable situation, requiring training from scratch for each new task.
Recent work therefore explores generalist world models that are pre-trained on large-scale heterogeneous data.
DreamZero~\citep{ye2026dreamzero}, for instance, builds a World-Actor-Model (WAM) upone a pretrained video diffusion backbone, jointly predicting future video frames and robot actions.
It treats video as a dense representation of physical dynamics, learns diverse skills without repetitive demonstrations, and achieves zero-shot tasks generalization as well as few-shot embodiment adaptation in physical situations, while maintaining real-time closed-loop control at 7Hz.
This line of work extends MBRL from environment-specific model learning to pre-trained genralist world models, positioning the world model not merely as a proxy for policy improvement but as a foundation encapsulating broad physical priors.

\paragraph{Sim2Real Gap in MBRL.}
A fundamental obstacle to MBRL methods deployment in real-world embodied systems is the sim2real gap, i.e., the discrepancy between the dynamics learned by $\widehat P_\theta$ and the true physical dynamics $P^\star$ of the target environment.
In MBRL, this gap manifests at two levels: (i) \textit{model bias} due to imperfect learning of $\widehat P_\theta$, which introduces systematic prediction errors even within the training distribution; and (ii) \textit{distribution shift}, where the states visited by the policy $\pi_\phi$ during deployment diverge from those encountered during model training, causing the learned model to extrapolate unreliably~\citep{janner2019mbpo,ross2011dagger}.
We can formulate this problem from the lens of model discrepancy and policy deployment.

We summarize the approaches to reduce the gap is to improve the fedility during model exploitation or the robustness during policy execution.
DreamZero~\citep{ye2026dreamzero} takes the most direct approach to this issue. By building the world model upon a pretrained video diffusion backbone trained on large-scale real-world vidoe data, DreamZero's dynamics prior is anchored to genuine physical observations rather than simulated ones.
However, DreamZero does not really resolve the covariate shift and compounding error from the theoretical perspective, but it has been mitigated through technical means.
However, such paradigm also poses its limitations from some aspects:
1) computational overhead during pre-training, but still far below the high-frequency control rates required for dynamicor dexterous manipulation tasks; 2) data-hungry at the embodiment level as it realies on learning from full real-world data; 3) the quality of action prediction is fundamentally tied to the quality of video generation; 4) the current framework does not support multi-embodiment joint training, limiting the scalability of its cross-embodiment generalization.

Because MBRL agents possess an explicit predictive model, they also admit a principled answer to the exploration problem: seek experience where the model is wrong. Curiosity-driven exploration operationalizes this by converting model error or model uncertainty into an \emph{intrinsic reward}. Prediction-error methods such as the intrinsic curiosity module reward the agent in proportion to its forward-model surprise in a learned feature space \citep{pathak2017curiosity}, while random network distillation provides a robust proxy for state novelty \citep{burda2019rnd}. Information-theoretic formulations recast exploration as active learning of the dynamics, rewarding expected information gain about model parameters \citep{houthooft2016vime}, and Plan2Explore shows that an agent can use its world model to \emph{plan to be surprised}, seeking out states where an ensemble of latent dynamics models disagrees, thereby learning a task-agnostic world model that transfers zero-shot to downstream rewards \citep{sekar2020plan2explore}. Intrinsic rewards thus play a dual role in the world-model research agenda: they are both a mechanism for acquiring the sparse embodied interaction data that internet video cannot supply, and an early example of the model itself directing the data collection that improves it---a self-improvement loop we return to throughout this article.

\subsection{Policy Learning Inside World Models}
Before turning to how imagined futures are converted into behavior, it is worth making explicit a structural distinction that separates classical MBRL from the WAM paradigm introduced above.
Traditional MBRL, as formalized above, decouples model learning from policy optimization: an explicit transition model $\hat{P}_\theta$ is first fit to approximate $P^\star$ by minimizing a
state-prediction likelihood (Eq.~\eqref{eq:learn_dynamic_mbrl}), and only afterward is this model exploited---through background planning, decision-time search, or actor-critic learning on imagined rollouts---to improve a separately parameterized policy $\pi_\phi$.
WAMs depart from this two-stage recipe by coupling dynamics prediction and action generation within a single generative process, learned jointly from video-action data rather than from a dedicated $(s,a,s')$ transition objective~\citep{ye2026dreamzero,wang2026wam_survey}.
In a WAM, the predicted future state and the action that realizes it are co-modeled---either through a cascaded factorization that first imagines a future and then derives an action, or through a fully joint distribution over states and actions~\citep{ye2026dreamzero}---so that, at the system level, there is no longer a clean separation between ``the model'' and ``the policy.''

This architectural shift trades one set of strengths for another.
Decoupled MBRL retains an inspectable, swappable dynamics model whose training objective can be made decision-aware~\citep{grimm2020valueequivalence,farahmand2017vaml} and whose epistemic uncertainty can be explicitly quantified and exploited for conservative planning~\citep{chua2018pets,janner2019mbpo}, but it is correspondingly more exposed to objective mismatch between predictive likelihood and downstream control performance~\citep{lambert2020objectivemismatch}, and its dynamics model
is typically task- and embodiment-specific, offering little purchase on the
internet-scale, largely action-free video that dominates available physical data.
WAMs, conversely, inherit broad physical priors directly from video-action pretraining and have been shown to support zero-shot task generalization and few-shot embodiment adaptation while sustaining real-time closed-loop control~\citep{Ye2026b}, but at the cost of the modularity and calibrated uncertainty that decoupled model learning affords: the joint formulation makes it harder to isolate dynamics error from policy error, the fidelity of the predicted action is tied to the fidelity of the underlying video-generation backbone, and---as discussed above---current instantiations remain comparatively data-hungry at the embodiment level and ill-suited to the kind of principled out-of-distribution detection that ensemble- or uncertainty-based MBRL provides.

A world model becomes useful for embodied intelligence not only when it predicts future observations, but when its internal futures can be transformed into actions. This shifts the emphasis from model exploitation in classical MBRL to policy learning \emph{inside} the model's representational space. The Dreamer family is the canonical example. Dreamer learns a recurrent state-space model (RSSM) that compresses observation histories into stochastic latents, predicts rewards and continuation, and unrolls latent trajectories under candidate actions~\citep{hafner2019dreamer,hafner2020dreamerv2,hafner2023dreamerv3}. Actor and critic networks are trained on these imagined trajectories: value estimates provide long-horizon learning signals, while the actor amortizes planning into a reactive policy. Thus, the model is not merely a synthetic data generator; it defines the state abstraction, imagination horizon, and gradient-bearing interface through which behavior is improved. DreamerV2 and DreamerV3 further show that this recipe can scale beyond continuous control through discrete latents, robust reward/value transformations, and mostly fixed hyperparameters across diverse domains.

In embodied foundation models, the question reappears with a different interface: should the policy directly decode actions, or should action be grounded in explicit future prediction? Vision-Language-Action (VLA) models, such as RT-2~\citep{zitkovich2023rt}, extend pretrained vision-language models by representing robot actions as output tokens conditioned on images and language. Their strength is semantic transfer: web-scale pretraining helps robots interpret objects, instructions, and relations that are rare in robot datasets. Yet the physical dynamics connecting an action to its future consequence remain mostly implicit in the action-token mapping. World Action Models (WAMs) make this channel explicit. Following the embodied-world-model view that world models can serve as data engines, policy evaluators, planning modules, or on-device robot brains~\citep{shang2026survey}, WAMs jointly predict future perceptual states and actions, treating video as a dense carrier of physical change. DreamZero exemplifies this paradigm: built on a pretrained video diffusion backbone, it co-generates future video and robot actions, so the policy becomes a readout of imagined world evolution rather than only a direct imitation head~\citep{ye2026dreamzero}. In short, VLAs ask ``what action token should follow this instruction and observation?'', whereas WAMs ask ``what future should occur, and what action sequence realizes it?''

Therefore, policy learning inside world models is adjacent to, but broader than, MBRL. MBRL supplies the decision-theoretic foundation: a predictive model should improve control. This subsection emphasizes the architectural evolution of that idea: from task-specific latent imagination in Dreamer, to foundation-model policies in VLAs, to WAMs where prediction and action generation are coupled. The same risks remain---model bias, compounding error, latency, and distribution shift---but the world model changes from an auxiliary transition approximator to the computational substrate in which policies are represented, optimized, and generalized.

\subsection{Chain-of-Imagination: Reasoning Through World Models}

Classical world models use imagination as a planning substrate: the agent compresses experience into latent states, rolls out possible futures, and optimizes actions or values inside the learned simulator \citep{ha2018worldmodels,hafner2019dreamer,Schrittwieser2020}. Chain-of-Imagination (CoI) marks a sharper transition. The world model is no longer only a simulator queried by an external planner; it becomes the computational medium in which reasoning itself unfolds. In this view, a thought is not necessarily a sentence. It can be an action-conditioned transition in a learned dynamical space.

This reframing is motivated by latent reasoning. Chain-of-thought prompting showed that intermediate computations can make difficult inference tractable \citep{wei2022chain}, but natural language is an inefficient substrate for embodied decision making: it serializes geometry, dynamics, uncertainty, and affordances into tokens that may remain weakly coupled to control. Coconut makes the representational issue explicit by allowing LLMs to feed continuous hidden states back as ``continuous thoughts,'' avoiding premature commitment to discrete text and preserving unresolved alternatives in latent space \citep{hao2024coconut}. For embodied agents, the analogous workspace is a spatiotemporal belief state,
\begin{equation}
    \mathbf{z}_t
    =
    E_\eta(\mathbf{o}_{\le t},\mathbf{a}_{<t}),
\end{equation}
which summarizes the current scene, interaction context, and uncertainty before the policy commits to action.

A generic CoI trace should therefore distinguish deliberation depth from hypothesis diversity. Let $B_t$ denote the number of imagined branches and $K_b$ the rollout depth of branch $b$. Then
\begin{equation}
    \mathcal{C}_t =
    \left\{
    \left[
    (\mathbf{a}_{b,1}, \widehat{\mathbf{z}}_{b,1\mid t}),
    \ldots,
    (\mathbf{a}_{b,K_b}, \widehat{\mathbf{z}}_{b,K_b\mid t})
    \right]
    \right\}_{b=1}^{B_t},
    \qquad
    \widehat{\mathbf{z}}_{b,k\mid t}
    =
    W_\theta(\widehat{\mathbf{z}}_{b,k-1\mid t}, \mathbf{a}_{b,k}),
    \quad
    \widehat{\mathbf{z}}_{b,0\mid t}=\mathbf{z}_t ,
\end{equation}
and the final decision is conditioned on both the present belief and the imagined counterfactuals,
\begin{equation}
    \widehat{\mathbf{a}}_t
    =
    \pi_\phi\!\left(
    \mathbf{z}_t,
    \operatorname{Agg}_\psi(\mathcal{C}_t)
    \right).
\end{equation}
This abstraction captures the essence of CoI: reasoning is a structured set of imagined action--consequence pairs, not a post-hoc verbal rationale.

Recent systems instantiate different points in this design space. MineDreamer provides a visual form of CoI: it imagines stepwise goal images and uses them as visual prompts for low-level control, making intermediate imagination directly actionable \citep{zhou2024minedreamer}. LCDrive moves the idea into compact latent space for end-to-end driving. Its latent CoT interleaves action-proposal tokens with latent world-model prediction tokens, so each candidate maneuver is evaluated through its predicted consequence before the final trajectory is decoded \citep{tan2025lcdrive}. Crucially, the proposal tokens share the action vocabulary of the policy, while the world-model tokens ground the trace in counterfactual dynamics; reasoning and control are therefore aligned in representation rather than connected by an after-the-fact explanation. FutureX adds the systems-level principle that deliberation should be allocated, not blindly spent. Its Auto-think Switch routes routine scenes through an instant policy and activates latent world-model rollout only for difficult scenes, where a summarizer refines the initial trajectory using imagined future latents \citep{lin2025futurex}. More generally, one may view adaptive CoI as learning
\begin{equation}
    (B_t,K_t)=\rho_\omega(\mathbf{z}_t),
\end{equation}
so that reasoning budget is governed by risk, uncertainty, interaction complexity, and latency constraints.

CoI is therefore the convergence of chain-of-thought reasoning and world-model imagination. CoT contributes deliberate intermediate computation; world models contribute grounded counterfactual dynamics; CoI fuses them into spatiotemporal chains of imagined action and consequence. Its training objective should not reward explanations that merely sound plausible, but internal futures that improve decisions. High-quality CoI requires latent future supervision, action--reasoning consistency, trajectory- or reward-level feedback, uncertainty calibration, and selective computation. Its central risk is equally clear: an uncalibrated world model turns reasoning into structured hallucination. The promise of CoI is not that agents will explain their actions better, but that they will think in the same variables through which they must act.

\subsection{Physics-Informed and Constrained Learning}

A complementary direction to purely data-driven world-model learning is to inject physical structure directly into the learning process. The motivation is straightforward: a world model should not only generate futures that look plausible, but also produce trajectories that remain admissible under the regularities of the physical world. These regularities may take the form of differential equations, conservation laws, boundary conditions, object permanence, contact constraints, stability requirements, or known causal mechanisms. In this sense, physics-informed learning provides an inductive bias that narrows the hypothesis space from all statistically likely futures to futures that are also physically coherent.

How physical knowledge is embedded, however, varies substantially in the strength of the guarantee it provides and in where it acts within the model. It is useful to organize existing approaches along three levels of increasing structural commitment: \emph{penalty-based (soft) constraints}, which encode physics as auxiliary loss terms; \emph{architecture-based (hard) constraints}, which build physical laws into the model so that they hold by construction; and \emph{hybrid physics--learning schemes}, which couple a differentiable physics engine with learned components in a dual-pathway design. These levels trade flexibility against the reliability and out-of-distribution behavior of the resulting model.

\paragraph{Penalty-based (soft) constraints.}
The most common and least intrusive approach is to add physical residuals as soft penalties in the training objective. A generic objective can be written as
\begin{equation}
    \mathcal{L}
    =
    \mathcal{L}_{\mathrm{pred}}
    +
    \lambda_{\mathrm{dyn}}\mathcal{L}_{\mathrm{dyn}}
    +
    \lambda_{\mathrm{cons}}\mathcal{L}_{\mathrm{cons}}
    +
    \lambda_{\mathrm{bc}}\mathcal{L}_{\mathrm{bc}},
\end{equation}
where $\mathcal{L}_{\mathrm{pred}}$ measures prediction error in observation or latent space, $\mathcal{L}_{\mathrm{dyn}}$ penalizes violations of known equations of motion, $\mathcal{L}_{\mathrm{cons}}$ enforces conservation or invariance properties, and $\mathcal{L}_{\mathrm{bc}}$ encodes boundary, contact, or feasibility constraints. This formulation is closely related to physics-informed neural networks (PINNs), which use neural function approximators while regularizing them with governing equations and boundary conditions \citep{raissi2019physics}. For world models, the same idea can be applied not only to continuous physical fields, but also to latent state transitions, object trajectories, contact events, and action-conditioned rollouts. Because the physics enters only as a weighted penalty, the constraint is \emph{encouraged rather than guaranteed}: the approach is easy to attach to almost any differentiable model, but the predicted trajectories can still violate the intended law when the penalty is outweighed by the data term, and balancing the multipliers $\lambda$ across terms of differing scale and optimization stiffness is often delicate.

\paragraph{Architecture-based (hard) constraints.}
A stronger form of physics-informed learning builds the constraint into the architecture rather than treating it as an additional loss, so that the relevant physical law is satisfied by construction. Instead of directly predicting arbitrary state updates, Hamiltonian, Lagrangian, symplectic, and port-Hamiltonian neural networks parameterize an energy function or a structured dynamical system from which the evolution law is derived, conserving energy or momentum exactly up to integration error \citep{greydanus2019hamiltonian,cranmer2020lagrangian,zhong2020symplectic}. Similarly, graph-based neural simulators encode locality and relational interaction by representing particles, meshes, or objects as nodes and physical interactions as edges, building translation invariance and pairwise-interaction structure into the model \citep{sanchez2020learning,pfaff2021learning}. Such structure-preserving designs are particularly attractive for long-horizon prediction, because unconstrained rollout errors often accumulate into physically impossible states, whereas encoding the invariant in the architecture reduces energy drift, improves stability, and generalizes better outside the training distribution. The cost is reduced flexibility and a dependence on the assumed structure being correct: a hard constraint that encodes the wrong invariant biases the model in ways a soft penalty would not.

\paragraph{Hybrid physics--learning schemes.}
A third route couples an explicit, differentiable physics engine with learned neural components in a dual-pathway design, combining the strengths of the previous two levels. When part of the underlying physics is known, a differentiable simulator can be embedded inside the world model, allowing gradients to flow through physical state updates, collision handling, or control objectives \citep{de2018end,hu2020difftaichi,freeman2021brax}. Learning then focuses on the unknown or hard-to-model components, such as friction, drag, actuator delay, deformable contact, material parameters, or residual sim-to-real gaps. This hybrid strategy is often more data-efficient than learning the entire transition function from scratch: analytic physics supplies a coarse causal scaffold, while neural components compensate for model mismatch. In embodied settings, this is especially valuable because real interaction data are expensive, and small prediction errors can lead to unsafe or ineffective plans. The boundary between this level and the previous one is not sharp in practice: architecture-based components such as Lagrangian and Hamiltonian networks can serve as the analytic pathway in a hybrid scheme when the governing form is only partially known.

For visual world models, however, physics is informative but incomplete. Raw observations contain lighting, texture, occlusion, camera motion, background clutter, and other factors that are not themselves part of the physical state. Therefore, directly enforcing physical laws in pixel space can be too restrictive. A more suitable strategy is to constrain the latent variables that correspond to action-relevant physical quantities, while leaving appearance-related factors to more flexible generative components. This suggests a layered design: perception modules infer compact physical states from observations; structured dynamics modules evolve these states under actions; rendering or decoding modules map latent states back to observations. In such a design, physical constraints need not explain every pixel, but they should discipline the variables that matter for prediction, intervention, and control.

Physics-informed learning also clarifies an important distinction between visual plausibility and physical correctness. A video prediction model may generate realistic-looking frames while still violating object permanence, momentum, contact consistency, or causal response to action. Conversely, a physically structured model may predict coarse trajectories accurately while producing visually imperfect images. This tension is central to world models: the model must eventually support both high-fidelity perception and reliable physical reasoning. Physics-informed and constrained learning should therefore be viewed not as a replacement for large-scale self-supervision, but as a mechanism for aligning learned representations with the invariants needed for planning and decision making.

Despite its promise, constrained learning introduces its own difficulties. Soft physical penalties can be hard to balance against reconstruction or prediction losses, especially when different terms have different scales or optimization stiffness. Hard constraints may improve extrapolation when the assumed structure is correct, but can bias the model when the prior is incomplete or wrong. Real-world systems also involve partial observability, discontinuous contact, hidden material properties, dissipation, multi-physics coupling, and stochastic external disturbances, all of which make exact physical modeling difficult. As a result, the central question is not whether world models should be purely data-driven or physics-based, but where and how strongly physical structure should be imposed.

The most plausible path forward is therefore hybrid. Future world models will likely learn from broad observational and interaction data, while selectively imposing physical constraints at the levels where reliability matters most: latent state representation, transition dynamics, numerical integration, contact handling, and planning-time feasibility. In this view, physics-informed learning is not merely a technique for improving prediction accuracy; it is a way to make world models more causal, more interpretable, and more trustworthy when deployed in the physical world.

\subsection{Counterfactual Reasoning}

A conventional world model answers a predictive question: given the present, what futures are plausible under different actions? Counterfactual reasoning asks a sharper and more diagnostic question: given the episode that actually occurred, what would have happened in that same episode if one decision had been different? In this sense, counterfactual reasoning turns a world model from a simulator of possible futures into an instrument for causal attribution. The key is not merely to generate another plausible rollout, but to replay the same world while surgically changing only the decision under analysis.

This idea can be formalized using a structural causal model
\begin{equation}
    \mathcal{M}_{\mathrm{SCM}}
    =
    (\mathcal{U},\mathcal{V},\mathcal{F},P_{\mathcal{U}}),
\end{equation}
where $\mathcal{U}$ denotes the space of exogenous background variables, $\mathcal{V}$ the endogenous variables inside the model, $\mathcal{F}$ the mechanisms that determine how variables influence one another, and $P_{\mathcal{U}}$ the distribution over latent circumstances. A counterfactual such as \(Y_x(u)\) means: keep the same background circumstance \(u\), replace the mechanism for \(X\) by the intervention \(X=x\), and then ask what value \(Y\) would take in the modified model $(\mathcal{M}_{\mathrm{SCM}})_x$. After observing factual evidence \(E=e\), the corresponding query is
\begin{equation}
    \mathbb{P}_{\mathcal{M}_{\mathrm{SCM}}}(Y_x \mid E=e)
    =
    \int
    \mathbb{P}_{(\mathcal{M}_{\mathrm{SCM}})_x}(Y\mid U=u)\,
    p_{\mathcal{M}_{\mathrm{SCM}}}(u\mid E=e)\,du .
\end{equation}
The equation is simply the formal version of a three-step procedure: infer what hidden circumstances made the factual episode happen; change one mechanism by intervention; then predict the consequences while holding the inferred background fixed \citep{pearl2009causality,rubin1974estimating}.

This is often described as the abduction--action--prediction pipeline. \textbf{Abduction} reconstructs the latent situation behind the observed trajectory: hidden state, unobserved context, physical parameters, stochastic noise, or the intentions of other agents. \textbf{Action} performs a surgical edit, for example replacing the factual action \(A_t=\mathbf{a}\) with \(do(A_t=\mathbf{a}')\). \textbf{Prediction} rolls the edited model forward under the same abducted circumstances. Thus a counterfactual does not ask what usually happens after action \(\mathbf{a}'\); it asks what would have happened here, in this very episode, had \(\mathbf{a}'\) been chosen. This is why, in general,
\begin{equation}
    \mathbb{P}(Y_{\mathbf{a}'}\mid E=e)
    \neq
    \mathbb{P}(Y\mid A_t=\mathbf{a}',E=e).
\end{equation}
The right-hand side compares against other episodes in which \(\mathbf{a}'\) happened to be taken; those episodes may involve different states, goals, noise, opponents, or environments. The left-hand side instead keeps the factual world fixed and changes only the decision \citep{pearl2009causality,pawlowski2020deep}.

The practical importance of this distinction is the same-world constraint. A model that produces two visually plausible rollouts under two actions may still fail as a counterfactual model if the two rollouts differ in hidden background factors. For example, when evaluating a robot's grasp, changing the grasp direction should change the object's downstream motion, not its mass, identity, lighting, or friction. If these background factors silently change, the model has not isolated the effect of the decision; it has changed the world itself. A genuine counterfactual contrast therefore takes the form
\begin{equation}
    \Delta_Y(\mathbf{a}',\mathbf{a}\mid E=e)
    =
    \mathbb{E}_{\mathcal{M}_{\mathrm{SCM}}}
    \left[
    Y_{\mathbf{a}'}-Y_{\mathbf{a}}
    \mid E=e
    \right],
\end{equation}
where the difference is attributed to the decision because the latent circumstances are held fixed. Such contrasts are central to causal credit assignment, regret analysis, policy debugging, off-policy explanation, and safety evaluation.

This perspective also clarifies why counterfactual world models require more than accurate next-step prediction. Observational prediction can learn that certain events tend to follow others, but it does not by itself determine what would have happened under a different action in the same latent situation. To support counterfactual inference, a model must learn not only temporal regularities but also the causal structure that separates manipulable variables, stable mechanisms, and nuisance context. Interventional or action-labelled data helps distinguish \(P(Y\mid A)\) from \(P(Y\mid do(A))\); paired simulations and counterfactual data augmentation provide direct same-context contrasts; invariant representation objectives encourage the model to factor out stable causal mechanisms from changing background conditions \citep{scholkopf2021toward,pitis2022mocoda,richens2024robust,gupta2024essential}. The abduction step is especially important: without a calibrated posterior over the exogenous factors, a purported counterfactual rollout is only conditional generation with causal language attached.

The limitations are equally fundamental. Counterfactuals are generally not identifiable from observational accuracy alone. Multiple structural models can agree on the observed distribution, and sometimes even on interventional distributions, while disagreeing about counterfactual quantities such as \(Y_x\), because they encode different couplings between the factual and hypothetical worlds. Deep generative parameterizations do not remove this ambiguity. Assumptions such as monotonicity, invertibility, modularity, known mechanisms, or sufficiently rich interventions are structural commitments, not automatic consequences of scale \citep{bareinboim2022pearl,nasr2023counterfactual}. A counterfactual-capable world model should therefore treat counterfactual validity as a semantic constraint, not merely a visual or predictive one. It must preserve the factual evidence, keep exogenous identity fixed, modify only the intervened mechanism, respect causal support, and avoid impossible or out-of-distribution actions. When these conditions cannot be verified, the appropriate output is not a single confident imagined trajectory, but uncertainty, bounds, or sensitivity to the assumed causal structure.

\subsection{Long-Horizon and Hierarchical Planning}

World models make long-horizon planning tractable not by rolling farther, but by changing the unit of planning. A flat planner over primitive actions faces exponential branching, weak credit assignment, and optimizer-amplified model error. A useful world model therefore learns a decision geometry of controllability: which latent regions are reachable, which transitions are reliable, and which events should become subgoals. Under partial observability, this geometry is carried by a recurrent belief and imagined through compact latent dynamics,
\begin{equation}
    b_t
    =
    \mathcal{B}_\theta(b_{t-1},\mathbf{a}_{t-1},\mathbf{o}_t),
    \qquad
    \widehat{\mathbf{z}}_{t+1\mid t}
    \sim
    \widehat p_\theta(\cdot\mid \mathbf{z}_t,\mathbf{a}_t),
    \label{eq:belief-world-model}
\end{equation}
turning planning into counterfactual search over beliefs. This connects Dyna-style simulation with World Models, PlaNet, Dreamer, and MuZero-style value-equivalent planning, where the model is useful insofar as it preserves reward, value, reachability, and action-relevant causal structure \citep{sutton1991dyna,ha2018worldmodels,hafner2019planet,hafner2019dreamer,Schrittwieser2020}.

Hierarchical planning turns this geometry into temporal decomposition. Classical options and MAXQ formalized temporally extended actions and value decompositions; modern world-model agents learn the goal space itself \citep{sutton1999between,dietterich2000maxq}. A high-level policy selects latent subgoals, skills, or bottleneck states, while a low-level policy grounds them through feedback control,
\begin{equation}
    \mathbf{g}_k
    \sim
    \pi_{\mathrm{hi}}(\cdot\mid b_{t_k},\mathbf{m}_{t_k}),
    \qquad
    \mathbf{a}_t
    \sim
    \pi_{\mathrm{lo}}(\cdot\mid b_t,\mathbf{g}_k),
    \quad
    t_k\le t<t_{k+1}.
    \label{eq:hierarchical-policy}
\end{equation}
The gain is not merely a shorter rollout, but a lower-entropy search problem: $H$ primitive actions are replaced by $K\ll H$ abstract commitments when each commitment corresponds to an executable closed-loop skill. Director realizes this idea by planning over latent goals from pixels, while THICK shows how hierarchical world models can expose multi-timescale, interpretable temporal abstractions \citep{hafner2022director,gumbsch2024learning}. A subgoal that cannot be executed is not an abstraction; it is a model-induced fiction.

Memory supplies temporal continuity. Working memory maintains hidden variables, instructions, commitments, and recent failures in the belief state. Episodic memory retrieves concrete trajectories---rare successes, dead ends, and landmark transitions---to seed subgoals, constrain rollouts, or provide non-parametric value evidence. MERLIN, Neural Episodic Control, GEM, and R2I illustrate complementary forms of predictive, episodic, and sequence-level memory for decision-making \citep{wayne2018merlin,pritzel2017neural,hu2021gem,samsami2024mastering}. Memory is therefore not an auxiliary cache, but a constraint on what the planner may regard as possible.

The principal failure mode is hallucinated planning under model misspecification. Because planning optimizes inside the learned model, it actively searches for model errors: a policy may be optimal in imagination yet infeasible or unsafe in the environment. One diagnostic is the horizon-limited value gap,
\begin{equation}
    \Delta_H(\pi_\phi;b_t)
    =
    \left|
    V_{\widehat p_\theta,H}^{\pi_\phi}(b_t)
    -
    V_{P^\star,H}^{\pi_\phi}(b_t)
    \right|,
    \label{eq:model-misspecification-error}
\end{equation}
which can grow as rollouts leave the data-supported region. Hierarchy shortens rollouts and enables replanning at subgoal boundaries, but it amplifies hallucination when abstract transitions are not calibrated against real executions. Robust agents therefore treat imagination as a proposal mechanism, not an oracle: rollout length, value expansion, uncertainty, episodic retrieval, feasibility checks, and online correction must anchor imagined futures to experience \citep{talvitie2017self,%
jafferjee2020hallucinating}. World models do not eliminate planning risk; they relocate it into representation, memory, and calibration.

\section{Application Domains}

\subsection{Robotics and Embodied AI}
{
In robotic systems, world models serve multiple complementary roles: they can act as predictive simulators for policy evaluation and planning, representation learners that capture task-relevant dynamics from large-scale observations, and increasingly as data engines that generate synthetic experiences for downstream policy training.
These capabilities are particularly important in embodied settings, where collecting real-world interaction data is expensive and where agents must generalize across diverse tasks, environments, objects, and embodiments.

Recent advances in generative modeling have substantially broadened the scope and effectiveness of robotic world models. Inspired by the success of large-scale video generation models~\citep{rombach2022high}, a growing line of work has moved beyond compact latent-state prediction toward video-based world modeling, where future visual observations are explicitly generated and used for decision making. Such models have been applied to robotic manipulation, navigation, policy evaluation, reinforcement learning, and synthetic data generation, leveraging the rich spatiotemporal knowledge encoded in large-scale video datasets. Building upon this trend, recent World Action Models (WAMs) jointly model future observations and robot actions within a unified generative framework~\citep{hu2024video,ye2026dreamzero}.
Rather than treating planning and control as separate stages, WAMs learn the joint distribution of future world states and actions, enabling action prediction to directly benefit from learned environment dynamics.
Representative works have shown that combining world modeling objectives with action prediction improves sample efficiency, robustness, cross-embodiment transfer, and generalization to novel tasks and environments.
More importantly, by inheriting visual and physical priors from internet-scale video pretraining, WAMs suggest a promising path toward scalable embodied intelligence, where advances in generative world modeling can directly translate into improved robotic capabilities.
As a result, world models are increasingly viewed not merely as planning modules, but as a unified framework that integrates perception, prediction, simulation, and action generation for future embodied agents.

Beyond the unified perception-prediction-action formulation embodied by WAMs, the functional contribution of world models to robotic systems can be more sharply delineated along three complementary axes, which differ in \emph{when}, and at what computational budget, predicted futures are consumed within the embodied agent's pipeline.

\paragraph{Data Engines.}
In an offline regime, a learned world model can be queried repeatedly, decoupled from real-time control, to synthesize large volumes of visually and physically plausible interaction data that supplement or substitute for costly teleoperated demonstrations. Recent compositional and embodiment-aware world models illustrate this role concretely: RoboDreamer factorizes language-conditioned video generation into reusable primitives in order to synthesize plans for unseen combinations of objects and actions~\citep{zhou2024robodreamer}, while UniSim orchestrates heterogeneous image, robotics, and navigation data into a universal interactive simulator from which both high-level and low-level policies can be trained purely in imagination prior to zero-shot real-world transfer~\citep{yang2024unisim}.

\paragraph{Environment Simulators.}
Rather than generating training data, the model is queried at evaluation to score or rank candidate policies, action sequences, or hyperparameters by rolling out their predicted consequences, thereby reducing the cost and risk of physical trial-and-error prior to deployment.
This idea dates back to Ha and Schmidhuber's original world model, in which a policy is trained and evaluated entirely inside a learned ``dream'' environment before being transferred back to the real one~\citep{ha2018worldmodels}.
In robotic and driving settings, MILE jointly learns a world model and a driving policy, allowing complex manoeuvres to be planned and assessed purely through imagination before being verified in the CARLA simulator~\citep{hu2022mile}, while ReSim explicitly targets the reliability of such in silico evaluation by addressing the tendency of driving world models trained only on safe expert data to fail on hazardous or non-expert behaviors~\citep{yang2025resim}.
A complementary, non-generative strand pursues the same functional goal through high-fidelity rendered simulation rather than learned dynamics: SIMPLER demonstrates that carefully constructed simulated environments can serve as a scalable and reproducible proxy for real-world manipulation policy evaluation, correlating strongly with real-robot performance while avoiding the cost and reproducibility issues of physical testing~\citep{li2024simpler}.

\paragraph{Action Planners.}
A third role embeds the world model directly within the control loop itself, where it is queried online and at low latency to infer plausible near-future states conditioned on candidate actions, supporting receding-horizon or imagination-based planning under the real-time constraints of physical deployment. DayDreamer offers an early demonstration of this regime, learning latent dynamics and acting directly on physical quadrupeds and manipulators without recourse to an external simulator~\citep{wu2022daydreamer}.

\paragraph{Embodiment World Models.}
The roles surveyed above all concern modeling the dynamics of the \emph{external environment}, conditioned on the agent's actions.
An orthogonal and comparatively underexplored line of work instead targets the agent's own \emph{embodiment}: rather than predicting what will happen in the world given an action, it asks what behaviors the agent's own body is capable of producing in order to reach a goal or satisfy a control objective.
The recently proposed Behavior Foundation Model (BFM) exemplifies this direction for humanoid whole-body control~\citep{zeng2025bfm}.
Motivated by the observation that existing whole-body control frameworks remain largely task-specific and rely on labor-intensive reward engineering, BFM is pretrained on large-scale behavioral data and combines a masked online distillation framework with a conditional variational autoencoder to model a broad, reusable distribution over feasible humanoid behaviors, which can subsequently be steered across diverse control modes or rapidly adapted to novel behaviors without retraining from scratch.
Whereas a world model encodes a generative prior over how the external scene evolves, BFM and related behavior(al) foundation models instead encode a generative prior over the embodiment's own morphology-constrained skill manifold, analogous to an internal body schema rather than an external scene model. 

We contend that these two families of generative priors, one over the world and one over the body, are complementary rather than competing, and that their integration constitutes a promising and largely open direction for general-purpose embodied agents that must simultaneously anticipate environmental consequences and know what their own body can do.

\textbf{Driving Simulators.}
A family of driving world models has emerged that explicitly casts the evolution of the driving scene as a conditional generation problem over future sensory observations, given past observations and ego-vehicle actions.
GAIA-1 generates realistic future driving videos jointly conditioned on video, text, and action inputs, and demonstrates that large-scale video-action pretraining endows the model with an implicit understanding of traffic conventions and agent dynamics, allowing it to additionally serve as a neural simulator for generating training and validation data~\citep{hu2023gaia1}.
DriveDreamer constructs a world model entirely from real-world driving scenarios, harnessing diffusion models to represent structured traffic constraints and supporting both controllable video generation and future driving-action prediction~\citep{wang2023drivedreamer}.
Vista further extends this paradigm toward generalizable, high-fidelity prediction with versatile controllability across heterogeneous action representations, explicitly targeting the limited generalization to unseen environments and the loss of safety-critical detail that affect earlier driving world models~\citep{gao2024vista}. Across these systems, the temporal dimension is not incidental but constitutive: autonomous driving requires reasoning simultaneously over short-horizon reactive prediction, e.g., imminent collision avoidance, and longer-horizon trajectory forecasting for route-level decision-making, and driving world models address this by amortizing both through autoregressive rollouts in a learned latent or pixel space. In this view, scene prediction and trajectory forecasting are not independent submodules but two temporally coupled instantiations of the same underlying predictive substrate, recovering, within an explicitly generative and temporally extended framework, the joint perception-planning objective first foregrounded by UniAD.
}

\subsection{Scientific Discovery}

\begin{figure}[t]
    \centering
    \includegraphics[width=\linewidth]{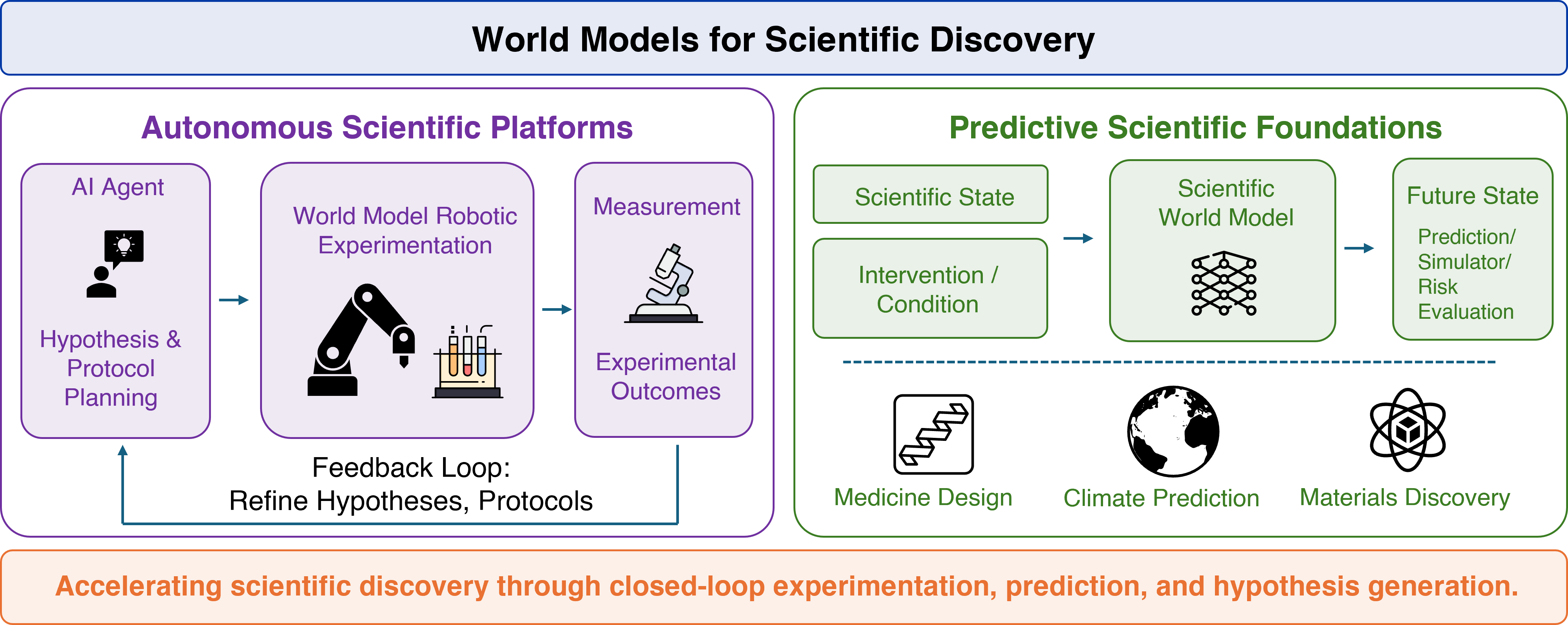}
    \caption{World model application on scientific discovery}
    \label{fig:future}
\end{figure}

World models have emerged as a foundational paradigm underpinning AI for Science (AI4S), enabling the comprehensive intelligence-driven upgrading and optimization of scientific discovery platforms \citep{wang2026towards}. By providing a systematic digital framework to model the underlying dynamics, physical laws, and operational principles of the objective world, these models contribute to scientific discovery in two highly complementary ways: by empowering autonomous experimental systems and by serving as predictive foundations for complex scientific phenomena.

First, world models integrate with AI agents, robotic instruments, and experimental feedback to support AI-driven autonomous laboratories \citep{szymanski2023autonomous,wei2025ai,wu2025ai}. Spanning critical modalities—such as complex robotic manipulation and dynamic instrument state variations—they operate in tandem with cognitive large language models through a synergistic closed-loop mechanism. Massive streams of experimental data generated from real-world platform deployments are continuously recycled to iteratively train and refine the world model. This forms a bidirectional, application-driven paradigm where physical execution enhances model accuracy, while the model's simulations empower practical execution. In wet-lab settings, this integration significantly accelerates the operational efficiency of embodied systems, enabling predictive \textit{in silico} rollouts to anticipate procedural and safety risks before execution, thereby mitigating hazards and stabilizing workflows \citep{lewis2026acht}.

Second, world models serve as predictive foundations by learning how scientific states evolve under different conditions or interventions. They extend the embodied AI concept of “state + action $\rightarrow$ next state” to scientific domains, where the goal becomes modeling “scientific state + intervention / condition $\rightarrow$ future state / response.” This role is increasingly supported by scientific foundation models and discovery platforms, such as Intern-S1 \citep{bai2025intern}, Intern-S1-Pro \citep{zou2026intern}, and Intern-Discovery, which provide reasoning, multimodal understanding, tool use, and workflow infrastructure. Concurrently, domain-specific predictive world models are emerging across various disciplines. In weather and climate, models like GraphCast \citep{lam2023learning}, GenCast \citep{price2025probabilistic}, FengWu \citep{chen2023fengwu}, and FengWu-W2S \citep{ling2024fengwu} treat the Earth system as a high-dimensional dynamical state and learn its future evolution. Similar advancements are appearing in medicine and life sciences, where the Medical World Model \citep{yang2025medical} simulates treatment-conditioned tumor evolution, and Delphi-2M \citep{shmatko2025learning} models long-term disease progression.

Ultimately, world models drive frontier scientific discovery by mapping natural laws that lie beyond existing human cognitive boundaries. Through physics-consistent forward deduction and counterfactual analysis, they isolate latent research pathways and inspire novel hypotheses. This capability not only reduces the cost of real-world experiments and accelerates closed-loop iterations but also allows researchers to explore complex processes that are difficult to observe directly, securing the long-term evolution and breakthroughs of the AI4S ecosystem.

\section{Open Challenges and Bottlenecks}

World models represent the definitive frontier for Physical AI, yet their trajectory toward true autonomy is fundamentally bottlenecked by a systemic schism between perceptual synthesis and physical execution. While current paradigms scale effortlessly on passive, internet-scale visual data, they remain profoundly brittle when confronting the interactive, safety-critical constraints of the real world. This section deconstructs the open challenges facing world models through a holistic lens: from the structural data asymmetry that leaves models "visually rich but physically blind," to the divergence between superficial fidelity and actionable precision, and the compounding drifts of open-loop rollouts. We argue that overcoming these bottlenecks requires a fundamental paradigm shift—moving beyond world models as mere generative renderers toward grounded physical simulators and closed-loop interactive planners. This transition demands not only physics-aware inductive biases and continuous ambient data harvesting, but also a radical reimagining of evaluation frameworks, triadic safety guarantees, and decentralized governance models capable of reconciling collective intelligence with localized physical sovereignty.

\subsection{Data Asymmetry}

Under the functional taxonomy~\citep{worldlabs2026taxonomy} of renderers, simulators, and planners, world models face a clear data asymmetry. Renderers can scale on internet-scale image and video data, driving rapid progress in image/video generation~\citep{kong2024hunyuanvideo,gao2025seedance,seedance2026seedance2,happyhorse2026}. However, such data is mostly passive and visual: it captures how the world looks, but provides limited supervision about actions, physical properties, contact dynamics, or task outcomes.

By contrast, simulators and planners depend on more structured and actionable data. For simulators, the bottleneck lies in simulation-ready 3D assets. Although 3D generation~\citep{zhao2025hunyuan3d,hunyuan3d2025hunyuan3d,li2025triposg} is rapidly increasing the amount of visual assets, most generated objects remain render-ready rather than physically executable, often lacking scale, material, collision geometry, mass, friction, articulation, and affordance. Recent research efforts such as PhysX-Anything~\citep{cao2026physx} and PhysX-Omni~\citep{cao2026phyomni}, together with industrial platforms such as Lightwheel~\citep{lightwheel2026} and Genesis World~\citep{genesisworld2026}, directly target this gap, but realism, non-rigid dynamics, long-tail interactions, and scalability remain open challenges. For planners, the bottleneck further shifts to action-conditioned interaction data, such as robot demonstrations, state-action trajectories, and task outcomes. Recent models such as Cosmos 3~\citep{nvidia2026cosmos3} explicitly incorporate action sequences into omnimodal world modeling, highlighting the importance of action-conditioned data for Physical AI; however, such data remains far less scalable and transparent than internet video. Synthetic data and simulation~\citep{chang2017matterport3d,savva2019habitat,puig2024habitat,dosovitskiy2017carla,gan2020threedworld,gao2026nvidia} can partially relieve this bottleneck, but introduce visual, geometric, physical, behavioral, and task-level distribution shifts. Together, these limitations form an embodied AI data pyramid: abundant passive video at the base, fewer simulation-ready assets and interaction trajectories in the middle, and scarce but highly valuable tactile and force feedback at the top. 

Confronting these systemic limitations requires a fundamental paradigm shift in how physical intelligence ingests the world. Recognizing that existing data acquisition methods are inherently constrained by scalability boundaries and distribution shifts, the future architecture of world model data acquisition must transition toward a ubiquitous, non-intrusive, and continuous framework. First, data harvesting must be ubiquitous, utilizing widespread and heterogeneous collection mechanisms across diverse physical settings to thoroughly capture long-tail environments. Second, the ingestion pipeline must be completely non-intrusive, operating seamlessly in the background to ensure data gathering never becomes a burden to human participants or a disruption to daily workflows; ideally, this ambient collection should actively enhance operational efficiency through synergistic human-AI collaboration. Finally, data collection must be continuous, establishing an unbroken, perpetual loop of real-world interaction feedback that dynamically and iteratively updates the underlying model in real time.

\subsection{Fidelity vs.\  Precision}

A key challenge for world models is the mismatch between perceptual fidelity and physical precision. Recent benchmarks and studies provide empirical evidence for this mismatch. Physical-reasoning evaluations~\citep{kang2024far,motamed2025generative,zhang2025morpheus,meng2024towards,gu2025phyworldbench,zheng2025vbench} show that video generation models still struggle with physical laws, conservation constraints, and out-of-distribution generalization, even as their visual quality improves. Embodied benchmarks~\citep{jiang2026robowm,shang2026worldarena} further reveal that visually realistic predictions do not necessarily translate into physically plausible or executable robot behaviors, exposing a perception--functionality gap for world models used in manipulation and planning.

Recent works~\citep{zhao2026phyworld,chen2026abot,wang2025physcorr,lin2026mmphysvideo} use physical rewards, preference pairs, DPO-style alignment, or structured physical conditions to improve physical plausibility while preserving visual quality. However, current rewards and preference signals are still incomplete proxies for real physical correctness, especially for fine-grained contact, spatial consistency, and long-tail interactions. Thus, the gap between perceptual fidelity and physical precision remains an important open challenge.

Confronting the mismatch between perceptual fidelity and physical precision requires a shift from fidelity-first generation to physical-precision-centered world modeling. First, pretraining should incorporate scalable physics-rich supervision that connects visual outcomes with latent physical states, such as 3D geometry, object motion, contact, force, material properties, and task outcomes. Second, post-training should move beyond visual preference alignment by introducing physical rewards and evaluations that measure constraint satisfaction, counterfactual validity, long-horizon stability, and task success. Finally, world models should be calibrated through closed-loop feedback from real or high-confidence physical environments, where predictions are tested against actual outcomes and refined iteratively. The goal is not merely to generate realistic futures, but to make world-model predictions actionable, verifiable, and reliable under physical intervention.

\subsection{Compounding Prediction Errors}
{Even when a world model is locally accurate, long-horizon deployment can still fail because small one-step errors are recursively fed back into future predictions. In autoregressive rollouts, each predicted observation, latent state, or belief state becomes part of the context for the next step. Minor inaccuracies in geometry, velocity, contact timing, object permanence, or hidden-state estimation can therefore accumulate, pushing imagined trajectories away from the data manifold and eventually away from physically realizable behavior. This phenomenon has long been recognized as a central source of model bias in model-based reinforcement learning and planning \citep{talvitie2017selfcorrecting,lambert2022compounding}. Analogous failure modes are also well documented in video prediction and autoregressive video generation, where exposure bias and temporal drift degrade long-horizon rollouts \citep{nair2019hvf,chen2024diffusionforcing,huang2025selfforcing}. It is also a decision-making problem rather than merely a prediction problem: once a planner begins to optimize against an imperfect model, it may exploit model errors and assign high value to trajectories that would never succeed in the real environment, a failure mode highlighted in Dyna-style planning as ``hallucinating value'' \citep{jafferjee2020hallucinating}. The challenge becomes especially acute for visually rich and contact-heavy domains, where partial observability, branching futures, and sharp discontinuities in dynamics make long open-loop rollout intrinsically brittle.
}

{Several mitigation strategies have proved useful, but none eliminates the problem. A first line of attack is to predict in compact latent state spaces rather than raw pixels, reducing irrelevant perceptual noise and forcing the model to concentrate on decision-relevant dynamics, as in PlaNet and the Dreamer family \citep{hafner2019planet,hafner2019dreamer,hafner2021dreamerv2,hafner2023dreamerv3}. A second line is to make planning uncertainty-aware through probabilistic dynamics models, ensembles, and conservative rollout horizons, so that action selection depends less on a single brittle imagined future and more on calibrated uncertainty over possible futures \citep{chua2018pets,janner2019mbpo,hansen2022tdmpc,hansen2023tdmpc2}. A third line is to improve temporal consistency through self-correcting objectives, hierarchical abstraction, and multiscale planning, which periodically re-anchor long-horizon reasoning to observed evidence instead of trusting a single open-loop rollout indefinitely \citep{talvitie2017selfcorrecting,hafner2022director,gumbsch2024learning}. Taken together, these methods suggest that the key objective is not simply lower one-step prediction loss, but world models whose errors remain bounded, calibrated, and decision-compatible over the horizons that real planning requires.
}

\subsection{Sim-to-Real Transfer}
{
A fundamental challenge in deploying world models in robotics is the gap between learned simulations and real-world physics. While world models aim to approximate environment dynamics from data, even small modeling errors can accumulate rapidly during long-horizon rollouts, leading to compounding distribution shift and unreliable predictions in the real world. This issue is particularly severe in robotics, where contact-rich dynamics, partial observability, and high-dimensional action spaces make accurate long-term prediction extremely difficult.

To mitigate this sim-to-real gap, a variety of strategies have been explored. A classical approach is domain randomization, which improves robustness by exposing policies or dynamics models to diverse simulated variations during training, thereby encouraging invariance to visual and physical perturbations~\citep{tobin2017domain,peng2018sim}. Another line of work focuses on system identification, where unknown physical parameters are inferred from limited real-world observations and used to calibrate simulators or learned dynamics models~\citep{nagabandi2018learning}. More recently, online adaptation methods have been introduced, enabling world models to continuously update their dynamics using real-world feedback, thereby reducing prediction errors under distribution shift~\citep{hafner2019dreamer,nagabandi2018neural}.

In the context of video-based world models and World Action Models (WAMs), the sim-to-real challenge becomes even more pronounced, as these models operate in high-dimensional pixel space and rely heavily on learned visual dynamics priors from large-scale pretraining. While such models benefit from strong generalization across diverse internet-scale videos, transferring these priors to precise robotic control remains non-trivial due to differences in physics, embodiment, and action semantics. 
Moreover, relying solely on visual observations is often insufficient for robotics, as videos lack explicit information about underlying physical states, proprioceptive feedback, and fine-grained action dynamics. Incorporating additional modalities, such as proprioception, tactile sensing, force feedback, and structured robot states, can provide complementary constraints on the environment dynamics and help reduce the sim-to-real gap.
Addressing this gap requires better integration of physics-aware inductive biases, structured state representations, and adaptive inference-time correction mechanisms, which remain open challenges for scalable deployment of world models in real-world robotics.
}

\subsection{Evaluation and Benchmarks}

The evaluation landscape for world models remains fragmented. Since world models span visual prediction, physical simulation, action planning, and downstream decision-making, different communities often rely on different evaluation metrics and benchmarks. For predictive and generative models, commonly used metrics include paired reconstruction errors such as MSE, as well as distributional or perceptual metrics such as FID and FVD, which measure the quality of generated images or videos relative to real data. For control and planning tasks, evaluation is usually based on task-level performance, such as reward, episode return, success rate, or task completion. Generalization is often evaluated under OOD or zero-shot settings, where the same prediction or task metrics are measured on unseen environments, objects, tasks, or dynamics. In safety-critical settings such as robotics and autonomous driving, reliability-oriented metrics such as uncertainty calibration are also important, as agents must know when their predictions or plans are unreliable for safe trajectory rollouts and risk-aware planning.

At the benchmark level, these evaluation matrix are used in a range of representative benchmarks. For predictive and generative video models, benchmark suites such as VBench~\citep{huang2023vbench} and VBench-2.0~\citep{zheng2025vbench} evaluate video quality, temporal consistency, prompt alignment and intrinsic faithfulness, while human-evaluation protocols such as Cosmos-HUE, introduced with Cosmos 3~\citep{nvidia2025cosmos}, assess generative and omnimodal world-model outputs. For 3D/4D world generation, WorldScore~\citep{duan2025worldscore} and 4DWorldBench~\citep{lu20264dworldbench} extend evaluation beyond 2D video quality to controllability, dynamics, and spatiotemporal consistency. For physical consistency and world-simulator fidelity, physics-oriented benchmarks such as PhyGenBench~\citep{meng2024towards}, WorldModelBench~\citep{li2026worldmodelbench}, and PhyWorldBench~\citep{gu2025phyworldbench} assess whether generated videos respect physical commonsense and physical realism, while WorldSimBench~\citep{qin2024worldsimbench} further evaluates video generation models as world simulators through both perceptual and action-level evaluation. For action-conditioned and downstream functional evaluation, Atari / ALE~\citep{bellemare2013arcade} and DM Control~\citep{tassa2018deepmind} are commonly used for model-based reinforcement learning and continuous control; Habitat~\citep{savva2019habitat} and CARLA~\citep{dosovitskiy2017carla} evaluate embodied navigation and autonomous driving in simulation; RoboArena~\citep{atreya2025roboarena} extends evaluation to real-world robot policies; EWMBench~\citep{yue2025ewmbench} evaluates embodied world models in terms of scene consistency, motion correctness, and semantic alignment; and WorldArena~\citep{shang2026worldarena} evaluates the functional utility of embodied world models. For interactive world-model evaluation, recent benchmarks such as WBench~\citep{ying2026wbench} and WorldMark~\citep{xu2026worldmark} assess whether models can follow multi-turn interactions, maintain control consistency, and preserve physical plausibility. Complementarily, long-horizon state reasoning benchmarks such as MBench~\citep{zhang2026mbench}, WorldPrediction~\citep{chen2025worldprediction}, and WorldReasonBench~\citep{wu2026worldreasonbench} evaluate whether world models can maintain persistent states, support procedural planning, and generate future videos with physically, socially, logically, and informationally consistent state evolution.

Recent CoW-Bench further frames world-model evaluation around a ``Trinity of Consistency''---modal, spatial, and temporal consistency---testing whether generative models can preserve cross-modal alignment, 3D-aware spatial structure, object permanence, and physically plausible temporal evolution under complex multimodal scenarios~\citep{wei2026trinityconsistencydefiningprinciple}.
Crucially, existing benchmarks often overlook inference efficiency, yet real-world physical interaction demands closed-loop control frequencies that cannot tolerate high-latency rollouts. Sub-second generative speed is not merely a computational metric but a foundational safety requirement, as inference lag directly triggers control instability in highly dynamic environments.

Ultimately, the current evaluation paradigm faces a fundamental bottleneck: the lack of physical grounding. Virtual simulators operate on simplified physics engines, allowing world models to exploit non-causal shortcuts that fail in reality. True validation requires real-world deployment and continuous physical interaction to confront unmodeled stochastic dynamics. However, shifting to physical platforms introduces a critical trade-off between realism and experimental rigor. The core open challenge lies in benchmarking standardization and fairness. Unlike reproducible software containers, physical evaluation suffers from hardware degradation, inconsistent initial states, and extreme replication costs. Developing protocols that can normalize heterogeneous robotic embodiments and isolate hardware noise from model capability remains the critical next frontier.

\subsection{Safety, Transparency, and Sustainability}

\paragraph{Safe Exploration and Triadic Interaction Safety.}
The deployment of embodied AI and autonomous agents introduces significant physical risks, as real-time decision-making in unconstrained environments poses far greater hazards than isolated data analysis. Because the residual gap between simulation and reality can never be fully closed, untested exploration in real-world environments remains inherently dangerous~\citep{billard2025roadmap}. Physical world models mitigate this risk by acting as learned internal simulators, allowing agent policies to be rigorously optimized within imagined rollouts prior to deployment. During inference, these models provide continuous state representations and multi-step planning to forecast action consequences. To guarantee safety when interacting with heavy or high-throughput machinery, constraint methods and strict uncertainty thresholds must be embedded into the rollout engine to halt execution whenever predictions extrapolate into unsafe state regions. Crucially, because environments, embodiments, and tasks evolve over an agent's operational lifespan, safe exploration must function as a persistent capability coupled with lifelong learning. 

Furthermore, this safety paradigm scales non-linearly within a dynamic, triadic ecosystem encompassing human-robot alignment, environmental feedback co-evolution, and heterogeneous multi-agent synchronization. First, regarding human-robot interaction, safety transcends mere reactive obstacle avoidance to become a game-theoretic cognitive alignment; because human counter-actions shift dynamically based on a robot's trajectory, world models must integrate theory-of-mind frameworks into their latent rollouts to proactively adapt to implicit human intent. Second, in machine-environment loops, world models must exhibit causal bi-directionality—predicting not only how an agent alters its surroundings, but how non-stationary environmental feedback ({\it e.g.}, structural fatigue or thermal deformation) retroactively destabilizes control guarantees. Third, when heterogeneous agents with distinct morphologies cohabitate, the primary risk shifts from individual controller failure to systemic resonance, where asymmetric misalignments between separate world models can cascade into distributed collisions.

\paragraph{Transparency, Explainability, and Verifiable Control.}
For world models to be deployed ethically and reliably, explainability is essential for building operator trust and attributing accountability. Unlike traditional black-box end-to-end policies, world models offer a more structured path to transparency because they rely on an explicit representation of environmental dynamics. However, because deep neural networks are inherently uninterpretable, translating these dynamics into verifiable predictions remains an open engineering challenge. Relying solely on engineered interpretability layers to maintain legible latent dimensions for runtime invariant verification (such as velocity boundaries or energy conservation) is insufficient. Instead, these learned components must be systematically integrated with control theory or formal verification frameworks to achieve provable stability and constraint satisfaction, ensuring predictions inherit rigorous mathematical guarantees rather than functioning as unconstrained extrapolations. 

Furthermore, technical explainability alone does not eliminate systemic risk once human operators are integrated into the loop. The highly realistic, simulated outputs of world models can induce cognitive vulnerabilities like automation bias—the documented tendency for operators to over-rely on authoritative model predictions even when they are flawed or out-of-distribution~\citep{parasuraman1997humans}. Because this vulnerability stems from how explanations are cognitively consumed rather than the accuracy of the underlying model, it cannot be resolved by merely improving interpretability. Mitigating automation bias requires user-facing interfaces to explicitly visualize epistemic uncertainty—such as flagging rollouts that deviate from validated training distributions—to ensure human trust remains accurately calibrated to the system's objective reliability.

\paragraph{Ethical Deployment and Privacy-Preserving Governance.}
Ethical deployment constitutes a distinct imperative concerning the legitimacy of acting on a world model's human-centric predictions and data. This introduces three central obligations. First, when predicting human behavior to plan robotic actions, world models must not rely on bias-based profiling~\citep{billard2025roadmap}—a constraint that is exceptionally difficult to enforce because these predictions feed directly into physical actions rather than rejectable informational recommendations. Second, data collection faces strict privacy boundaries: human subjects in physical interaction datasets must remain unidentifiable, no sensitive attributes should be inferable, and recordings must not be repurposed beyond their original scope~\citep{billard2025roadmap}. While necessary, these constraints severely restrict the construction of large-scale human-interaction datasets. Third, while transparency is a precondition for assigning liability during accidents, this requirement sits in direct tension with the commercial prevalence of closed-source perception and control stacks that resist independent auditing~\citep{billard2025roadmap}, leaving accountability as an open governance challenge.

To resolve these cross-institutional data aggregation barriers and navigate commercial confidentiality, collaborative frameworks must shift toward Federated World Models.
Under this paradigm, localized data silos remain untouched, and participating entities exclusively exchange latent gradients or network weights. To safeguard these architectures against gradient inversion attacks, they must integrate secure multi-party computation or leverage hardware-enforced trusted execution environments (TEEs) for encrypted joint training. Crucially, to accommodate the non-identical and independently distributed (Non-IID) nature of diverse institutional environments, these frameworks must decouple invariant physical commonsense (learned globally and universally shared to anchor collective intelligence) from highly variant, domain-specific latent representations (retained locally), thereby successfully balancing collective intelligence with localized data sovereignty.

However, a critical caveat remains: current optimizations focus almost exclusively on the operational-phase energy costs of training and inference, completely omitting the hardware's comprehensive life-cycle footprint. A complete sustainability assessment must incorporate a life-cycle management framework that accounts for material extraction, manufacturing, and end-of-life disposal~\citep{billard2025designing}. Upstream environmental costs—including rare-earth mining, polymer synthesis for soft actuators, and semiconductor fabrication—are largely ignored in contemporary AI benchmarks. As demonstrated in sustainable computing literature, embodied carbon and operational carbon often exist in a trade-off state; optimizing one in isolation can inadvertently amplify the other~\citep{lee2025view}. Consequently, efficiency claims based solely on compute metrics remain fundamentally incomplete and must be interpreted with caution.

\section{Roadmap}\label{sec:roadmap}

\begin{figure}[t]
    \centering
    \includegraphics[width=\linewidth]{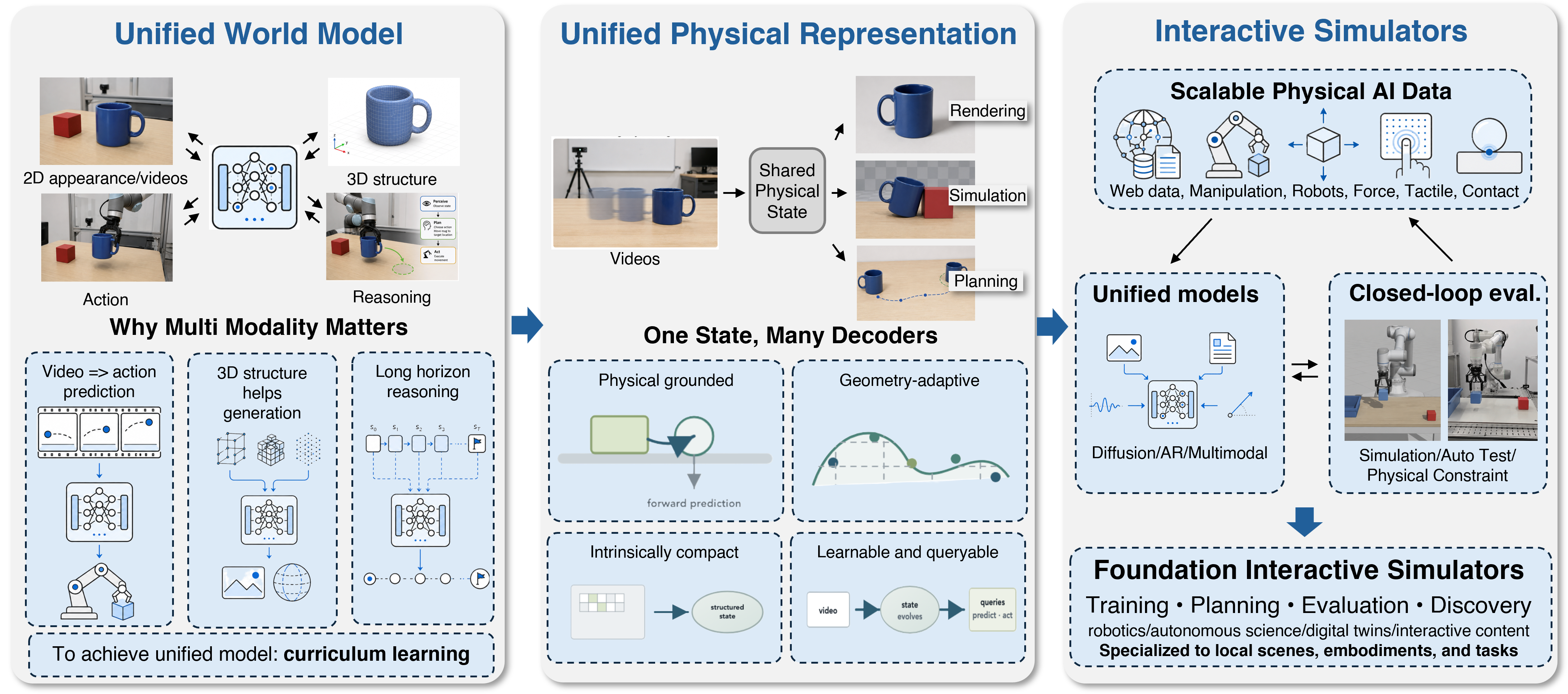}
    \caption{A staged roadmap for developing next-generation world models.}
    \label{fig:roadmap}
\end{figure}

Building on the previous discussion, developing a robust future physical world model requires the three progressive stages shown in Figure~\ref{fig:roadmap}. 
First, the world model must make use of multiple, unified modalities, integrating various asynchronously sampled signals to ensure that the models can work in most physical environments. 
Second, these diverse modalities must be distilled into a unified physical representation, i.e., a single, highly compressed internal state from which rendering, simulation, and planning can all be directly decoded for various downstream tasks. 
Third, by scaling these representations with advanced architectures and physically grounded data, the field can realize foundation-scale interactive simulators. In this ultimate stage, world models transition into closed-loop, reusable environments where agents can safely explore, evaluate, and refine their actions prior to real-world execution.

\subsection{Towards Unified Multimodal World Models}

Unified multimodality is central to future world models because accurate prediction requires appearance, spatial structure, state, action, and long-horizon reasoning to be represented together.
Video captures motion, 3D structure supports view-consistent memory, action-state signals explain controllable change, and semantic reasoning links local predictions to long-term goals.
A unified world model should therefore do more than merge modalities: it should correlate these signals with how the world changes. There are several reasons why unified multimodality is critical for this problem.

First, long-horizon reasoning requires multimodal information.
Although a video diffusion model can generate plausible futures, many tasks, such as manipulation, require long-term future planning, which a simple video diffusion model may struggle to achieve with limited computing resources.
The model therefore needs search, a policy, or a value signal to choose among generated branches.
Diffusion-based control systems often pair generation with tree search or action policies~\citep{huang2025forgetree,chi2023diffusionpolicy}. With additional modalities, the model has greater potential to scale for long-term perception, prediction, and action selection.

Second, some modalities, such as 3D, are helpful complements and improve generalization when data are scarce.
Robot and driving datasets rarely cover every camera view, object pose, or spatial configuration an agent may encounter.
Sparse-view reconstruction and geometry-grounded video generation can synthesize missing views, poses, and motion-consistent examples for training or testing controllable models~\citep{yu2020pixelnerf,liu2023zero123,kang2026geonvs}.
This makes multimodality a practical route to better generalization, because no single modality covers rare spatial cases well.

Third, action and state signals turn video prediction from passive forecasting into embodied prediction.
Visual observations alone do not determine which grasp will succeed, because the same scene can lead to different futures after a push, a grasp, or no action.
Recent video-action world models pair visual context with explicit action and state inputs~\citep{zhou2026tau0wm,alradi2026aeroworld}.
This suggests that visual scale alone is not enough; the model must also encode the variables that make change controllable.

A practical path is to treat unification as a curriculum rather than a single pooled token stream.
Training can begin with static representation learning, where image objectives such as JEPA build semantic features~\citep{assran2023ijepa}.
It can then add video, state, action, and embodied data, while compact latent actions separate controllable change from background appearance~\citep{bruce2024genie,bi2025motus}.

\subsection{Towards a Unified Physical Representation}

As stated before, information compression is key to achieving a unified world model. To compress high-dimensional sensory observations into a shared internal state that preserves the physical structure of the world, the choice of representation is more important than the choice of any individual renderer, simulator, or planner. Most current systems instead maintain three separate definitions of the world: appearance-centric primitives for rendering, such as radiance fields or Gaussian primitives~\citep{mildenhall2020nerf,kerbl20233dgaussians}; meshes or particles for simulation; and occupancy grids or object slots for planning. Translating among them is lossy and ad hoc. A world model that must jointly support perception, simulation, and control therefore raises a prior question that these decoders leave unanswered: what single internal state can all three be decoded from?

The most promising long-term direction is not just a better renderer, simulator, or planner in isolation, but a \emph{shared physical representation} from which each is recovered as a decoding operation. Such a representation should have the following properties. First, it should be physically grounded and simulation-ready, so that the same state carries the dynamical and contact structure required for forward prediction rather than only surface appearance. Second, it should be geometry-adaptive, accommodating the heterogeneous and irregular geometry of real scenes without imposing a regular grid or a fixed template. Furthermore, it should be intrinsically compact, encoding a structured state rather than a dense field, so that capacity is spent on slowly varying, decision-relevant physical factors instead of pixel-level details. Under such a representation, geometry, motion, material properties, appearance, semantic identity, uncertainty, and interaction state would be jointly encoded into one persistent state, and rendering, simulation, and planning would cease to require three separate world definitions: they would become different decoding operations on the same compressed physical state---decoded into pixels or splatting primitives for rendering, into deformation, stress, and contact variables for simulation, or into object--part--affordance structure for action planning.

This ``one state, many decoders'' principle already has partial precedents: PhysGaussian~\citep{xie2024physgaussian}, for instance, drives both physical simulation and rendering from a single set of Gaussian kernels. The broader challenge is to design a unified representation that fits all requirements: 1) it is appearance-decodable, simulation-grounded, geometry-flexible, and compact, 2) it can be learned from internet-scale visual data and compress such data into a persistent physical state, and 3) the state can evolve through analytical or learned dynamics and expose perceptual prediction, counterfactual simulation, and control through task-specific query interfaces. The concrete choice of substrate is still an open-ended question. In the end, the central problem on the path toward physical world models is: \textit{what compact internal structure can preserve sufficient physical and semantic information to support all downstream projections of an embodied intelligence?}

\subsection{Foundation-Scale Interactive Simulators}
Finally, another use of world models is to build interactive simulators that support action imagination, task generation, feedback integration, or continual refinement. World models help simulators scale to provide a reusable environment in which agents and scientific systems can test actions, evaluate risks, and explore possible outcomes before real-world experiments.

This vision is motivated by the success of scaling in large language models and recent video generation models, where larger models, broader data, and greater computation have led to systematic improvements in reasoning and generalization. This raises a central question for world models: can similar scaling behavior emerge for physical dynamics? More specifically, can larger models, more diverse data, scalable interactive training datasets, and physics-rich annotations improve a world model's ability to maintain physical consistency, predict long-horizon dynamics, respond reliably to agent actions, and act as an interactive simulator? 

To achieve this goal, three key aspects need to be considered. First, scalable model architectures are required to support both high-fidelity world prediction and reliable action-conditioned interaction. Recent diffusion and flow-matching models provide high quality for visual futures, while autoregressive and LLM-style architectures offer advantages in long-context sequence modeling and reasoning. More recent unified multimodal models, such as Cosmos 3, and planner-renderer frameworks, such as Bernini, further suggest that language, vision, video, audio, and actions can be processed or coordinated within more unified modeling frameworks. For foundation-scale interactive simulators, the key architectural question remains open: what kind of model design can scale to controllable interaction while incorporating richer physical modalities, such as force, tactile feedback, contact states, and proprioception?

Second, the development of foundation-scale interactive simulators will depend on both internet videos and the construction of scalable physical datasets. Internet videos introduce general visual and action priors; first-person manipulation videos complement them with action or hand-motion annotations; and real robot trajectories provide grounded control signals. Beyond these sources, richer physical supervision, including 3D states, object motion, contact events, force, tactile feedback, thermal signals, and material properties, is also important. 

Third, a physical interactive simulator must be verified in closed-loop settings. A learned simulator may provide a low-cost interaction environment for policy training and evaluation, but its own predictions must still be validated against real-world outcomes or reliable physical constraints. Future systems therefore need evaluation protocols that go beyond visual plausibility, measuring whether predicted futures are causally consistent with actions, stable over long horizons, and predictive of real execution in robots or AI-driven autonomous wet lab environments. 
\section{Outlook: A Path to Physical AGI}
\label{sec:path-to-physical-agi}
Large language models (LLMs) grant machines an unprecedented mastery over human syntax, semantics, and stored knowledge. Yet, while LLMs possess an encyclopedic grasp of how we describe reality, they lack a fundamental grounding in physical reality itself. We are rapidly approaching the limits of disembodied intelligence \citep{coveney2025wall}.
The relationship between language models and world models defines the next major paradigm shift in AI development. To chart the broader trajectory toward AGI, we must recognize a fundamental truth about artificial cognition: i) language gave machines a way to talk about the world; ii) world models are how they will come to understand, imagine, reason with, and act in it. If AGI is required to navigate and manipulate the physical domain, it cannot accomplish these tasks by simply predicting the next word in a sequence. It must, in a native way, possess an intuitive grasp of physics, causality, spatial reasoning, and time. It must be able to simulate the consequences of its actions before taking them—predicting the next state of reality itself, and, more importantly, it needs to evaluate what tasks are possibly achievable in the real world given the current physical agent.

\paragraph{The Trinity Architecture: An Engine for Autonomous Evolution}
To realize this vision of Physical AGI, static models trained passively on historical data are insufficient. AGI requires a dynamic, self-evolving system capable of continuous interacting with the real world and self-directed learning. To this end, here we conceptualize this through a \textbf{Trinity Architecture}, which is a three-part cognitive loop designed to autonomously bridge the gap between digital reasoning and physical competence.
This architecture consists of three interdependent components:
\begin{itemize}
\item \textbf{Agent: The task execution engine.} The Actor, i.e., the physical agent, receives a task instruction and physically (or simulated-physically) attempts to accomplish it. It translates high-level intent into granular, sequential actions within its environment.
\item \textbf{Evaluator: The judge of task completion.} As the Agent operates, the Evaluator observes the trajectory data. It assesses how effectively and efficiently the task was accomplished, providing precise feedback on failures, physics violations, or suboptimal movements.
\item \textbf{World Model: The core simulator and curriculum designer.} The World Model ingests the trajectory data generated by the Agent and graded by the Evaluator. By observing these interactions, it learns the underlying physics, dynamics, and causality of the world. Crucially, the World Model understands the exact \emph{edge} of the Agent's current capabilities. Using this internal simulation, it imagines and proposes new, progressively complex tasks that are \textit{just beyond} the Agent's current limits, acting as an automated curriculum generator.
\end{itemize}
Note that here the World Model component act as a crucial role that i) internalize the knowledge of the world, ii) know about the edge the feasible tasks of the current Actor, iii) guide the next-round agent-world interaction by proposing the new tasks. This is also similar to how we human learn and use our world model in the brain \citep{taniguchi2023world,deng2025simura}.
Also, the Actor and Critic components can be surely empowered with LLMs, e.g., implemented via a LLM-based multi-agent systems \citep{yang2025unlocking}, thus the planning, reasoning and decision-making parts of these components natually depend on the cognitive intelligence from LLMs. In such a way, the Trinity Architecture provide a mutual enhancement protocol of both LLMs and the world model.

\begin{figure}
    \centering
    \includegraphics[width=0.85\linewidth]{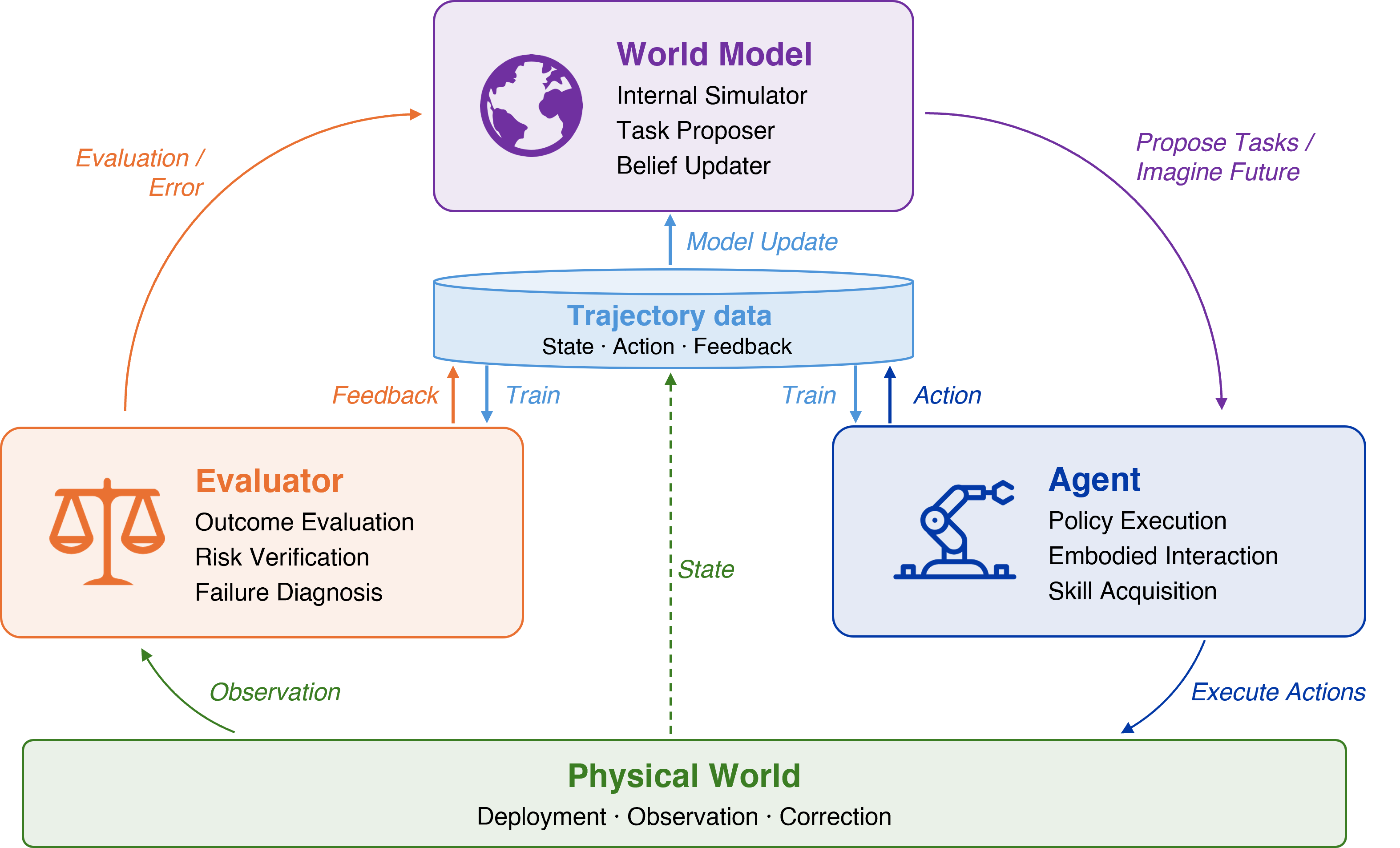}
    \caption{An overview of the trinity architecture.}
    \label{fig:trinity}
\end{figure}

\paragraph{Interacting with the World to Approach AGI}
The true power of this trinity lies in its continuous interaction with reality. This loop operates across both digital simulations and the physical world. The World Model can safely imagine thousands of scenarios and propose tasks for the Actor to attempt in a high-fidelity digital twin. Once the Actor masters these tasks, the system deploys its learned intuition to the physical world, where the Critic evaluates the inevitable friction and noise of reality, feeding that new data back into the World Model.
Through this relentless cycle of acting, evaluating, and updating its internal simulation, the AI stops relying solely on human-curated datasets. It begins to learn exactly the way biological intelligence does: through physical trial, error, and imagination. This Trinity Architecture is not just a blueprint for better robotics—it is the evolutionary engine that will transform AI from a passive conversationalist into an active Physical AGI.

\section*{Contributors}
The following are the contributors of this article, listed in alphabetically order of the family name:
Xinyuan Chen, Haoyu Guo, Shi Guo, Bingqi Jiang, Chunhua Shen, Xing Shen, Tianfan Xue, Yufei Xue, Mulin Yu, Weinan Zhang, Bin Zhao, Bowen Zhou, Ming Zhou.

\bibliographystyle{plainnat}
\bibliography{wm-pjlab}

@inproceedings{ha2018worldmodels,
  title={Recurrent World Models Facilitate Policy Evolution},
  author={Ha, David and Schmidhuber, J{\"u}rgen},
  booktitle={Advances in Neural Information Processing Systems (NeurIPS)},
  volume={31},
  year={2018}
}

@inproceedings{hu2022mile,
  title={Model-Based Imitation Learning for Urban Driving},
  author={Hu, Anthony and Corrado, Gianluca and Griffiths, Nicolas and Murez, Zak and Gurau, Corina and Yeo, Hudson and Kendall, Alex and Cipolla, Roberto and Shotton, Jamie},
  booktitle={Advances in Neural Information Processing Systems (NeurIPS)},
  volume={35},
  pages={20703--20716},
  year={2022}
}

@article{yang2025resim,
  title={ReSim: Reliable World Simulation for Autonomous Driving},
  author={Yang, Jiazhi and Chitta, Kashyap and Gao, Shenyuan and Chen, Long and Shao, Yuqian and Jia, Xiaosong and Li, Hongyang and Geiger, Andreas and Yue, Xiangyu and Chen, Li},
  journal={arXiv preprint arXiv:2506.09981},
  year={2025}
}

@inproceedings{li2024simpler,
  title={Evaluating Real-World Robot Manipulation Policies in Simulation},
  author={Li, Xuanlin and Hsu, Kyle and Gu, Jiayuan and Pertsch, Karl and Mees, Oier and Walke, Homer Rich and Fu, Chuyuan and Lunawat, Ishikaa and Sieh, Isabel and Kirmani, Sean and Levine, Sergey and Wu, Jiajun and Finn, Chelsea and Su, Hao and Vuong, Quan and Xiao, Ted},
  booktitle={Conference on Robot Learning (CoRL)},
  year={2024}
}

@inproceedings{wu2022daydreamer,
  title={DayDreamer: World Models for Physical Robot Learning},
  author={Wu, Philipp and Escontrela, Alejandro and Hafner, Danijar and Goldberg, Ken and Abbeel, Pieter},
  booktitle={Conference on Robot Learning (CoRL)},
  year={2022}
}

@inproceedings{yang2024unisim,
  title={Learning Interactive Real-World Simulators},
  author={Yang, Sherry and Du, Yilun and Ghasemipour, Seyed Kamyar Seyed and Tompson, Jonathan and Kaelbling, Leslie Pack and Schuurmans, Dale and Abbeel, Pieter},
  booktitle={International Conference on Learning Representations (ICLR)},
  year={2024}
}

@inproceedings{zhou2024robodreamer,
  title={RoboDreamer: Learning Compositional World Models for Robot Imagination},
  author={Zhou, Siyuan and Du, Yilun and Chen, Jiaben and Li, Yandong and Yeung, Dit-Yan and Gan, Chuang},
  booktitle={International Conference on Machine Learning (ICML)},
  year={2024}
}

@article{zeng2025bfm,
  title={Behavior Foundation Model for Humanoid Robots},
  author={Zeng, Weishuai and Lu, Shunlin and Yin, Kangning and Niu, Xiaojie and others},
  journal={arXiv preprint arXiv:2509.13780},
  year={2025}
}

@inproceedings{gao2024vista,
  title={Vista: A Generalizable Driving World Model with High Fidelity and Versatile Controllability},
  author={Gao, Shenyuan and Yang, Jiazhi and Chen, Li and others},
  booktitle={Advances in Neural Information Processing Systems (NeurIPS)},
  year={2024}
}

@article{taniguchi2023world,
  title={World models and predictive coding for cognitive and developmental robotics: frontiers and challenges},
  author={Taniguchi, Tadahiro and Murata, Shingo and Suzuki, Masahiro and Ognibene, Dimitri and Lanillos, Pablo and Ugur, Emre and Jamone, Lorenzo and Nakamura, Tomoaki and Ciria, Alejandra and Lara, Bruno and others},
  journal={Advanced Robotics},
  volume={37},
  number={13},
  pages={780--806},
  year={2023},
  publisher={Taylor \& Francis}
}

@inproceedings{yang2025unlocking,
  title={Unlocking the potential of decentralized llm-based mas: Privacy preservation and monetization in collective intelligence},
  author={Yang, Yingxuan and Peng, Qiuying and Wang, Jun and Wen, Ying and Zhang, Weinan},
  booktitle={Proceedings of the 24th International Conference on Autonomous Agents and Multiagent Systems},
  pages={2896--2900},
  year={2025}
}

@article{coveney2025wall,
  title={The wall confronting large language models},
  author={Coveney, Peter V and Succi, Sauro},
  journal={arXiv preprint arXiv:2507.19703},
  year={2025}
}

@article{deng2025simura,
  title={General Agentic Planning Through Simulative Reasoning with World Models},
  author={Deng, Mingkai and Hou, Jinyu and Hu, Zhiting and Xing, Eric},
  journal={arXiv preprint arXiv:2507.23773},
  year={2025},
  doi={10.48550/arXiv.2507.23773},
  url={https://arxiv.org/abs/2507.23773}
}

@misc{moerland2022modelbasedreinforcementlearningsurvey,
      title={Model-based Reinforcement Learning: A Survey}, 
      author={Thomas M. Moerland and Joost Broekens and Aske Plaat and Catholijn M. Jonker},
      year={2022},
      eprint={2006.16712},
      archivePrefix={arXiv},
      primaryClass={cs.LG},
      url={https://arxiv.org/abs/2006.16712}, 
}

@article{sutton1991dyna,
  title={Dyna, an integrated architecture for learning, planning, and reacting},
  author={Sutton, Richard S},
  journal={ACM SIGART Bulletin},
  volume={2},
  number={4},
  pages={160--163},
  year={1991},
  publisher={ACM New York, NY, USA}
}

@inproceedings{lewis2026acht,
  title     = {{ACHT-World}: Causal World Models for Closed-Loop Self-Driving Laboratories},
  author    = {Lewis, David Scott and Zueco, Enrique},
  booktitle = {AI4X Accelerate Conference},
  year      = {2026}
}

@article{parasuraman1997humans,
  title={Humans and automation: Use, misuse, disuse, abuse},
  author={Parasuraman, Raja and Riley, Victor},
  journal={Human factors},
  volume={39},
  number={2},
  pages={230--253},
  year={1997},
  publisher={SAGE Publications Sage CA: Los Angeles, CA}
}

@misc{billard2025designing,
  title={Designing and using robots for environmental sustainability},
  author={Billard, Aude G},
  journal={Science},
  volume={388},
  number={6744},
  pages={eadx2444},
  year={2025},
  publisher={American Association for the Advancement of Science}
}

@article{lee2025view,
  title={A view of the sustainable computing landscape},
  author={Lee, Benjamin C and Brooks, David and van Benthem, Arthur and Elgamal, Mariam and Gupta, Udit and Hills, Gage and Liu, Vincent and Phan, Linh Thi Xuan and Pierce, Benjamin and Stewart, Christopher and others},
  journal={Patterns},
  volume={6},
  number={7},
  year={2025},
  publisher={Elsevier}
}

@article{wang2026towards,
  title={Towards World Models in Biomedical Research},
  author={Wang, Guangyu and Yue, Jingkun and Zhang, Siqi and Liu, Yu and Wang, Xiaoyu and Meng, Mingyuan and Ji, Changwei and Han, Zongbo and Wang, Yulin and Yue, Yang and others},
  journal={arXiv preprint arXiv:2606.05925},
  year={2026}
}

@article{billard2025roadmap,
  title={A roadmap for AI in robotics},
  author={Billard, Aude and Albu-Schaeffer, Alin and Beetz, Michael and Burgard, Wolfram and Corke, Peter and Ciocarlie, Matei and Dahiya, Ravinder and Kragic, Danica and Goldberg, Ken and Nagai, Yukie and others},
  journal={Nature Machine Intelligence},
  volume={7},
  number={6},
  pages={818--824},
  year={2025},
  publisher={Nature Publishing Group UK London}
}

@inproceedings{deisenroth2011pilco,
  title     = {{PILCO}: A Model-Based and Data-Efficient Approach to Policy Search},
  author    = {Deisenroth, Marc and Rasmussen, Carl E.},
  booktitle = {Proceedings of the 28th International Conference on Machine Learning (ICML)},
  pages     = {465--472},
  year      = {2011},
  publisher = {Omnipress}
}

@inproceedings{nagabandi2018mbmf,
  title     = {Neural Network Dynamics for Model-Based Deep Reinforcement Learning 
               with Model-Free Fine-Tuning},
  author    = {Nagabandi, Anusha and Kahn, Gregory and Fearing, Ronald S. and Levine, Sergey},
  booktitle = {2018 IEEE International Conference on Robotics and Automation (ICRA)},
  pages     = {7559--7566},
  year      = {2018},
  organization = {IEEE}
}

@inproceedings{chua2018pets,
  title     = {Deep Reinforcement Learning in a Handful of Trials using 
               Probabilistic Dynamics Models},
  author    = {Chua, Kurtland and Calandra, Roberto and McAllister, Rowan and Levine, Sergey},
  booktitle = {Advances in Neural Information Processing Systems (NeurIPS)},
  volume    = {31},
  year      = {2018}
}

@inproceedings{janner2019mbpo,
  title     = {When to Trust Your Model: Model-Based Policy Optimization},
  author    = {Janner, Michael and Fu, Justin and Zhang, Marvin and Levine, Sergey},
  booktitle = {Advances in Neural Information Processing Systems (NeurIPS)},
  volume    = {32},
  year      = {2019}
}

@inproceedings{hafner2019dreamer,
  title     = {Dream to Control: Learning Behaviors by Latent Imagination},
  author    = {Hafner, Danijar and Lillicrap, Timothy and Ba, Jimmy and Norouzi, Mohammad},
  booktitle = {International Conference on Learning Representations (ICLR)},
  year      = {2020}
}

@inproceedings{hafner2020dreamerv2,
  title     = {Mastering {Atari} with Discrete World Models},
  author    = {Hafner, Danijar and Lillicrap, Timothy and Norouzi, Mohammad and Ba, Jimmy},
  booktitle = {International Conference on Learning Representations (ICLR)},
  year      = {2021}
}

@inproceedings{sutton1990dyna,
  title     = {Integrated Architectures for Learning, Planning, and Reacting Based on Approximating Dynamic Programming},
  author    = {Sutton, Richard S.},
  booktitle = {Proceedings of the 7th International Conference on Machine Learning (ICML)},
  pages     = {216--224},
  year      = {1990},
  publisher = {Morgan Kaufmann}
}

@inproceedings{williams2017mppi,
  title     = {Information Theoretic {MPC} for Model-Based Reinforcement Learning},
  author    = {Williams, Grady and Wagener, Nolan and Goldfain, Brian and Drews, Paul 
               and Rehg, James M. and Boots, Byron and Theodorou, Evangelos A.},
  booktitle = {2017 IEEE International Conference on Robotics and Automation (ICRA)},
  pages     = {1714--1721},
  year      = {2017},
  organization = {IEEE}
}

@inproceedings{heess2015svg,
  title     = {Learning Continuous Control Policies by Stochastic Value Gradients},
  author    = {Heess, Nicolas and Wayne, Gregory and Silver, David and Lillicrap, Timothy 
               and Tassa, Yuval and Erez, Tom},
  booktitle = {Advances in Neural Information Processing Systems (NeurIPS)},
  volume    = {28},
  year      = {2015}
}

@article{schrittwieser2020muzero,
  title={Mastering Atari, go, chess and shogi by planning with a learned model},
  author={Schrittwieser, Julian and Antonoglou, Ioannis and Hubert, Thomas and Simonyan, Karen and Sifre, Laurent and Schmitt, Simon and Guez, Arthur and Lockhart, Edward and Degrave, Nicolas and Oolrich, Stig and et al.},
  journal={Nature},
  volume={588},
  number={7839},
  pages={604--609},
  year={2020},
  publisher={Nature Publishing Group}
}

@inproceedings{yu2020mopo,
  author    = {Yu, Tianhe and Thomas, Garrett and Yu, Lantao and Ermon, Stefano and Zou, James Y. and Levine, Sergey and Finn, Chelsea and Ma, Tengyu},
  title     = {{MOPO}: Model-Based Offline Policy Optimization},
  booktitle = {Advances in Neural Information Processing Systems (NeurIPS)},
  year      = {2020}
}

@inproceedings{kidambi2020morel,
  author    = {Kidambi, Rahul and Rajeswaran, Aravind and Netrapalli, Praneeth and Joachims, Thorsten},
  title     = {{MOReL}: Model-Based Offline Reinforcement Learning},
  booktitle = {Advances in Neural Information Processing Systems (NeurIPS)},
  year      = {2020}
}

@inproceedings{lambert2020objectivemismatch,
  author    = {Lambert, Nathan and Amos, Brandon and Yadan, Omry and Calandra, Roberto},
  title     = {Objective Mismatch in Model-Based Reinforcement Learning},
  booktitle = {Learning for Dynamics and Control (L4DC)},
  year      = {2020}
}

@inproceedings{curi2020hucrl,
  author    = {Curi, Sebastian and Berkenkamp, Felix and Krause, Andreas},
  title     = {Efficient Model-Based Reinforcement Learning through Optimistic Policy Search and Planning},
  booktitle = {Advances in Neural Information Processing Systems (NeurIPS)},
  year      = {2020}
}

@inproceedings{farahmand2017vaml,
  author    = {Farahmand, Amir-massoud and Barreto, Andr{\'e} and Nikovski, Daniel},
  title     = {Value-Aware Loss Function for Model-Based Reinforcement Learning},
  booktitle = {Proceedings of the 20th International Conference on Artificial Intelligence and Statistics (AISTATS)},
  year      = {2017}
}

@inproceedings{grimm2020valueequivalence,
  author    = {Grimm, Christopher and Barreto, Andr{\'e} and Singh, Satinder and Silver, David},
  title     = {The Value Equivalence Principle for Model-Based Reinforcement Learning},
  booktitle = {Advances in Neural Information Processing Systems (NeurIPS)},
  year      = {2020}
}

@inproceedings{antonoglou2022stochasticmuzero,
  author    = {Antonoglou, Ioannis and Schrittwieser, Julian and Ozair, Sherjil and Hubert, Thomas K. and Silver, David},
  title     = {Planning in Stochastic Environments with a Learned Model},
  booktitle = {International Conference on Learning Representations (ICLR)},
  year      = {2022}
}

@inproceedings{talvitie2017selfcorrecting,
  author    = {Talvitie, Erik},
  title     = {Self-Correcting Models for Model-Based Reinforcement Learning},
  booktitle = {Proceedings of the AAAI Conference on Artificial Intelligence},
  year      = {2017}
}

@article{lambert2022compounding,
  author  = {Lambert, Nathan and Pister, Kristofer and Calandra, Roberto},
  title   = {Investigating Compounding Prediction Errors in Learned Dynamics Models},
  journal = {arXiv preprint arXiv:2203.09637},
  year    = {2022}
}

@inproceedings{hubert2021sampledmuzero,
  author    = {Hubert, Thomas and Schrittwieser, Julian and Antonoglou, Ioannis and Barekatain, Mohammadamin and Schmitt, Simon and Silver, David},
  title     = {Learning and Planning in Complex Action Spaces},
  booktitle = {Proceedings of the 38th International Conference on Machine Learning (ICML)},
  year      = {2021}
}

@inproceedings{hansen2022tdmpc,
  author    = {Hansen, Nicklas and Wang, Xiaolong and Su, Hao},
  title     = {Temporal Difference Learning for Model Predictive Control},
  booktitle = {Proceedings of the 39th International Conference on Machine Learning (ICML)},
  year      = {2022}
}

@inproceedings{pathak2017curiosity,
  author    = {Pathak, Deepak and Agrawal, Pulkit and Efros, Alexei A. and Darrell, Trevor},
  title     = {Curiosity-Driven Exploration by Self-Supervised Prediction},
  booktitle = {Proceedings of the 34th International Conference on Machine Learning (ICML)},
  year      = {2017}
}

@inproceedings{burda2019rnd,
  author    = {Burda, Yuri and Edwards, Harrison and Storkey, Amos and Klimov, Oleg},
  title     = {Exploration by Random Network Distillation},
  booktitle = {International Conference on Learning Representations (ICLR)},
  year      = {2019}
}

@inproceedings{houthooft2016vime,
  author    = {Houthooft, Rein and Chen, Xi and Duan, Yan and Schulman, John and De Turck, Filip and Abbeel, Pieter},
  title     = {{VIME}: Variational Information Maximizing Exploration},
  booktitle = {Advances in Neural Information Processing Systems (NeurIPS)},
  year      = {2016}
}

@inproceedings{sekar2020plan2explore,
  author    = {Sekar, Ramanan and Rybkin, Oleh and Daniilidis, Kostas and Abbeel, Pieter and Hafner, Danijar and Pathak, Deepak},
  title     = {Planning to Explore via Self-Supervised World Models},
  booktitle = {Proceedings of the 37th International Conference on Machine Learning (ICML)},
  year      = {2020}
}

@book{sutton2018rl,
  author    = {Sutton, Richard S. and Barto, Andrew G.},
  title     = {Reinforcement Learning: An Introduction},
  edition   = {2nd},
  publisher = {MIT Press},
  year      = {2018}
}

@article{silver2017alphagozero,
  author  = {Silver, David and Schrittwieser, Julian and Simonyan, Karen and Antonoglou, Ioannis and Huang, Aja and Guez, Arthur and Hubert, Thomas and Baker, Lucas and Lai, Matthew and Bolton, Adrian and others},
  title   = {Mastering the Game of {Go} without Human Knowledge},
  journal = {Nature},
  volume  = {550},
  pages   = {354--359},
  year    = {2017}
}

@article{silver2016alphago,
  author  = {Silver, David and Huang, Aja and Maddison, Chris J. and Guez, Arthur and Sifre, Laurent and van den Driessche, George and Schrittwieser, Julian and Antonoglou, Ioannis and Panneershelvam, Veda and Lanctot, Marc and others},
  title   = {Mastering the Game of {Go} with Deep Neural Networks and Tree Search},
  journal = {Nature},
  volume  = {529},
  pages   = {484--489},
  year    = {2016}
}

@book{kahneman2011thinking,
  author    = {Kahneman, Daniel},
  title     = {Thinking, Fast and Slow},
  publisher = {Farrar, Straus and Giroux},
  year      = {2011}
}

@inproceedings{ross2011dagger,
  title     = {A Reduction of Imitation Learning and Structured Prediction 
               to No-Regret Online Learning},
  author    = {Ross, St{\'e}phane and Gordon, Geoffrey and Bagnell, Drew},
  booktitle = {Proceedings of the 14th International Conference on 
               Artificial Intelligence and Statistics (AISTATS)},
  pages     = {627--635},
  year      = {2011}
}

@inproceedings{clavera2020mbmpo,
  title     = {Model-Based Meta-Policy Optimization},
  author    = {Clavera, Ignasi and Fu, Jonas and Abbeel, Pieter},
  booktitle = {Proceedings of the 2nd Conference on Learning for Dynamics 
               and Control (L4DC)},
  pages     = {119--129},
  year      = {2020},
  publisher = {PMLR}
}

@article{Schrittwieser2020,
  title     = {Mastering Atari, Go, chess and shogi by planning with a learned model},
  author    = {Schrittwieser, Julian and Antonoglou, Ioannis and Hubert, Thomas and Simonyan, Karen and Sifre, Laurent and Schmitt, Simon and Guez, Arthur and Lockhart, Edward and Degrave, Herbert and Ovsepyan, Oriol and Sonnerat, Timothée and Silver, David and Hassabis, Demis},
  journal   = {Nature},
  year      = {2020},
  volume    = {588},
  pages     = {604--609},
  doi       = {10.1038/s41586-020-03051-4},
  publisher = {Nature Publishing Group}
}

@article{kang2024far,
  title={How far is video generation from world model: A physical law perspective},
  author={Kang, Bingyi and Yue, Yang and Lu, Rui and Lin, Zhijie and Zhao, Yang and Wang, Kaixin and Huang, Gao and Feng, Jiashi},
  journal={arXiv preprint arXiv:2411.02385},
  year={2024}
}

@article{motamed2025generative,
  title={Do generative video models learn physical principles from watching videos?},
  author={Motamed, Saman and Culp, Laura and Swersky, Kevin and Jaini, Priyank and Geirhos, Robert},
  journal={arXiv e-prints},
  pages={arXiv--2501},
  year={2025}
}

@article{zhang2025morpheus,
  title={Evaluating Newtonian Mechanics in Video Generative Models with Real Physical Systems},
  author={Tragoudaras, Antonios and Zhang, Chenyu and Cherniavskii, Daniil and Vozikis, Antonios and Nijdam, Thijmen and Prinzhorn, Derck W. E. and Bodracska, Mark and Sebe, Nicu and Zadaianchuk, Andrii and Gavves, Efstratios},
  journal={arXiv preprint arXiv:2504.02918},
  year={2025},
  doi={10.48550/arXiv.2504.02918},
  url={https://arxiv.org/abs/2504.02918},
  note={ICML 2026}
}

@article{meng2024towards,
  title={Towards world simulator: Crafting physical commonsense-based benchmark for video generation},
  author={Meng, Fanqing and Liao, Jiaqi and Tan, Xinyu and Shao, Wenqi and Lu, Quanfeng and Zhang, Kaipeng and Cheng, Yu and Li, Dianqi and Qiao, Yu and Luo, Ping},
  journal={arXiv preprint arXiv:2410.05363},
  year={2024}
}

@article{gu2025phyworldbench,
  title={" PhyWorldBench": A Comprehensive Evaluation of Physical Realism in Text-to-Video Models},
  author={Gu, Jing and Liu, Xian and Zeng, Yu and Nagarajan, Ashwin and Zhu, Fangrui and Hong, Daniel and Fan, Yue and Yan, Qianqi and Zhou, Kaiwen and Liu, Ming-Yu and others},
  journal={arXiv preprint arXiv:2507.13428},
  year={2025}
}

@article{zheng2025vbench,
  title={Vbench-2.0: Advancing video generation benchmark suite for intrinsic faithfulness},
  author={Zheng, Dian and Huang, Ziqi and Liu, Hongbo and Zou, Kai and He, Yinan and Zhang, Fan and Gu, Lulu and Zhang, Yuanhan and He, Jingwen and Zheng, Wei-Shi and others},
  journal={arXiv preprint arXiv:2503.21755},
  year={2025}
}

@inproceedings{jiang2026robowm,
  title={RoboWM-Bench: A Benchmark for Evaluating World Models in Robotic Manipulation},
  author={Jiang, Feng and Chen, Yang and Xu, Kyle and Liu, Yuchen and Wang, Haifeng and Shen, Zhenhao and Lu, Jasper and Huang, Shengze and Wang, Yuanfei and Xie, Chen and others},
  booktitle={Proceedings of the IEEE/CVF Conference on Computer Vision and Pattern Recognition},
  pages={4455--4460},
  year={2026}
}

@article{shang2026worldarena,
  title={Worldarena: A unified benchmark for evaluating perception and functional utility of embodied world models},
  author={Shang, Yu and Li, Zhuohang and Ma, Yiding and Su, Weikang and Jin, Xin and Wang, Ziyou and Jin, Lei and Zhang, Xin and Tang, Yinzhou and Su, Haisheng and others},
  journal={arXiv preprint arXiv:2602.08971},
  year={2026}
}

@article{zhao2026phyworld,
  title={PhyWorld: Physics-Faithful World Model for Video Generation},
  author={Zhao, Pu and Lin, Juyi and Rupprecht, Timothy and Akbari, Arash and Yang, Chence and Chowdhury, Rahul and Motamedi, Elaheh and Akbari, Arman and He, Yumei and Wang, Chen and others},
  journal={arXiv preprint arXiv:2605.19242},
  year={2026}
}

@article{chen2026abot,
  title={Abot-physworld: Interactive world foundation model for robotic manipulation with physics alignment},
  author={Chen, Yuzhi and Chen, Ronghan and Huo, Dongjie and Yang, Yandan and Qi, Dekang and Liu, Haoyun and Lin, Tong and Zeng, Shuang and Xiao, Junjin and Chang, Xinyuan and others},
  journal={arXiv preprint arXiv:2603.23376},
  year={2026}
}

@article{wang2025physcorr,
  title={PhysCorr: Dual-Reward DPO for Physics-Constrained Text-to-Video Generation with Automated Preference Selection},
  author={Wang, Peiyao and Wang, Weining and Li, Qi},
  journal={arXiv preprint arXiv:2511.03997},
  year={2025}
}

@article{lin2026mmphysvideo,
  title={MMPhysVideo: Scaling Physical Plausibility in Video Generation via Joint Multimodal Modeling},
  author={Lin, Shubo and Zhang, Xuanyang and Cheng, Wei and Hu, Weiming and Yu, Gang and Gao, Jin},
  journal={arXiv preprint arXiv:2604.02817},
  year={2026}
}

@misc{openai2024sora,
  title        = {Video Generation Models as World Simulators},
  author       = {{OpenAI}},
  year         = {2024},
  url          = {https://openai.com/index/video-generation-models-as-world-simulators/}
}

@misc{chen2024diffusionforcing,
  title        = {Diffusion Forcing: Next-token Prediction Meets Full-Sequence Diffusion},
  author       = {Boyuan Chen and Diego Marti Monso and Yilun Du and Max Simchowitz and Russ Tedrake and Vincent Sitzmann},
  year         = {2024},
  eprint       = {2407.01392},
  archivePrefix = {arXiv},
  primaryClass = {cs.LG},
  url          = {https://arxiv.org/abs/2407.01392}
}

@misc{kondratyuk2024videopoet,
  title        = {VideoPoet: A Large Language Model for Zero-Shot Video Generation},
  author       = {Dan Kondratyuk and Lijun Yu and Xiuye Gu and Jos{\'e} Lezama and Jonathan Huang and Grant Schindler and Rachel Hornung and Vighnesh Birodkar and Jimmy Yan and Ming-Chang Chiu and Krishna Somandepalli and Hassan Akbari and Yair Alon and Yong Cheng and Josh Dillon and Agrim Gupta and Meera Hahn and Anja Hauth and David Hendon and Alonso Martinez and David Minnen and Mikhail Sirotenko and Kihyuk Sohn and Xuan Yang and Hartwig Adam and Ming-Hsuan Yang and Irfan Essa and Huisheng Wang and David A. Ross and Bryan Seybold and Lu Jiang},
  year         = {2024},
  eprint       = {2312.14125},
  archivePrefix = {arXiv},
  primaryClass = {cs.CV},
  url          = {https://arxiv.org/abs/2312.14125}
}

@misc{wan2025wan,
  title        = {Wan: Open and Advanced Large-Scale Video Generative Models},
  author       = {{Wan Team} and Ang Wang and Baole Ai and Bin Wen and Chaojie Mao and Chen-Wei Xie and Di Chen and Feiwu Yu and Haiming Zhao and Jianxiao Yang and Jianyuan Zeng and Jiayu Wang and Jingfeng Zhang and Jingren Zhou and Jinkai Wang and Jixuan Chen and Kai Zhu and Kang Zhao and Keyu Yan and Lianghua Huang and Mengyang Feng and Ningyi Zhang and Pandeng Li and Pingyu Wu and Ruihang Chu and Ruili Feng and Shiwei Zhang and Siyang Sun and Tao Fang and Tianxing Wang and Tianyi Gui and Tingyu Weng and Tong Shen and Wei Lin and Wei Wang and Wei Wang and Wenmeng Zhou and Wente Wang and Wenting Shen and Wenyuan Yu and Xianzhong Shi and Xiaoming Huang and Xin Xu and Yan Kou and Yangyu Lv and Yifei Li and Yijing Liu and Yiming Wang and Yingya Zhang and Yitong Huang and Yong Li and You Wu and Yu Liu and Yulin Pan and Yun Zheng and Yuntao Hong and Yupeng Shi and Yutong Feng and Zeyinzi Jiang and Zhen Han and Zhi-Fan Wu and Ziyu Liu},
  year         = {2025},
  eprint       = {2503.20314},
  archivePrefix = {arXiv},
  primaryClass = {cs.CV},
  doi          = {10.48550/arXiv.2503.20314},
  url          = {https://arxiv.org/abs/2503.20314}
}

@misc{nvidia2025cosmos,
  title        = {Cosmos World Foundation Model Platform for Physical AI},
  author       = {{NVIDIA} and Niket Agarwal and Arslan Ali and Maciej Bala and Yogesh Balaji and Erik Barker and Tiffany Cai and Prithvijit Chattopadhyay and Yongxin Chen and Yin Cui and Yifan Ding and Daniel Dworakowski and Jiaojiao Fan and Michele Fenzi and Francesco Ferroni and Sanja Fidler and Dieter Fox and Songwei Ge and Yunhao Ge and Jinwei Gu and Siddharth Gururani and Ethan He and Jiahui Huang and Jacob Huffman and Pooya Jannaty and Jingyi Jin and Seung Wook Kim and Gergely Kl{\'a}r and Grace Lam and Shiyi Lan and Laura Leal-Taixe and Anqi Li and Zhaoshuo Li and Chen-Hsuan Lin and Tsung-Yi Lin and Huan Ling and Ming-Yu Liu and Xian Liu and Alice Luo and Qianli Ma and Hanzi Mao and Kaichun Mo and Arsalan Mousavian and Seungjun Nah and Sriharsha Niverty and David Page and Despoina Paschalidou and Zeeshan Patel and Lindsey Pavao and Morteza Ramezanali and Fitsum Reda and Xiaowei Ren and Vasanth Rao Naik Sabavat and Ed Schmerling and Stella Shi and Bartosz Stefaniak and Shitao Tang and Lyne Tchapmi and Przemek Tredak and Wei-Cheng Tseng and Jibin Varghese and Hao Wang and Haoxiang Wang and Heng Wang and Ting-Chun Wang and Fangyin Wei and Xinyue Wei and Jay Zhangjie Wu and Jiashu Xu and Wei Yang and Lin Yen-Chen and Xiaohui Zeng and Yu Zeng and Jing Zhang and Qinsheng Zhang and Yuxuan Zhang and Qingqing Zhao and Artur Zolkowski},
  year         = {2025},
  eprint       = {2501.03575},
  archivePrefix = {arXiv},
  primaryClass = {cs.CV},
  url          = {https://arxiv.org/abs/2501.03575}
}

@misc{bruce2024genie,
  title        = {Genie: Generative Interactive Environments},
  author       = {Jake Bruce and Michael Dennis and Ashley Edwards and Jack Parker-Holder and Yuge Shi and Edward Hughes and Matthew Lai and Aditi Mavalankar and Richie Steigerwald and Chris Apps and Yusuf Aytar and Sarah Bechtle and Feryal Behbahani and Stephanie Chan and Nicolas Heess and Lucy Gonzalez and Simon Osindero and Sherjil Ozair and Scott Reed and Jingwei Zhang and Konrad Zolna and Jeff Clune and Nando de Freitas and Satinder Singh and Tim Rockt{\"a}schel},
  year         = {2024},
  eprint       = {2402.15391},
  archivePrefix = {arXiv},
  primaryClass = {cs.LG},
  url          = {https://arxiv.org/abs/2402.15391}
}

@misc{parkerholder2024genie2,
  title        = {{Genie 2}: A Large-Scale Foundation World Model},
  author       = {Parker-Holder, Jack and Ball, Philip and Bruce, Jake and Dasagi, Vibhavari and Holsheimer, Kristian and Kaplanis, Christos and Moufarek, Alexandre and Scully, Guy and Shar, Jeremy and Shi, Jimmy and Spencer, Stephen and Yung, Jessica and Dennis, Michael and Kenjeyev, Sultan and Long, Shangbang and Mnih, Vlad and Chan, Harris and Gazeau, Maxime and Li, Bonnie and Pardo, Fabio and Wang, Luyu and Zhang, Lei and Besse, Frederic and Harley, Tim and Mitenkova, Anna and Wang, Jane and Clune, Jeff and Hassabis, Demis and Hadsell, Raia and Bolton, Adrian and Singh, Satinder and Rockt{\"a}schel, Tim},
  year         = {2024},
  howpublished = {Google DeepMind blog},
  url          = {https://deepmind.google/discover/blog/genie-2-a-large-scale-foundation-world-model/}
}

@misc{genie3,
  title        = {{Genie 3}: A New Frontier for World Models},
  author       = {{Google DeepMind}},
  year         = {2025},
  howpublished = {Google DeepMind blog},
  url          = {https://deepmind.google/discover/blog/genie-3-a-new-frontier-for-world-models/}
}

@misc{valevski2025gamengen,
  title        = {Diffusion Models Are Real-Time Game Engines},
  author       = {Dani Valevski and Yaniv Leviathan and Moab Arar and Shlomi Fruchter},
  year         = {2025},
  eprint       = {2408.14837},
  archivePrefix = {arXiv},
  primaryClass = {cs.LG},
  url          = {https://arxiv.org/abs/2408.14837}
}

@misc{alonso2024diamond,
  title        = {Diffusion for World Modeling: Visual Details Matter in Atari},
  author       = {Eloi Alonso and Adam Jelley and Vincent Micheli and Anssi Kanervisto and Amos Storkey and Tim Pearce and Fran{\c{c}}ois Fleuret},
  year         = {2024},
  eprint       = {2405.12399},
  archivePrefix = {arXiv},
  primaryClass = {cs.LG},
  url          = {https://arxiv.org/abs/2405.12399}
}

@misc{decart2024oasis,
  title        = {Oasis: A Universe in a Transformer},
  author       = {{Decart} and {Etched}},
  year         = {2024},
  url          = {https://oasis-model.github.io/}
}

@misc{hu2023gaia1,
  title        = {GAIA-1: A Generative World Model for Autonomous Driving},
  author       = {Anthony Hu and Lloyd Russell and Hudson Yeo and Zak Murez and George Fedoseev and Alex Kendall and Jamie Shotton and Gianluca Corrado},
  year         = {2023},
  eprint       = {2309.17080},
  archivePrefix = {arXiv},
  primaryClass = {cs.CV},
  url          = {https://arxiv.org/abs/2309.17080}
}

@misc{wang2023drivedreamer,
  title        = {DriveDreamer: Towards Real-world-driven World Models for Autonomous Driving},
  author       = {Xiaofeng Wang and Zheng Zhu and Guan Huang and Xinze Chen and Jiagang Zhu and Jiwen Lu},
  year         = {2023},
  eprint       = {2309.09777},
  archivePrefix = {arXiv},
  primaryClass = {cs.CV},
  url          = {https://arxiv.org/abs/2309.09777}
}

@misc{wang2023drivewm,
  title        = {Driving into the Future: Multiview Visual Forecasting and Planning with World Model for Autonomous Driving},
  author       = {Yuqi Wang and Jiawei He and Lue Fan and Hongxin Li and Yuntao Chen and Zhaoxiang Zhang},
  year         = {2023},
  eprint       = {2311.17918},
  archivePrefix = {arXiv},
  primaryClass = {cs.CV},
  url          = {https://arxiv.org/abs/2311.17918}
}

@misc{waymo2026worldmodel,
  title        = {The Waymo World Model: A New Frontier for Autonomous Driving Simulation},
  author       = {Chiyu Max Jiang and Xander Masotto and Bo Sun},
  year         = {2026},
  month        = feb,
  url          = {https://waymo.com/blog/2026/02/the-waymo-world-model-a-new-frontier-for-autonomous-driving-simulation/}
}

@misc{gao2025seedance,
  title        = {Seedance 1.0: Exploring the Boundaries of Video Generation Models},
  author       = {Yu Gao and Haoyuan Guo and Tuyen Hoang and Weilin Huang and Lu Jiang and Fangyuan Kong and Huixia Li and Jiashi Li and Liang Li and Xiaojie Li and Xunsong Li and Yifu Li and Shanchuan Lin and Zhijie Lin and Jiawei Liu and Shu Liu and Xiaonan Nie and Zhiwu Qing and Yuxi Ren and Li Sun and Zhi Tian and Rui Wang and Sen Wang and Guoqiang Wei and Guohong Wu and Jie Wu and Ruiqi Xia and Fei Xiao and Xuefeng Xiao and Jiangqiao Yan and Ceyuan Yang and Jianchao Yang and Runkai Yang and Tao Yang and Yihang Yang and Zilyu Ye and Xuejiao Zeng and Yan Zeng and Heng Zhang and Yang Zhao and Xiaozheng Zheng and Peihao Zhu and Jiaxin Zou and Feilong Zuo},
  year         = {2025},
  eprint       = {2506.09113},
  archivePrefix = {arXiv},
  primaryClass = {cs.CV},
  url          = {https://arxiv.org/abs/2506.09113}
}

@misc{seedance2026seedance2,
  title        = {Seedance 2.0: Advancing Video Generation for World Complexity},
  author       = {{ByteDance Seed Team}},
  year         = {2026},
  eprint       = {2604.14148},
  archivePrefix = {arXiv},
  primaryClass = {cs.CV},
  doi          = {10.48550/arXiv.2604.14148},
  url          = {https://arxiv.org/abs/2604.14148}
}

@misc{happyhorse2026,
  title        = {Happy Horse 1.0},
  author       = {{Alibaba ATH Innovation Unit}},
  year         = {2026},
  url          = {https://fal.ai/happyhorse-1.0}
}

@misc{polyak2024moviegen,
  title        = {Movie Gen: A Cast of Media Foundation Models},
  author       = {Adam Polyak and Amit Zohar and Andrew Brown and Andros Tjandra and Animesh Sinha and Ann Lee and Apoorv Vyas and Bowen Shi and Chih-Yao Ma and Ching-Yao Chuang and David Yan and Dhruv Choudhary and Dingkang Wang and Geet Sethi and Guan Pang and Haoyu Ma and Ishan Misra and Ji Hou and Jialiang Wang and Kiran Jagadeesh and Kunpeng Li and Luxin Zhang and Mannat Singh and Mary Williamson and Matt Le and Matthew Yu and Mitesh Kumar Singh and Peizhao Zhang and Peter Vajda and Quentin Duval and Rohit Girdhar and Roshan Sumbaly and Sai Saketh Rambhatla and Sam Tsai and Samaneh Azadi and Samyak Datta and Sanyuan Chen and Sean Bell and Sharadh Ramaswamy and Shelly Sheynin and Siddharth Bhattacharya and Simran Motwani and Tao Xu and Tianhe Li and Tingbo Hou and Wei-Ning Hsu and Xi Yin and Xiaoliang Dai and Yaniv Taigman and Yaqiao Luo and Yen-Cheng Liu and Yi-Chiao Wu and Yue Zhao and Yuval Kirstain and Zecheng He and Zijian He and Albert Pumarola and Ali Thabet and Artsiom Sanakoyeu and Arun Mallya and Baishan Guo and Boris Araya and Breena Kerr and Carleigh Wood and Ce Liu and Cen Peng and Dimitry Vengertsev and Edgar Schonfeld and Elliot Blanchard and Felix Juefei-Xu and Fraylie Nord and Jeff Liang and John Hoffman and Jonas Kohler and Kaolin Fire and Karthik Sivakumar and Lawrence Chen and Licheng Yu and Luya Gao and Markos Georgopoulos and Rashel Moritz and Sara K. Sampson and Shikai Li and Simone Parmeggiani and Steve Fine and Tara Fowler and Vladan Petrovic and Yuming Du},
  year         = {2024},
  eprint       = {2410.13720},
  archivePrefix = {arXiv},
  primaryClass = {cs.CV},
  url          = {https://arxiv.org/abs/2410.13720}
}

@misc{kong2024hunyuanvideo,
  title        = {HunyuanVideo: A Systematic Framework For Large Video Generative Models},
  author       = {Weijie Kong and Qi Tian and Zijian Zhang and Rox Min and Zuozhuo Dai and Jin Zhou and Jiangfeng Xiong and Xin Li and Bo Wu and Jianwei Zhang and Kathrina Wu and Qin Lin and Junkun Yuan and Yanxin Long and Aladdin Wang and Andong Wang and Changlin Li and Duojun Huang and Fang Yang and Hao Tan and Hongmei Wang and Jacob Song and Jiawang Bai and Jianbing Wu and Jinbao Xue and Joey Wang and Kai Wang and Mengyang Liu and Pengyu Li and Shuai Li and Weiyan Wang and Wenqing Yu and Xinchi Deng and Yang Li and Yi Chen and Yutao Cui and Yuanbo Peng and Zhentao Yu and Zhiyu He and Zhiyong Xu and Zixiang Zhou and Zunnan Xu and Yangyu Tao and Qinglin Lu and Songtao Liu and Dax Zhou and Hongfa Wang and Yong Yang and Di Wang and Yuhong Liu and Jie Jiang and Caesar Zhong},
  year         = {2024},
  eprint       = {2412.03603},
  archivePrefix = {arXiv},
  primaryClass = {cs.CV},
  doi          = {10.48550/arXiv.2412.03603},
  url          = {https://arxiv.org/abs/2412.03603}
}

@misc{yang2024cogvideox,
  title        = {CogVideoX: Text-to-Video Diffusion Models with An Expert Transformer},
  author       = {Zhuoyi Yang and Jiayan Teng and Wendi Zheng and Ming Ding and Shiyu Huang and Jiazheng Xu and Yuanming Yang and Wenyi Hong and Xiaohan Zhang and Guanyu Feng and Da Yin and Yuxuan Zhang and Weihan Wang and Yean Cheng and Bin Xu and Xiaotao Gu and Yuxiao Dong and Jie Tang},
  year         = {2024},
  eprint       = {2408.06072},
  archivePrefix = {arXiv},
  primaryClass = {cs.CV},
  url          = {https://arxiv.org/abs/2408.06072}
}

@misc{kling2025omni,
  title        = {Kling-Omni Technical Report},
  author       = {{Kling Team} and Jialu Chen and Yuanzheng Ci and Xiangyu Du and Zipeng Feng and Kun Gai and Sainan Guo and Feng Han and Jingbin He and Kang He and Xiao Hu and Xiaohua Hu and Boyuan Jiang and Fangyuan Kong and Hang Li and Jie Li and Qingyu Li and Shen Li and Xiaohan Li and Yan Li and Jiajun Liang and Borui Liao and Yiqiao Liao and Weihong Lin and Quande Liu and Xiaokun Liu and Yilun Liu and Yuliang Liu and Shun Lu and Hangyu Mao and Yunyao Mao and Haodong Ouyang and Wenyu Qin and Wanqi Shi and Xiaoyu Shi and Lianghao Su and Haozhi Sun and Peiqin Sun and Pengfei Wan and Chao Wang and Chenyu Wang and Meng Wang and Qiulin Wang and Runqi Wang and Xintao Wang and Xuebo Wang and Zekun Wang and Min Wei and Tiancheng Wen and Guohao Wu and Xiaoshi Wu and Zhenhua Wu and Da Xie and Yingtong Xiong and Yulong Xu and Sile Yang and Zikang Yang and Weicai Ye and Ziyang Yuan and Shenglong Zhang and Shuaiyu Zhang and Yuanxing Zhang and Yufan Zhang and Wenzheng Zhao and Ruiliang Zhou and Yan Zhou and Guosheng Zhu and Yongjie Zhu},
  year         = {2025},
  eprint       = {2512.16776},
  archivePrefix = {arXiv},
  primaryClass = {cs.CV},
  doi          = {10.48550/arXiv.2512.16776},
  url          = {https://arxiv.org/abs/2512.16776}
}

@misc{liu2025lwm,
  title        = {World Model on Million-Length Video And Language With Blockwise RingAttention},
  author       = {Hao Liu and Wilson Yan and Matei Zaharia and Pieter Abbeel},
  year         = {2025},
  eprint       = {2402.08268},
  archivePrefix = {arXiv},
  primaryClass = {cs.LG},
  url          = {https://arxiv.org/abs/2402.08268}
}

@misc{ge2024worldgpt,
  title        = {WorldGPT: Empowering LLM as Multimodal World Model},
  author       = {Zhiqi Ge and Hongzhe Huang and Mingze Zhou and Juncheng Li and Guoming Wang and Siliang Tang and Yueting Zhuang},
  year         = {2024},
  eprint       = {2404.18202},
  archivePrefix = {arXiv},
  primaryClass = {cs.AI},
  url          = {https://arxiv.org/abs/2404.18202}
}

@misc{zhou2025hermes,
  title        = {HERMES: A Unified Self-Driving World Model for Simultaneous 3D Scene Understanding and Generation},
  author       = {Xin Zhou and Dingkang Liang and Sifan Tu and Xiwu Chen and Yikang Ding and Dingyuan Zhang and Feiyang Tan and Hengshuang Zhao and Xiang Bai},
  year         = {2025},
  eprint       = {2501.14729},
  archivePrefix = {arXiv},
  primaryClass = {cs.CV},
  url          = {https://arxiv.org/abs/2501.14729}
}

@misc{zhou2026hermespp,
  title        = {HERMES++: Toward a Unified Driving World Model for 3D Scene Understanding and Generation},
  author       = {Xin Zhou and Dingkang Liang and Xiwu Chen and Feiyang Tan and Dingyuan Zhang and Hengshuang Zhao and Xiang Bai},
  year         = {2026},
  eprint       = {2604.28196},
  archivePrefix = {arXiv},
  primaryClass = {cs.CV},
  url          = {https://arxiv.org/abs/2604.28196}
}

@misc{deng2026gaussiandwm,
  title        = {GaussianDWM: 3D Gaussian Driving World Model for Unified Scene Understanding and Multi-Modal Generation},
  author       = {Tianchen Deng and Xuefeng Chen and Yi Chen and Qu Chen and Yuyao Xu and Lijin Yang and Le Xu and Yu Zhang and Bo Zhang and Wuxiong Huang and Hesheng Wang},
  year         = {2026},
  eprint       = {2512.23180},
  archivePrefix = {arXiv},
  primaryClass = {cs.CV},
  url          = {https://arxiv.org/abs/2512.23180}
}

@misc{xiong2026unidrivewm,
  title        = {UniDrive-WM: Unified Understanding, Planning and Generation World Model For Autonomous Driving},
  author       = {Zhexiao Xiong and Xin Ye and Burhan Yaman and Sheng Cheng and Yiren Lu and Jingru Luo and Nathan Jacobs and Liu Ren},
  year         = {2026},
  eprint       = {2601.04453},
  archivePrefix = {arXiv},
  primaryClass = {cs.CV},
  url          = {https://arxiv.org/abs/2601.04453}
}

@misc{chameleon2025,
  title        = {Chameleon: Mixed-Modal Early-Fusion Foundation Models},
  author       = {{Chameleon Team}},
  year         = {2025},
  eprint       = {2405.09818},
  archivePrefix = {arXiv},
  primaryClass = {cs.CL},
  url          = {https://arxiv.org/abs/2405.09818}
}

@misc{wu2024janus,
  title        = {Janus: Decoupling Visual Encoding for Unified Multimodal Understanding and Generation},
  author       = {Chengyue Wu and Xiaokang Chen and Zhiyu Wu and Yiyang Ma and Xingchao Liu and Zizheng Pan and Wen Liu and Zhenda Xie and Xingkai Yu and Chong Ruan and Ping Luo},
  year         = {2024},
  eprint       = {2410.13848},
  archivePrefix = {arXiv},
  primaryClass = {cs.CV},
  url          = {https://arxiv.org/abs/2410.13848}
}

@misc{chen2025januspro,
  title        = {Janus-Pro: Unified Multimodal Understanding and Generation with Data and Model Scaling},
  author       = {Xiaokang Chen and Zhiyu Wu and Xingchao Liu and Zizheng Pan and Wen Liu and Zhenda Xie and Xingkai Yu and Chong Ruan},
  year         = {2025},
  eprint       = {2501.17811},
  archivePrefix = {arXiv},
  primaryClass = {cs.AI},
  url          = {https://arxiv.org/abs/2501.17811}
}

@misc{wei2026trinityconsistencydefiningprinciple,
      title={The Trinity of Consistency as a Defining Principle for General World Models}, 
      author={Jingxuan Wei and Siyuan Li and Yuhang Xu and Zheng Sun and Junjie Jiang and Hexuan Jin and Caijun Jia and Honghao He and Xinglong Xu and Xi bai and Chang Yu and Yumou Liu and Junnan Zhu and Xuanhe Zhou and Jintao Chen and Xiaobin Hu and Shancheng Pang and Bihui Yu and Ran He and Zhen Lei and Stan Z. Li and Conghui He and Shuicheng Yan and Cheng Tan},
      year={2026},
      eprint={2602.23152},
      archivePrefix={arXiv},
      primaryClass={cs.AI},
      url={https://arxiv.org/abs/2602.23152}, 
}

@misc{wang2024emu3,
  title        = {Emu3: Next-Token Prediction is All You Need},
  author       = {Xinlong Wang and Xiaosong Zhang and Zhengxiong Luo and Quan Sun and Yufeng Cui and Jinsheng Wang and Fan Zhang and Yueze Wang and Zhen Li and Qiying Yu and Yingli Zhao and Yulong Ao and Xuebin Min and Tao Li and Boya Wu and Bo Zhao and Bowen Zhang and Liangdong Wang and Guang Liu and Zheqi He and Xi Yang and Jingjing Liu and Yonghua Lin and Tiejun Huang and Zhongyuan Wang},
  year         = {2024},
  eprint       = {2409.18869},
  archivePrefix = {arXiv},
  primaryClass = {cs.CV},
  url          = {https://arxiv.org/abs/2409.18869}
}

@misc{cui2025emu35,
  title        = {Emu3.5: Native Multimodal Models are World Learners},
  author       = {Yufeng Cui and Honghao Chen and Haoge Deng and Xu Huang and Xinghang Li and Jirong Liu and Yang Liu and Zhuoyan Luo and Jinsheng Wang and Wenxuan Wang and Yueze Wang and Chengyuan Wang and Fan Zhang and Yingli Zhao and Ting Pan and Xianduo Li and Zecheng Hao and Wenxuan Ma and Zhuo Chen and Yulong Ao and Tiejun Huang and Zhongyuan Wang and Xinlong Wang},
  year         = {2025},
  eprint       = {2510.26583},
  archivePrefix = {arXiv},
  primaryClass = {cs.CV},
  url          = {https://arxiv.org/abs/2510.26583}
}

@misc{wang2026wam_survey,
  title        = {World Action Models: The Next Frontier in Embodied AI},
  author       = {Siyin Wang and Junhao Shi and Zhaoyang Fu and Xinzhe He and Feihong Liu and Chenchen Yang and Yikang Zhou and Zhaoye Fei and Jingjing Gong and Jinlan Fu and Mike Zheng Shou and Xuanjing Huang and Xipeng Qiu and Yu-Gang Jiang},
  year         = {2026},
  eprint       = {2605.12090},
  archivePrefix = {arXiv},
  primaryClass = {cs.AI},
  url          = {https://arxiv.org/abs/2605.12090}
}

@misc{ye2026dreamzero,
  title        = {World Action Models are Zero-shot Policies},
  author       = {Seonghyeon Ye and Yunhao Ge and Kaiyuan Zheng and Shenyuan Gao and Sihyun Yu and George Kurian and Suneel Indupuru and You Liang Tan and Chuning Zhu and Jiannan Xiang and Ayaan Malik and Kyungmin Lee and William Liang and Nadun Ranawaka and Jiasheng Gu and Yinzhen Xu and Guanzhi Wang and Fengyuan Hu and Avnish Narayan and Johan Bjorck and Jing Wang and Gwanghyun Kim and Dantong Niu and Ruijie Zheng and Yuqi Xie and Jimmy Wu and Qi Wang and Ryan Julian and Danfei Xu and Yilun Du and Yevgen Chebotar and Scott Reed and Jan Kautz and Yuke Zhu and Linxi Jim Fan and Joel Jang},
  year         = {2026},
  eprint       = {2602.15922},
  archivePrefix = {arXiv},
  primaryClass = {cs.RO},
  url          = {https://arxiv.org/abs/2602.15922}
}

@misc{Ye2026b,
  title         = {World Action Models are Zero-shot Policies},
  author        = {Seonghyeon Ye and Yunhao Ge and Kaiyuan Zheng and Shenyuan Gao and Sihyun Yu and George Kurian and Suneel Indupuru and You Liang Tan and Chuning Zhu and Jiannan Xiang and Ayaan Malik and Kyungmin Lee and William Liang and Nadun Ranawaka and Jiasheng Gu and Yinzhen Xu and Guanzhi Wang and Fengyuan Hu and Avnish Narayan and Johan Bjorck and Jing Wang and Gwanghyun Kim and Dantong Niu and Ruijie Zheng and Yuqi Xie and Jimmy Wu and Qi Wang and Ryan Julian and Danfei Xu and Yilun Du and Yevgen Chebotar and Scott Reed and Jan Kautz and Yuke Zhu and Linxi Jim Fan and Joel Jang},
  year          = {2026},
  eprint        = {2602.15922},
  archivePrefix = {arXiv},
  primaryClass  = {cs.RO},
  doi           = {10.48550/arXiv.2602.15922},
  url           = {https://arxiv.org/abs/2602.15922}
}

@misc{li2026lingbotva,
  title        = {Causal World Modeling for Robot Control},
  author       = {Lin Li and Qihang Zhang and Yiming Luo and Shuai Yang and Ruilin Wang and Fei Han and Mingrui Yu and Zelin Gao and Nan Xue and Xing Zhu and Yujun Shen and Yinghao Xu},
  year         = {2026},
  eprint       = {2601.21998},
  archivePrefix = {arXiv},
  primaryClass = {cs.RO},
  url          = {https://arxiv.org/abs/2601.21998}
}

@misc{zhou2026tau0wm,
  title        = {{$\tau$0-WM: A Unified Video-Action World Model for Robotic Manipulation}},
  author       = {Pengfei Zhou and Shengcong Chen and Di Chen and Jiaxu Wang and Rongjun Jin and Bingwen Zhu and Yike Pan and Songen Gu and Kuanning Wang and Shufeng Nan and Xingyu Qiu and Chenhao Qiu and Pu Yang and Yunuo Cai and Jianxiong Gao and Yifan Li and Yanwei Fu and Xiangyu Yue and Zhi Chen and Jianlan Luo},
  year         = {2026},
  month        = may,
  url          = {https://finch.agibot.com/research/tau0-wm},
  note         = {Technical report}
}

@misc{zhu2025uwm,
  title        = {Unified World Models: Coupling Video and Action Diffusion for Pretraining on Large Robotic Datasets},
  author       = {Chuning Zhu and Raymond Yu and Siyuan Feng and Benjamin Burchfiel and Paarth Shah and Abhishek Gupta},
  year         = {2025},
  eprint       = {2504.02792},
  archivePrefix = {arXiv},
  primaryClass = {cs.RO},
  url          = {https://arxiv.org/abs/2504.02792}
}

@misc{kim2026cosmospolicy,
  title        = {Cosmos Policy: Fine-Tuning Video Models for Visuomotor Control and Planning},
  author       = {Moo Jin Kim and Yihuai Gao and Tsung-Yi Lin and Yen-Chen Lin and Yunhao Ge and Grace Lam and Percy Liang and Shuran Song and Ming-Yu Liu and Chelsea Finn and Jinwei Gu},
  year         = {2026},
  eprint       = {2601.16163},
  archivePrefix = {arXiv},
  primaryClass = {cs.RO},
  url          = {https://arxiv.org/abs/2601.16163}
}

@misc{ye2026gigaworldpolicy,
  title        = {GigaWorld-Policy: An Efficient Action-Centered World--Action Model},
  author       = {Angen Ye and Boyuan Wang and Chaojun Ni and Guan Huang and Guosheng Zhao and Hao Li and Hengtao Li and Jie Li and Jindi Lv and Jingyu Liu and Min Cao and Peng Li and Qiuping Deng and Wenjun Mei and Xiaofeng Wang and Xinze Chen and Xinyu Zhou and Yang Wang and Yifan Chang and Yifan Li and Yukun Zhou and Yun Ye and Zhichao Liu and Zheng Zhu},
  year         = {2026},
  eprint       = {2603.17240},
  archivePrefix = {arXiv},
  primaryClass = {cs.RO},
  url          = {https://arxiv.org/abs/2603.17240}
}

@misc{yuan2026fastwam,
  title        = {Fast-WAM: Do World Action Models Need Test-time Future Imagination?},
  author       = {Tianyuan Yuan and Zibin Dong and Yicheng Liu and Hang Zhao},
  year         = {2026},
  eprint       = {2603.16666},
  archivePrefix = {arXiv},
  primaryClass = {cs.RO},
  url          = {https://arxiv.org/abs/2603.16666}
}

@misc{akbari2026flashwam,
  title        = {Flash-WAM: Modality-Aware Distillation for World Action Models},
  author       = {Arman Akbari and Ci Zhang and Arash Akbari and Lin Zhao and Yixiao Chen and Weiwei Chen and Xuan Zhang and Geng Yuan and Yanzhi Wang},
  year         = {2026},
  eprint       = {2606.05254},
  archivePrefix = {arXiv},
  primaryClass = {cs.RO},
  url          = {https://arxiv.org/abs/2606.05254}
}

@misc{nvidia2026cosmos3,
  title        = {Cosmos 3: Omnimodal World Models for Physical AI},
  author       = {{NVIDIA Cosmos Team}},
  year         = {2026},
  eprint       = {2606.02800},
  archivePrefix = {arXiv},
  primaryClass = {cs.CV},
  doi          = {10.48550/arXiv.2606.02800},
  url          = {https://arxiv.org/abs/2606.02800}
}

@misc{hafner2019planet,
  title        = {Learning Latent Dynamics for Planning from Pixels},
  author       = {Danijar Hafner and Timothy Lillicrap and Ian Fischer and Ruben Villegas and David Ha and Honglak Lee and James Davidson},
  year         = {2019},
  eprint       = {1811.04551},
  archivePrefix = {arXiv},
  primaryClass = {cs.LG},
  url          = {https://arxiv.org/abs/1811.04551}
}

@misc{hafner2021dreamerv2,
  title        = {Mastering Atari with Discrete World Models},
  author       = {Danijar Hafner and Timothy Lillicrap and Mohammad Norouzi and Jimmy Ba},
  year         = {2021},
  eprint       = {2010.02193},
  archivePrefix = {arXiv},
  primaryClass = {cs.LG},
  url          = {https://arxiv.org/abs/2010.02193}
}

@misc{hafner2023dreamerv3,
  title        = {Mastering Diverse Domains through World Models},
  author       = {Danijar Hafner and Jurgis Pasukonis and Jimmy Ba and Timothy Lillicrap},
  year         = {2023},
  eprint       = {2301.04104},
  archivePrefix = {arXiv},
  primaryClass = {cs.LG},
  url          = {https://arxiv.org/abs/2301.04104}
}

@misc{hansen2023tdmpc2,
  title        = {TD-MPC2: Scalable, Robust World Models for Continuous Control},
  author       = {Nicklas Hansen and Hao Su and Xiaolong Wang},
  year         = {2023},
  eprint       = {2310.16828},
  archivePrefix = {arXiv},
  primaryClass = {cs.RO},
  url          = {https://arxiv.org/abs/2310.16828}
}

@misc{kipf2019cswm,
  title        = {Contrastive Learning of Structured World Models},
  author       = {Thomas Kipf and Elise van der Pol and Max Welling},
  year         = {2019},
  eprint       = {1911.12247},
  archivePrefix = {arXiv},
  primaryClass = {cs.LG},
  url          = {https://arxiv.org/abs/1911.12247}
}

@misc{kipf2019contrastive,
  title         = {Contrastive Learning of Structured World Models},
  author        = {Thomas Kipf and Elise van der Pol and Max Welling},
  year          = {2019},
  eprint        = {1911.12247},
  archivePrefix = {arXiv},
  primaryClass  = {cs.LG},
  doi           = {10.48550/arXiv.1911.12247},
  url           = {https://arxiv.org/abs/1911.12247}
}

@misc{okada2022dreamingv2,
  title        = {DreamingV2: Reinforcement Learning with Discrete World Models without Reconstruction},
  author       = {Masashi Okada and Tadahiro Taniguchi},
  year         = {2022},
  eprint       = {2203.00494},
  archivePrefix = {arXiv},
  primaryClass = {cs.LG},
  url          = {https://arxiv.org/abs/2203.00494}
}

@misc{poudel2023recore,
  title        = {ReCoRe: Regularized Contrastive Representation Learning of World Model},
  author       = {Rudra P. K. Poudel and Harit Pandya and Stephan Liwicki and Roberto Cipolla},
  year         = {2023},
  eprint       = {2312.09056},
  archivePrefix = {arXiv},
  primaryClass = {cs.LG},
  url          = {https://arxiv.org/abs/2312.09056}
}

@misc{assran2023ijepa,
  title        = {Self-Supervised Learning from Images with a Joint-Embedding Predictive Architecture},
  author       = {Mahmoud Assran and Quentin Duval and Ishan Misra and Piotr Bojanowski and Pascal Vincent and Michael Rabbat and Yann LeCun and Nicolas Ballas},
  year         = {2023},
  eprint       = {2301.08243},
  archivePrefix = {arXiv},
  primaryClass = {cs.CV},
  url          = {https://arxiv.org/abs/2301.08243}
}

@misc{bardes2024vjepa,
  title        = {Revisiting Feature Prediction for Learning Visual Representations from Video},
  author       = {Adrien Bardes and Quentin Garrido and Jean Ponce and Xinlei Chen and Michael Rabbat and Yann LeCun and Mahmoud Assran and Nicolas Ballas},
  year         = {2024},
  eprint       = {2404.08471},
  archivePrefix = {arXiv},
  primaryClass = {cs.CV},
  url          = {https://arxiv.org/abs/2404.08471}
}

@misc{assran2025vjepa2,
  title        = {V-JEPA 2: Self-Supervised Video Models Enable Understanding, Prediction and Planning},
  author       = {Mido Assran and Adrien Bardes and David Fan and Quentin Garrido and Russell Howes and Mojtaba Komeili and Matthew Muckley and Ammar Rizvi and Claire Roberts and Koustuv Sinha and Artem Zholus and Sergio Arnaud and Abha Gejji and Ada Martin and Francois Robert Hogan and Daniel Dugas and Piotr Bojanowski and Vasil Khalidov and Patrick Labatut and Francisco Massa and Marc Szafraniec and Kapil Krishnakumar and Yong Li and Xiaodong Ma and Sarath Chandar and Franziska Meier and Yann LeCun and Michael Rabbat and Nicolas Ballas},
  year         = {2025},
  eprint       = {2506.09985},
  archivePrefix = {arXiv},
  primaryClass = {cs.CV},
  url          = {https://arxiv.org/abs/2506.09985}
}

@misc{bi2025motus,
  title        = {Motus: A Unified Latent Action World Model},
  author       = {Hongzhe Bi and Hengkai Tan and Shenghao Xie and Zeyuan Wang and Shuhe Huang and Haitian Liu and Ruowen Zhao and Yao Feng and Chendong Xiang and Yinze Rong and Hongyan Zhao and Hanyu Liu and Zhizhong Su and Lei Ma and Hang Su and Jun Zhu},
  year         = {2025},
  eprint       = {2512.13030},
  archivePrefix = {arXiv},
  primaryClass = {cs.RO},
  url          = {https://arxiv.org/abs/2512.13030}
}

@misc{lyu2026lda1b,
  title        = {LDA-1B: Scaling Latent Dynamics Action Model via Universal Embodied Data Ingestion},
  author       = {Jiangran Lyu and Kai Liu and Xuheng Zhang and Haoran Liao and Yusen Feng and Wenxuan Zhu and Tingrui Shen and Jiayi Chen and Jiazhao Zhang and Yifei Dong and Wenbo Cui and Senmao Qi and Shuo Wang and Yixin Zheng and Mi Yan and Xuesong Shi and Haoran Li and Dongbin Zhao and Ming-Yu Liu and Zhizheng Zhang and Li Yi and Yizhou Wang and He Wang},
  year         = {2026},
  eprint       = {2602.12215},
  archivePrefix = {arXiv},
  primaryClass = {cs.RO},
  url          = {https://arxiv.org/abs/2602.12215}
}

@misc{worldlabs2026taxonomy,
  title        = {A Functional Taxonomy of World Models},
  author       = {Li Fei-Fei},
  year         = {2026},
  howpublished = {\url{https://www.worldlabs.ai/blog/taxonomy-of-world-models}},
  note         = {Accessed: 2026-06-15}
}

@misc{genesisworld2026,
  title        = {Genesis World: Simulation Platform for Physical AI Development},
  author       = {{Genesis-Embodied-AI}},
  year         = {2026},
  howpublished = {\url{https://github.com/Genesis-Embodied-AI/genesis-world}},
  note         = {Accessed: 2026-06-15}
}

@misc{lightwheel2026,
  title        = {Lightwheel: Data and Simulation Infrastructure for Physical AI},
  author       = {{Lightwheel}},
  year         = {2026},
  howpublished = {\url{https://lightwheel.ai/}},
  note         = {Accessed: 2026-06-15}
}

@inproceedings{cao2026physx,
  title={Physx-anything: Simulation-ready physical 3d assets from single image},
  author={Cao, Ziang and Hong, Fangzhou and Chen, Zhaoxi and Pan, Liang and Liu, Ziwei},
  booktitle={Proceedings of the IEEE/CVF Conference on Computer Vision and Pattern Recognition},
  pages={5839--5848},
  year={2026}
}

@article{cao2026phyomni,
  title={PhysX-Omni: Unified Simulation-Ready Physical 3D Generation for Rigid, Deformable, and Articulated Objects},
  author={Cao, Ziang and Liu, Yinghao and Li, Haitian and Yao, Runmao and Hong, Fangzhou and Chen, Zhaoxi and Pan, Liang and Liu, Ziwei},
  journal={arXiv preprint arXiv:2605.21572},
  year={2026}
}

@article{zhao2025hunyuan3d,
  title={Hunyuan3d 2.0: Scaling diffusion models for high resolution textured 3d assets generation},
  author={Zhao, Zibo and Lai, Zeqiang and Lin, Qingxiang and Zhao, Yunfei and Liu, Haolin and Yang, Shuhui and Feng, Yifei and Yang, Mingxin and Zhang, Sheng and Yang, Xianghui and others},
  journal={arXiv preprint arXiv:2501.12202},
  year={2025}
}

@article{hunyuan3d2025hunyuan3d,
  title={Hunyuan3d 2.1: From images to high-fidelity 3d assets with production-ready pbr material},
  author={Hunyuan3D, Team and Yang, Shuhui and Yang, Mingxin and Feng, Yifei and Huang, Xin and Zhang, Sheng and He, Zebin and Luo, Di and Liu, Haolin and Zhao, Yunfei and others},
  journal={arXiv preprint arXiv:2506.15442},
  year={2025}
}

@article{li2025triposg,
  title={Triposg: High-fidelity 3d shape synthesis using large-scale rectified flow models},
  author={Li, Yangguang and Zou, Zi-Xin and Liu, Zexiang and Wang, Dehu and Liang, Yuan and Yu, Zhipeng and Liu, Xingchao and Guo, Yuan-Chen and Liang, Ding and Ouyang, Wanli and others},
  journal={IEEE Transactions on Pattern Analysis and Machine Intelligence},
  year={2025},
  publisher={IEEE}
}

@article{chang2017matterport3d,
  title={Matterport3d: Learning from rgb-d data in indoor environments},
  author={Chang, Angel and Dai, Angela and Funkhouser, Thomas and Halber, Maciej and Niessner, Matthias and Savva, Manolis and Song, Shuran and Zeng, Andy and Zhang, Yinda},
  journal={arXiv preprint arXiv:1709.06158},
  year={2017}
}

@inproceedings{savva2019habitat,
  title={Habitat: A platform for embodied ai research},
  author={Savva, Manolis and Kadian, Abhishek and Maksymets, Oleksandr and Zhao, Yili and Wijmans, Erik and Jain, Bhavana and Straub, Julian and Liu, Jia and Koltun, Vladlen and Malik, Jitendra and others},
  booktitle={Proceedings of the IEEE/CVF international conference on computer vision},
  pages={9339--9347},
  year={2019}
}

@inproceedings{puig2024habitat,
  title={Habitat 3.0: A co-habitat for humans, avatars, and robots},
  author={Puig, Xavier and Undersander, Eric and Szot, Andrew and Dallaire Cote, Mikael and Yang, Tsung-Yen and Partsey, Ruslan and Desai, Ruta and Clegg, Alexander and Hlavac, Michal and Min, So Yeon and others},
  booktitle={International Conference on Learning Representations},
  volume={2024},
  pages={15306--15336},
  year={2024}
}

@inproceedings{dosovitskiy2017carla,
  title={CARLA: An open urban driving simulator},
  author={Dosovitskiy, Alexey and Ros, German and Codevilla, Felipe and Lopez, Antonio and Koltun, Vladlen},
  booktitle={Conference on robot learning},
  pages={1--16},
  year={2017},
  organization={PMLR}
}

@article{gan2020threedworld,
  title={Threedworld: A platform for interactive multi-modal physical simulation},
  author={Gan, Chuang and Schwartz, Jeremy and Alter, Seth and Mrowca, Damian and Schrimpf, Martin and Traer, James and De Freitas, Julian and Kubilius, Jonas and Bhandwaldar, Abhishek and Haber, Nick and others},
  journal={arXiv preprint arXiv:2007.04954},
  year={2020}
}

@article{gao2026nvidia,
  title={NVIDIA Isaac Sim: Enabling Scalable, GPU-Accelerated Simulation for Robotics},
  author={Gao, Sicong and Pagnucco, Maurice and Bednarz, Tomasz and Song, Yang},
  journal={arXiv preprint arXiv:2606.03551},
  year={2026}
}

@inproceedings{rombach2022high,
  title={High-resolution image synthesis with latent diffusion models},
  author={Rombach, Robin and Blattmann, Andreas and Lorenz, Dominik and Esser, Patrick and Ommer, Bj{\"o}rn},
  booktitle={Proceedings of the IEEE/CVF conference on computer vision and pattern recognition},
  pages={10684--10695},
  year={2022}
}

@article{hu2024video,
  title={Video prediction policy: A generalist robot policy with predictive visual representations},
  author={Hu, Yucheng and Guo, Yanjiang and Wang, Pengchao and Chen, Xiaoyu and Wang, Yen-Jen and Zhang, Jianke and Sreenath, Koushil and Lu, Chaochao and Chen, Jianyu},
  journal={arXiv preprint arXiv:2412.14803},
  year={2024}
}

@inproceedings{tobin2017domain,
  title={Domain randomization for transferring deep neural networks from simulation to the real world},
  author={Tobin, Josh and Fong, Rachel and Ray, Alex and Schneider, Jonas and Zaremba, Wojciech and Abbeel, Pieter},
  booktitle={2017 IEEE/RSJ international conference on intelligent robots and systems (IROS)},
  pages={23--30},
  year={2017},
  organization={IEEE}
}

@inproceedings{peng2018sim,
  title={Sim-to-real transfer of robotic control with dynamics randomization},
  author={Peng, Xue Bin and Andrychowicz, Marcin and Zaremba, Wojciech and Abbeel, Pieter},
  booktitle={2018 IEEE international conference on robotics and automation (ICRA)},
  pages={3803--3810},
  year={2018},
  organization={IEEE}
}

@inproceedings{nagabandi2018neural,
  title={Neural network dynamics for model-based deep reinforcement learning with model-free fine-tuning},
  author={Nagabandi, Anusha and Kahn, Gregory and Fearing, Ronald S and Levine, Sergey},
  booktitle={2018 IEEE international conference on robotics and automation (ICRA)},
  pages={7559--7566},
  year={2018},
  organization={IEEE}
}

@article{nagabandi2018learning,
  title={Learning to adapt in dynamic, real-world environments through meta-reinforcement learning},
  author={Nagabandi, Anusha and Clavera, Ignasi and Liu, Simin and Fearing, Ronald S and Abbeel, Pieter and Levine, Sergey and Finn, Chelsea},
  journal={arXiv preprint arXiv:1803.11347},
  year={2018}
}

@misc{finn2016unsupervised,
  title={Unsupervised Learning for Physical Interaction through Video Prediction},
  author={Chelsea Finn and Ian Goodfellow and Sergey Levine},
  year={2016},
  eprint={1605.07157},
  archivePrefix={arXiv},
  url={https://arxiv.org/abs/1605.07157}
}

@inproceedings{he2022mae,
  title={Masked Autoencoders Are Scalable Vision Learners},
  author={Kaiming He and Xinlei Chen and Saining Xie and Yanghao Li and Piotr Doll{\'a}r and Ross Girshick},
  booktitle={Proceedings of the IEEE/CVF Conference on Computer Vision and Pattern Recognition},
  pages={16000--16009},
  year={2022},
  url={https://arxiv.org/abs/2111.06377}
}

@inproceedings{tong2022videomae,
  title={VideoMAE: Masked Autoencoders are Data-Efficient Learners for Self-Supervised Video Pre-Training},
  author={Zhan Tong and Yibing Song and Jue Wang and Limin Wang},
  booktitle={Advances in Neural Information Processing Systems},
  volume={35},
  pages={10078--10093},
  year={2022},
  url={https://arxiv.org/abs/2203.12602}
}

@misc{kaplan2020scaling,
  title={Scaling Laws for Neural Language Models},
  author={Jared Kaplan and Sam McCandlish and Tom Henighan and Tom B. Brown and Benjamin Chess and Rewon Child and Scott Gray and Alec Radford and Jeffrey Wu and Dario Amodei},
  year={2020},
  eprint={2001.08361},
  archivePrefix={arXiv},
  url={https://arxiv.org/abs/2001.08361}
}

@misc{hoffmann2022training,
  title={Training Compute-Optimal Large Language Models},
  author={Jordan Hoffmann and Sebastian Borgeaud and Arthur Mensch and Elena Buchatskaya and Trevor Cai and Eliza Rutherford and Diego de Las Casas and Lisa Anne Hendricks and Johannes Welbl and Aidan Clark and others},
  year={2022},
  eprint={2203.15556},
  archivePrefix={arXiv},
  url={https://arxiv.org/abs/2203.15556}
}

@inproceedings{pearce2024scaling,
  title={Scaling Laws for Pre-training Agents and World Models},
  author={Tim Pearce and Tabish Rashid and David Bignell and Raluca Georgescu and Sam Devlin and Katja Hofmann},
  booktitle={Proceedings of the 42nd International Conference on Machine Learning},
  series={Proceedings of Machine Learning Research},
  volume={267},
  pages={48542--48562},
  year={2025},
  publisher={PMLR},
  url={https://proceedings.mlr.press/v267/pearce25a.html}
}

@misc{nvidia2022replicator,
  title={Build Custom Synthetic Data Generation Pipelines with Omniverse Replicator},
  author={{NVIDIA}},
  year={2022},
  howpublished={\url{https://developer.nvidia.com/blog/build-custom-synthetic-data-generation-pipelines-with-omniverse-replicator/}},
  note={Accessed: 2026-06-15}
}

@misc{nvidia2025grootdreams,
  title={Enhance Robot Learning with Synthetic Trajectory Data Generated by World Foundation Models},
  author={{NVIDIA}},
  year={2025},
  howpublished={\url{https://developer.nvidia.com/blog/enhance-robot-learning-with-synthetic-trajectory-data-generated-by-world-foundation-models/}},
  note={Accessed: 2026-06-15}
}

@inproceedings{wei2022chain,
  title={Chain-of-Thought Prompting Elicits Reasoning in Large Language Models},
  author={Jason Wei and Xuezhi Wang and Dale Schuurmans and Maarten Bosma and Brian Ichter and Fei Xia and Ed H. Chi and Quoc V. Le and Denny Zhou},
  booktitle={Advances in Neural Information Processing Systems},
  volume={35},
  pages={24824--24837},
  year={2022},
  url={https://proceedings.neurips.cc/paper_files/paper/2022/hash/9d5609613524ecf4f15af0f7b31abca4-Abstract-Conference.html}
}

@misc{hao2024coconut,
  title={Training Large Language Models to Reason in a Continuous Latent Space},
  author={Shibo Hao and Sainbayar Sukhbaatar and DiJia Su and Xian Li and Zhiting Hu and Jason Weston and Yuandong Tian},
  year={2024},
  eprint={2412.06769},
  archivePrefix={arXiv},
  url={https://arxiv.org/abs/2412.06769}
}

@misc{zhou2024minedreamer,
  title={MineDreamer: Learning to Follow Instructions via Chain-of-Imagination for Simulated-World Control},
  author={Enshen Zhou and Yiran Qin and Zhenfei Yin and Yuzhou Huang and Ruimao Zhang and Lu Sheng and Yu Qiao and Jing Shao},
  year={2024},
  eprint={2403.12037},
  archivePrefix={arXiv},
  url={https://arxiv.org/abs/2403.12037}
}

@misc{tan2025lcdrive,
  title={Latent Chain-of-Thought World Modeling for End-to-End Driving},
  author={Shuhan Tan and Kashyap Chitta and Yuxiao Chen and Ran Tian and Yurong You and Yan Wang and Wenjie Luo and Yulong Cao and Philipp Kr{\"a}henb{\"u}hl and Marco Pavone and Boris Ivanovic},
  year={2025},
  eprint={2512.10226},
  archivePrefix={arXiv},
  url={https://arxiv.org/abs/2512.10226}
}

@misc{lin2025futurex,
  title={FutureX: Enhance End-to-End Autonomous Driving via Latent Chain-of-Thought World Model},
  author={Hongbin Lin and Yiming Yang and Yifan Zhang and Chaoda Zheng and Jie Feng and Sheng Wang and Zhennan Wang and Shijia Chen and Boyang Wang and Yu Zhang and Xianming Liu and Shuguang Cui and Zhen Li},
  year={2025},
  eprint={2512.11226},
  archivePrefix={arXiv},
  url={https://arxiv.org/abs/2512.11226}
}

@book{pearl2009causality,
  title={Causality: Models, Reasoning, and Inference},
  author={Judea Pearl},
  edition={2},
  publisher={Cambridge University Press},
  year={2009}
}

@article{rubin1974estimating,
  title={Estimating Causal Effects of Treatments in Randomized and Nonrandomized Studies},
  author={Donald B. Rubin},
  journal={Journal of Educational Psychology},
  volume={66},
  number={5},
  pages={688--701},
  year={1974},
  doi={10.1037/h0037350}
}

@inproceedings{pawlowski2020deep,
  title={Deep Structural Causal Models for Tractable Counterfactual Inference},
  author={Nick Pawlowski and Daniel C. Castro and Ben Glocker},
  booktitle={Advances in Neural Information Processing Systems},
  volume={33},
  pages={857--869},
  year={2020},
  url={https://proceedings.neurips.cc/paper/2020/hash/0987b8b338d6c90bbedd8631bc499221-Abstract.html}
}

@article{scholkopf2021toward,
  title={Toward Causal Representation Learning},
  author={Bernhard Sch{\"o}lkopf and Francesco Locatello and Stefan Bauer and Nan Rosemary Ke and Nal Kalchbrenner and Anirudh Goyal and Yoshua Bengio},
  journal={Proceedings of the IEEE},
  volume={109},
  number={5},
  pages={612--634},
  year={2021},
  doi={10.1109/JPROC.2021.3058954}
}

@inproceedings{pitis2022mocoda,
  title={MoCoDA: Model-Based Counterfactual Data Augmentation},
  author={Silviu Pitis and Elliot Creager and Ajay Mandlekar and Animesh Garg},
  booktitle={Advances in Neural Information Processing Systems},
  volume={35},
  pages={18143--18156},
  year={2022},
  url={https://arxiv.org/abs/2210.11287}
}

@inproceedings{richens2024robust,
  title={Robust Agents Learn Causal World Models},
  author={Jonathan Richens and Tom Everitt},
  booktitle={The Twelfth International Conference on Learning Representations},
  year={2024},
  url={https://openreview.net/forum?id=pOoKI3ouv1}
}

@misc{gupta2024essential,
  title={The Essential Role of Causality in Foundation World Models for Embodied AI},
  author={Tarun Gupta and Wenbo Gong and Chao Ma and Nick Pawlowski and Agrin Hilmkil and Meyer Scetbon and Marc Rigter and Ade Famoti and Ashley Juan Llorens and Jianfeng Gao and Stefan Bauer and Danica Kragic and Bernhard Sch{\"o}lkopf and Cheng Zhang},
  year={2024},
  eprint={2402.06665},
  archivePrefix={arXiv},
  url={https://arxiv.org/abs/2402.06665}
}

@incollection{bareinboim2022pearl,
  title={On Pearl's Hierarchy and the Foundations of Causal Inference},
  author={Elias Bareinboim and Juan D. Correa and Duligur Ibeling and Thomas Icard},
  booktitle={Probabilistic and Causal Inference: The Works of Judea Pearl},
  editor={Hector Geffner and Rina Dechter and Joseph Y. Halpern},
  pages={507--556},
  publisher={ACM Books},
  year={2022},
  doi={10.1145/3501714.3501743}
}

@misc{nasr2023counterfactual,
  title={Counterfactual (Non-)identifiability of Learned Structural Causal Models},
  author={Arash Nasr-Esfahany and Emre K{\i}c{\i}man},
  year={2023},
  eprint={2301.09031},
  archivePrefix={arXiv},
  url={https://arxiv.org/abs/2301.09031}
}

@article{sutton1999between,
  title={Between MDPs and Semi-MDPs: A Framework for Temporal Abstraction in Reinforcement Learning},
  author={Richard S. Sutton and Doina Precup and Satinder Singh},
  journal={Artificial Intelligence},
  volume={112},
  number={1--2},
  pages={181--211},
  year={1999},
  doi={10.1016/S0004-3702(99)00052-1}
}

@article{dietterich2000maxq,
  title={Hierarchical Reinforcement Learning with the MAXQ Value Function Decomposition},
  author={Thomas G. Dietterich},
  journal={Journal of Artificial Intelligence Research},
  volume={13},
  pages={227--303},
  year={2000},
  doi={10.1613/jair.639}
}

@misc{hafner2022director,
  title={Deep Hierarchical Planning from Pixels},
  author={Danijar Hafner and Kuang-Huei Lee and Ian Fischer and Pieter Abbeel},
  year={2022},
  eprint={2206.04114},
  archivePrefix={arXiv},
  url={https://arxiv.org/abs/2206.04114}
}

@inproceedings{gumbsch2024learning,
  title={Learning Hierarchical World Models with Adaptive Temporal Abstractions from Discrete Latent Dynamics},
  author={Christian Gumbsch and Noor Sajid and Georg Martius and Martin V. Butz},
  booktitle={The Twelfth International Conference on Learning Representations},
  year={2024},
  url={https://openreview.net/forum?id=TjCDNssXKU}
}

@misc{wayne2018merlin,
  title={Unsupervised Predictive Memory in a Goal-Directed Agent},
  author={Greg Wayne and Chia-Chun Hung and David Amos and Mehdi Mirza and Arun Ahuja and Agnieszka Grabska-Barwinska and Jack Rae and Piotr Mirowski and Joel Z. Leibo and Adam Santoro and others},
  year={2018},
  eprint={1803.10760},
  archivePrefix={arXiv},
  url={https://arxiv.org/abs/1803.10760}
}

@inproceedings{pritzel2017neural,
  title={Neural Episodic Control},
  author={Alexander Pritzel and Benigno Uria and Sriram Srinivasan and Adri{\`a} Puigdom{\`e}nech Badia and Oriol Vinyals and Demis Hassabis and Daan Wierstra and Charles Blundell},
  booktitle={Proceedings of the 34th International Conference on Machine Learning},
  series={Proceedings of Machine Learning Research},
  volume={70},
  pages={2827--2836},
  year={2017},
  publisher={PMLR},
  url={https://proceedings.mlr.press/v70/pritzel17a.html}
}

@inproceedings{hu2021gem,
  title={Generalizable Episodic Memory for Deep Reinforcement Learning},
  author={Hao Hu and Jianing Ye and Guangxiang Zhu and Zhizhou Ren and Chongjie Zhang},
  booktitle={Proceedings of the 38th International Conference on Machine Learning},
  series={Proceedings of Machine Learning Research},
  volume={139},
  pages={4380--4390},
  year={2021},
  publisher={PMLR},
  url={https://proceedings.mlr.press/v139/hu21d.html}
}

@inproceedings{samsami2024mastering,
  title={Mastering Memory Tasks with World Models},
  author={Mohammad Reza Samsami and Artem Zholus and Janarthanan Rajendran and Sarath Chandar},
  booktitle={The Twelfth International Conference on Learning Representations},
  year={2024},
  url={https://arxiv.org/abs/2403.04253}
}

@inproceedings{talvitie2017self,
  title={Self-Correcting Models for Model-Based Reinforcement Learning},
  author={Erik Talvitie},
  booktitle={Proceedings of the Thirty-First AAAI Conference on Artificial Intelligence},
  year={2017},
  url={https://arxiv.org/abs/1612.06018}
}

@misc{jafferjee2020hallucinating,
  title={Mitigating Value Hallucination in Dyna Planning via Multistep Predecessor Models},
  author={Farzane Aminmansour and Taher Jafferjee and Ehsan Imani and Erin Talvitie and Micheal Bowling and Martha White},
  year={2020},
  eprint={2006.04363},
  archivePrefix={arXiv},
  primaryClass={cs.LG},
  doi={10.48550/arXiv.2006.04363},
  url={https://arxiv.org/abs/2006.04363}
}

@inproceedings{mildenhall2020nerf,
  title={NeRF: Representing Scenes as Neural Radiance Fields for View Synthesis},
  author={Mildenhall, Ben and Srinivasan, Pratul P. and Tancik, Matthew and Barron, Jonathan T. and Ramamoorthi, Ravi and Ng, Ren},
  booktitle={European Conference on Computer Vision},
  year={2020}
}

@article{kerbl20233dgaussians,
  title={3D Gaussian Splatting for Real-Time Radiance Field Rendering},
  author={Kerbl, Bernhard and Kopanas, Georgios and Leimk{\"u}hler, Thomas and Drettakis, George},
  journal={ACM Transactions on Graphics},
  volume={42},
  number={4},
  year={2023}
}

@inproceedings{zheng2024occworld,
  title={OccWorld: Learning a 3D Occupancy World Model for Autonomous Driving},
  author={Zheng, Wenzhao and Chen, Weiliang and Huang, Yuanhui and Zhang, Borui and Duan, Yueqi and Lu, Jiwen},
  booktitle={European Conference on Computer Vision},
  year={2024}
}

@article{zhang2025robooccworld,
  title={Occupancy World Model for Robots},
  author={Zhang, Zhang and Zhang, Qiang and Cui, Wei and Shi, Shuai and Guo, Yijie and Han, Gang and Zhao, Wen and Sun, Jingkai and Cao, Jiahang and Wang, Jiaxu and Cheng, Hao and Ju, Xiaozhu and Che, Zhengping and Xu, Renjing and Tang, Jian},
  journal={arXiv preprint arXiv:2505.05512},
  year={2025}
}

@article{yan2024renderworld,
  title={RenderWorld: World Model with Self-Supervised 3D Label},
  author={Yan, Ziyang and Dong, Wenzhen and Shao, Yihua and Lu, Yuhang and Liu, Haiyang and Liu, Jingwen and Wang, Haozhe and Wang, Zhe and Wang, Yan and Remondino, Fabio and Ma, Yuexin},
  journal={arXiv preprint arXiv:2409.11356},
  year={2024}
}

@inproceedings{locatello2020slotattention,
  title={Object-Centric Learning with Slot Attention},
  author={Locatello, Francesco and Weissenborn, Dirk and Unterthiner, Thomas and Mahendran, Aravindh and Heigold, Georg and Uszkoreit, Jakob and Dosovitskiy, Alexey and Kipf, Thomas},
  booktitle={Advances in Neural Information Processing Systems},
  year={2020}
}

@inproceedings{kipf2022savi,
  title={Conditional Object-Centric Learning from Video},
  author={Kipf, Thomas and Elsayed, Gamaleldin F. and Mahendran, Aravindh and Stone, Austin and Sabour, Sara and Heigold, Georg and Jampani, Varun and Greff, Klaus and Dosovitskiy, Alexey},
  booktitle={International Conference on Learning Representations},
  year={2022}
}

@inproceedings{veerapaneni2019op3,
  title={Entity Abstraction in Visual Model-Based Reinforcement Learning},
  author={Veerapaneni, Rishi and Co-Reyes, John D. and Chang, Michael and Janner, Michael and Finn, Chelsea and Wu, Jiajun and Tenenbaum, Joshua B. and Levine, Sergey},
  booktitle={Conference on Robot Learning},
  year={2019}
}

@article{wang2024occsora,
  title={OccSora: 4D Occupancy Generation Models as World Simulators for Autonomous Driving},
  author={Wang, Lening and Zheng, Wenzhao and Ren, Yilong and Jiang, Han and Cui, Zhiyong and Yu, Haiyang and Lu, Jiwen},
  journal={arXiv preprint arXiv:2405.20337},
  year={2024}
}

@article{yang2024driveoccworld,
  title={Driving in the Occupancy World: Vision-Centric 4D Occupancy Forecasting and Planning via World Models for Autonomous Driving},
  author={Yang, Yu and Mei, Jianbiao and Ma, Yukai and Du, Siliang and Chen, Wenqing and Qian, Yijie and Feng, Yuxiang and Liu, Yong},
  journal={arXiv preprint arXiv:2408.14197},
  year={2024}
}

@article{zhang2024bevworld,
  title={BEVWorld: A Multimodal World Simulator for Autonomous Driving via Scene-Level BEV Latents},
  author={Zhang, Yumeng and Gong, Shi and Xiong, Kaixin and Ye, Xiaoqing and Li, Xiaofan and Tan, Xiao and Wang, Fan and Huang, Jizhou and Wu, Hua and Wang, Haifeng},
  journal={arXiv preprint arXiv:2407.05679},
  year={2024},
  doi={10.48550/arXiv.2407.05679},
  url={https://arxiv.org/abs/2407.05679}
}

@article{zuo2025gaussianworld,
  title={GaussianWorld: Gaussian World Model for Streaming 3D Occupancy Prediction},
  author={Zuo, Sicheng and Zheng, Wenzhao and Huang, Yuanhui and Zhou, Jie and Lu, Jiwen},
  journal={arXiv preprint arXiv:2412.10373},
  year={2025}
}

@article{lu2025gwm,
  title={GWM: Towards Scalable Gaussian World Models for Robotic Manipulation},
  author={Lu, Guanxing and Jia, Baoxiong and Li, Puhao and Chen, Yixin and Wang, Ziwei and Tang, Yansong and Huang, Siyuan},
  journal={arXiv preprint arXiv:2508.17600},
  year={2025}
}

@inproceedings{wu2023slotformer,
  title={SlotFormer: Unsupervised Visual Dynamics Simulation with Object-Centric Models},
  author={Wu, Ziyi and Dvornik, Nikita and Greff, Klaus and Kipf, Thomas and Garg, Animesh},
  booktitle={International Conference on Learning Representations},
  year={2023}
}

@article{ferraro2023focus,
  title={FOCUS: Object-Centric World Models for Robotics Manipulation},
  author={Ferraro, Stefano and Mazzaglia, Pietro and Verbelen, Tim and Dhoedt, Bart},
  journal={arXiv preprint arXiv:2307.02427},
  year={2023}
}

@article{jeong2025objectcentricwm,
  title={Object-Centric World Model for Language-Guided Manipulation},
  author={Jeong, Youngjoon and Chun, Junha and Cha, Soonwoo and Kim, Taesup},
  journal={arXiv preprint arXiv:2503.06170},
  year={2025}
}

@article{huang2025enerverse,
  title={EnerVerse: Envisioning Embodied Future Space for Robotics Manipulation},
  author={Huang, Siyuan and Chen, Liliang and Zhou, Pengfei and Chen, Shengcong and Liao, Yue and Jiang, Zhengkai and Hu, Yue and Gao, Peng and Li, Hongsheng and Yao, Maoqing and Ren, Guanghui},
  journal={arXiv preprint arXiv:2501.01895},
  year={2025}
}

@article{jiang2025enerverseac,
  title={EnerVerse-AC: Envisioning Embodied Environments with Action Condition},
  author={Jiang, Yuxin and Chen, Shengcong and Huang, Siyuan and Chen, Liliang and Zhou, Pengfei and Liao, Yue and He, Xindong and Liu, Chiming and Li, Hongsheng and Yao, Maoqing and Ren, Guanghui},
  journal={arXiv preprint arXiv:2505.09723},
  year={2025}
}

@misc{worldlabs2025marble,
  title={Marble: A Multimodal World Model},
  author={{World Labs}},
  year={2025},
  howpublished={\url{https://www.worldlabs.ai/blog/marble-world-model}},
  note={Accessed: 2026-06-16}
}

@article{zou2026intern,
  title={Intern-s1-pro: Scientific multimodal foundation model at trillion scale},
  author={Zou, Yicheng and Zhu, Dongsheng and Zhu, Lin and Zhu, Tong and Zhou, Yunhua and Zhou, Peiheng and Zhou, Xinyu and Zhou, Dongzhan and Zhou, Zhiwang and Zhou, Yuhao and others},
  journal={arXiv preprint arXiv:2603.25040},
  year={2026}
}

@article{lam2023learning,
  title={Learning skillful medium-range global weather forecasting},
  author={Lam, Remi and Sanchez-Gonzalez, Alvaro and Willson, Matthew and Wirnsberger, Peter and Fortunato, Meire and Alet, Ferran and Ravuri, Suman and Ewalds, Timo and Eaton-Rosen, Zach and Hu, Weihua and others},
  journal={Science},
  volume={382},
  number={6677},
  pages={1416--1421},
  year={2023},
  publisher={American Association for the Advancement of Science}
}

@article{lam2023graphcast,
  title   = {Learning skillful medium-range global weather forecasting},
  author  = {Lam, Remi and Sanchez-Gonzalez, Alvaro and Willson, Matthew and Wirnsberger, Peter and Fortunato, Meire and Alet, Ferran and Ravuri, Suman and Ewalds, Timo and Eaton-Rosen, Zach and Hu, Weihua and others},
  journal = {Science},
  volume  = {382},
  number  = {6677},
  pages   = {1416--1421},
  year    = {2023},
  doi     = {10.1126/science.adi2336},
  url     = {https://www.science.org/doi/10.1126/science.adi2336}
}

@article{price2025probabilistic,
  title={Probabilistic weather forecasting with machine learning},
  author={Price, Ilan and Sanchez-Gonzalez, Alvaro and Alet, Ferran and Andersson, Tom R and El-Kadi, Andrew and Masters, Dominic and Ewalds, Timo and Stott, Jacklynn and Mohamed, Shakir and Battaglia, Peter and others},
  journal={Nature},
  volume={637},
  number={8044},
  pages={84--90},
  year={2025},
  publisher={Nature Publishing Group UK London}
}

@article{chen2023fengwu,
  title={Fengwu: Pushing the skillful global medium-range weather forecast beyond 10 days lead},
  author={Chen, Kang and Han, Tao and Gong, Junchao and Bai, Lei and Ling, Fenghua and Luo, Jing-Jia and Chen, Xi and Ma, Leiming and Zhang, Tianning and Su, Rui and others},
  journal={arXiv preprint arXiv:2304.02948},
  year={2023}
}

@article{ling2024fengwu,
  title={FengWu-W2S: A deep learning model for seamless weather-to-subseasonal forecast of global atmosphere},
  author={Ling, Fenghua and Chen, Kang and Wu, Jiye and Han, Tao and Luo, Jing-Jia and Ouyang, Wanli and Bai, Lei},
  journal={arXiv preprint arXiv:2411.10191},
  year={2024}
}

@article{yang2025medical,
  title={Medical world model: Generative simulation of tumor evolution for treatment planning},
  author={Yang, Yijun and Wang, Zhao-Yang and Liu, Qiuping and Sun, Shuwen and Wang, Kang and Chellappa, Rama and Zhou, Zongwei and Yuille, Alan and Zhu, Lei and Zhang, Yu-Dong and others},
  journal={arXiv preprint arXiv:2506.02327},
  year={2025}
}

@article{shmatko2025learning,
  title={Learning the natural history of human disease with generative transformers},
  author={Shmatko, Artem and Jung, Alexander Wolfgang and Gaurav, Kumar and Brunak, S{\o}ren and Mortensen, Laust Hvas and Birney, Ewan and Fitzgerald, Tom and Gerstung, Moritz},
  journal={Nature},
  volume={647},
  number={8088},
  pages={248--256},
  year={2025},
  publisher={Nature Publishing Group UK London}
}

@article{szymanski2023autonomous,
  title={An autonomous laboratory for the accelerated synthesis of inorganic materials},
  author={Szymanski, Nathan J and Rendy, Bernardus and Fei, Yuxing and Kumar, Rishi E and He, Tanjin and Milsted, David and McDermott, Matthew J and Gallant, Max and Cubuk, Ekin Dogus and Merchant, Amil and others},
  journal={Nature},
  volume={624},
  number={7990},
  pages={86},
  year={2023}
}

@article{wei2025ai,
  title={From ai for science to agentic science: A survey on autonomous scientific discovery},
  author={Wei, Jiaqi and Yang, Yuejin and Zhang, Xiang and Chen, Yuhan and Zhuang, Xiang and Gao, Zhangyang and Zhou, Dongzhan and Wang, Guangshuai and Gao, Zhiqiang and Cao, Juntai and others},
  journal={arXiv preprint arXiv:2508.14111},
  year={2025}
}

@article{wu2025ai,
  title={An AI-native experimental laboratory for autonomous biomolecular engineering},
  author={Wu, Mingyu and Wang, Zhaoguo and Wang, Jiabin and Dong, Zhiyuan and Yang, Jingkai and Li, Qingting and Huang, Tianyu and Zhao, Lei and Li, Mingqiang and Wang, Fei and others},
  journal={arXiv preprint arXiv:2507.02379},
  year={2025}
}

@article{bai2025intern,
  title={Intern-s1: A scientific multimodal foundation model},
  author={Bai, Lei and Cai, Zhongrui and Cao, Yuhang and Cao, Maosong and Cao, Weihan and Chen, Chiyu and Chen, Haojiong and Chen, Kai and Chen, Pengcheng and Chen, Ying and others},
  journal={arXiv preprint arXiv:2508.15763},
  year={2025}
}

@InProceedings{huang2023vbench,
     title={VBench: Comprehensive Benchmark Suite for Video Generative Models},
     author={Huang, Ziqi and He, Yinan and Yu, Jiashuo and Zhang, Fan and Si, Chenyang and Jiang, Yuming and Zhang, Yuanhan and Wu, Tianxing and Jin, Qingyang and Chanpaisit, Nattapol and Wang, Yaohui and Chen, Xinyuan and Wang, Limin and Lin, Dahua and Qiao, Yu and Liu, Ziwei},
     booktitle={Proceedings of the IEEE/CVF Conference on Computer Vision and Pattern Recognition},
     year={2024}
}

@misc{li2026worldmodelbench,
  title={WorldModelBench: Judging Video Generation Models As World Models},
  author={Li, Dacheng and Fang, Yunhao and Chen, Yukang and Yang, Shuo and Cao, Shiyi and Wong, Justin and Luo, Michael and Wang, Xiaolong and Yin, Hongxu and Gonzalez, Joseph E. and Stoica, Ion and Han, Song and Lu, Yao},
  year={2025},
  eprint={2502.20694},
  archivePrefix={arXiv},
  primaryClass={cs.CV},
  doi={10.48550/arXiv.2502.20694},
  url={https://arxiv.org/abs/2502.20694}
}

@article{bellemare2013arcade,
  title={The arcade learning environment: An evaluation platform for general agents},
  author={Bellemare, Marc G and Naddaf, Yavar and Veness, Joel and Bowling, Michael},
  journal={Journal of artificial intelligence research},
  volume={47},
  pages={253--279},
  year={2013}
}

@article{tassa2018deepmind,
  title={Deepmind control suite},
  author={Tassa, Yuval and Doron, Yotam and Muldal, Alistair and Erez, Tom and Li, Yazhe and Casas, Diego de Las and Budden, David and Abdolmaleki, Abbas and Merel, Josh and Lefrancq, Andrew and others},
  journal={arXiv preprint arXiv:1801.00690},
  year={2018}
}

@article{atreya2025roboarena,
  title={Roboarena: Distributed real-world evaluation of generalist robot policies},
  author={Atreya, Pranav and Pertsch, Karl and Lee, Tony and Kim, Moo Jin and Jain, Arhan and Kuramshin, Artur and Eppner, Clemens and Neary, Cyrus and Hu, Edward and Ramos, Fabio and others},
  journal={arXiv preprint arXiv:2506.18123},
  year={2025}
}

@book{craik1943nature,
  title     = {The Nature of Explanation},
  author    = {Craik, Kenneth J. W.},
  year      = {1943},
  publisher = {Cambridge University Press}
}

@misc{ha2018worldmodelsorig,
  title         = {World Models},
  author        = {David Ha and J{\"u}rgen Schmidhuber},
  year          = {2018},
  eprint        = {1803.10122},
  archivePrefix = {arXiv},
  primaryClass  = {cs.LG},
  url           = {https://arxiv.org/abs/1803.10122}
}

@misc{ding2024understanding,
  title         = {Understanding World or Predicting Future? A Comprehensive Survey of World Models},
  author        = {Jingtao Ding and Yunke Zhang and Yu Shang and Yuheng Zhang and Zefang Zong and Jie Feng and Yuan Yuan and Hongyuan Su and Nian Li and Nicholas Sukiennik and Fengli Xu and Yong Li},
  year          = {2024},
  eprint        = {2411.14499},
  archivePrefix = {arXiv},
  url           = {https://arxiv.org/abs/2411.14499}
}

@misc{dawid2023autonomous,
  title         = {Introduction to Latent Variable Energy-Based Models: A Path Towards Autonomous Machine Intelligence},
  author        = {Anna Dawid and Yann LeCun},
  year          = {2023},
  eprint        = {2306.02572},
  archivePrefix = {arXiv},
  primaryClass  = {cs.LG},
  url           = {https://arxiv.org/abs/2306.02572}
}

@article{lecun2022path,
  title={A path towards autonomous machine intelligence version 0.9. 2, 2022-06-27},
  author={LeCun, Yann},
  journal={Open Review},
  volume={62},
  number={1},
  pages={1--62},
  year={2022}
}

@misc{xing2025critiques,
  title         = {Critique of World Model},
  author        = {Eric Xing and Mingkai Deng and Jinyu Hou},
  year          = {2025},
  eprint        = {2507.05169},
  archivePrefix = {arXiv},
  primaryClass  = {cs.LG},
  doi           = {10.48550/arXiv.2507.05169},
  url           = {https://arxiv.org/abs/2507.05169}
}

@misc{nair2019hvf,
  title        = {Hierarchical Foresight: Self-Supervised Learning of Long-Horizon Tasks via Visual Subgoal Generation},
  author       = {Suraj Nair and Chelsea Finn},
  year         = {2019},
  eprint       = {1909.05829},
  archivePrefix = {arXiv},
  primaryClass = {cs.RO},
  url          = {https://arxiv.org/abs/1909.05829}
}

@misc{huang2025selfforcing,
  title        = {Self Forcing: Bridging the Train-Test Gap in Autoregressive Video Diffusion},
  author       = {Xun Huang and Zhengqi Li and Guande He and Mingyuan Zhou and Eli Shechtman},
  year         = {2025},
  eprint       = {2506.08009},
  archivePrefix = {arXiv},
  primaryClass = {cs.CV},
  url          = {https://arxiv.org/abs/2506.08009}
}

@article{raissi2019physics,
  title   = {Physics-informed neural networks: A deep learning framework for solving forward and inverse problems involving nonlinear partial differential equations},
  author  = {Raissi, Maziar and Perdikaris, Paris and Karniadakis, George Em},
  journal = {Journal of Computational Physics},
  volume  = {378},
  pages   = {686--707},
  year    = {2019},
  doi     = {10.1016/j.jcp.2018.10.045},
  url     = {https://doi.org/10.1016/j.jcp.2018.10.045}
}

@inproceedings{greydanus2019hamiltonian,
  title     = {Hamiltonian Neural Networks},
  author    = {Greydanus, Samuel and Dzamba, Misko and Yosinski, Jason},
  booktitle = {Advances in Neural Information Processing Systems},
  volume    = {32},
  year      = {2019},
  url       = {https://papers.nips.cc/paper_files/paper/2019/hash/26cd8ecadce0d4efd6cc8a8725cbd1f8-Abstract.html}
}

@misc{cranmer2020lagrangian,
  title         = {Lagrangian Neural Networks},
  author        = {Cranmer, Miles and Greydanus, Sam and Hoyer, Stephan and Battaglia, Peter and Spergel, David and Ho, Shirley},
  year          = {2020},
  eprint        = {2003.04630},
  archivePrefix = {arXiv},
  primaryClass  = {cs.LG},
  url           = {https://arxiv.org/abs/2003.04630}
}

@inproceedings{zhong2020symplectic,
  title     = {Symplectic ODE-Net: Learning Hamiltonian Dynamics with Control},
  author    = {Zhong, Yaofeng Desmond and Dey, Biswadip and Chakraborty, Amit},
  booktitle = {International Conference on Learning Representations},
  year      = {2020},
  url       = {https://openreview.net/forum?id=ryxmb1rKDS}
}

@inproceedings{sanchez2020learning,
  title     = {Learning to Simulate Complex Physics with Graph Networks},
  author    = {Sanchez-Gonzalez, Alvaro and Godwin, Jonathan and Pfaff, Tobias and Ying, Rex and Leskovec, Jure and Battaglia, Peter},
  booktitle = {Proceedings of the 37th International Conference on Machine Learning},
  series    = {Proceedings of Machine Learning Research},
  volume    = {119},
  pages     = {8459--8468},
  year      = {2020},
  publisher = {PMLR},
  url       = {https://proceedings.mlr.press/v119/sanchez-gonzalez20a.html}
}

@inproceedings{pfaff2021learning,
  title     = {Learning Mesh-Based Simulation with Graph Networks},
  author    = {Pfaff, Tobias and Fortunato, Meire and Sanchez-Gonzalez, Alvaro and Battaglia, Peter W.},
  booktitle = {International Conference on Learning Representations},
  year      = {2021},
  url       = {https://arxiv.org/abs/2010.03409}
}

@inproceedings{de2018end,
  title     = {End-to-End Differentiable Physics for Learning and Control},
  author    = {de Avila Belbute-Peres, Filipe and Smith, Kevin and Allen, Kelsey and Tenenbaum, Josh and Kolter, J. Zico},
  booktitle = {Advances in Neural Information Processing Systems},
  volume    = {31},
  year      = {2018},
  url       = {https://papers.nips.cc/paper_files/paper/2018/hash/842424a1d0595b76ec4fa03c46e8d755-Abstract.html}
}

@inproceedings{hu2020difftaichi,
  title     = {DiffTaichi: Differentiable Programming for Physical Simulation},
  author    = {Hu, Yuanming and Anderson, Luke and Li, Tzu-Mao and Sun, Qi and Carr, Nathan and Ragan-Kelley, Jonathan and Durand, Fr{\'e}do},
  booktitle = {International Conference on Learning Representations},
  year      = {2020},
  url       = {https://openreview.net/forum?id=B1eB5xSFvr}
}

@misc{freeman2021brax,
  title         = {Brax -- A Differentiable Physics Engine for Large Scale Rigid Body Simulation},
  author        = {Freeman, C. Daniel and Frey, Erik and Raichuk, Anton and Girgin, Sertan and Mordatch, Igor and Bachem, Olivier},
  year          = {2021},
  eprint        = {2106.13281},
  archivePrefix = {arXiv},
  primaryClass  = {cs.RO},
  doi           = {10.48550/arXiv.2106.13281},
  url           = {https://arxiv.org/abs/2106.13281}
}

@article{alradi2026aeroworld,
  title = {Aero-World: Action-Conditioned Aerial Video Generation from Inertial Controls},
  author = {Al Radi, Abdul Mohaimen and Li, Kunyang and Shang, Yuzhang and Shah, Mubarak and Tian, Yu},
  journal = {arXiv preprint arXiv:2605.19728},
  year = {2026},
  url = {https://arxiv.org/abs/2605.19728}
}

@article{huang2025forgetree,
  title = {{FORGE}-Tree: Diffusion-Forcing Tree Search for Long-Horizon Robot Manipulation},
  author = {Huang, Yanjia and Liu, Shuo and Liu, Sheng and Xu, Qingxiao and Wu, Mingyang and Gao, Xiangbo and Tu, Zhengzhong},
  journal = {arXiv preprint arXiv:2510.21744},
  year = {2025},
  url = {https://arxiv.org/abs/2510.21744}
}

@article{chi2023diffusionpolicy,
  title = {Diffusion Policy: Visuomotor Policy Learning via Action Diffusion},
  author = {Chi, Cheng and Xu, Zhenjia and Feng, Siyuan and Cousineau, Eric and Du, Yilun and Burchfiel, Benjamin and Tedrake, Russ and Song, Shuran},
  journal = {arXiv preprint arXiv:2303.04137},
  year = {2023},
  url = {https://arxiv.org/abs/2303.04137}
}

@article{yu2020pixelnerf,
  title = {{pixelNeRF}: Neural Radiance Fields from One or Few Images},
  author = {Yu, Alex and Ye, Vickie and Tancik, Matthew and Kanazawa, Angjoo},
  journal = {arXiv preprint arXiv:2012.02190},
  year = {2020},
  url = {https://arxiv.org/abs/2012.02190}
}

@article{liu2023zero123,
  title = {Zero-1-to-3: Zero-shot One Image to 3D Object},
  author = {Liu, Ruoshi and Wu, Rundi and Van Hoorick, Basile and Tokmakov, Pavel and Zakharov, Sergey and Vondrick, Carl},
  journal = {arXiv preprint arXiv:2303.11328},
  year = {2023},
  url = {https://arxiv.org/abs/2303.11328}
}

@article{kang2026geonvs,
  title = {{GeoNVS}: Geometry Grounded Video Diffusion for Novel View Synthesis},
  author = {Kang, Minjun and Shin, Inkyu and Lee, Taeyeop and Kim, Myungchul and Kweon, In So and Yoon, Kuk-Jin},
  journal = {arXiv preprint arXiv:2603.14965},
  year = {2026},
  url = {https://arxiv.org/abs/2603.14965}
}

@misc{hunyuanworld2025,
  title = {{HunyuanWorld 1.0}: Generating Immersive, Explorable, and Interactive 3D Worlds from Words or Pixels},
  author = {{HunyuanWorld Team} and Wang, Zhenwei and Liu, Yuhao and others},
  year = {2025},
  eprint = {2507.21809},
  archivePrefix = {arXiv},
  primaryClass = {cs.CV},
  doi = {10.48550/arXiv.2507.21809},
  url = {https://arxiv.org/abs/2507.21809}
}

@misc{qwen2025omni,
  title = {{Qwen2.5-Omni} Technical Report},
  author = {Xu, Jin and Guo, Zhifang and He, Jinzheng and Hu, Hangrui and He, Ting and Bai, Shuai and Chen, Keqin and Wang, Jialin and Fan, Yang and Dang, Kai and Zhang, Bin and Wang, Xiong and Chu, Yunfei and Lin, Junyang},
  year = {2025},
  eprint = {2503.20215},
  archivePrefix = {arXiv},
  primaryClass = {cs.CL},
  doi = {10.48550/arXiv.2503.20215},
  url = {https://arxiv.org/abs/2503.20215}
}

@inproceedings{zitkovich2023rt,
  title={Rt-2: Vision-language-action models transfer web knowledge to robotic control},
  author={Zitkovich, Brianna and Yu, Tianhe and Xu, Sichun and Xu, Peng and Xiao, Ted and Xia, Fei and Wu, Jialin and Wohlhart, Paul and Welker, Stefan and Wahid, Ayzaan and others},
  booktitle={Conference on Robot Learning},
  pages={2165--2183},
  year={2023},
  organization={PMLR}
}

@misc{shang2026survey,
  title        = {A Survey of Embodied World Models},
  author       = {Shang, Yu and Tang, Yinzhou and Zhang, Xin and Wang, Shengyuan and Yan, Yuwei and Zhang, Honglin and Zheng, Zhiheng and Zhao, Jie and Feng, Jie and Gao, Chen and others},
  year         = {2026},
  doi          = {10.20944/preprints202604.0928.v2},
  url          = {https://www.preprints.org/manuscript/202604.0928/v2},
  note         = {Preprints}
}

@misc{jing2024mdgen,
  title         = {Generative Modeling of Molecular Dynamics Trajectories},
  author        = {Jing, Bowen and St{\"a}rk, Hannes and Jaakkola, Tommi and Berger, Bonnie},
  year          = {2024},
  eprint        = {2409.17808},
  archivePrefix = {arXiv},
  primaryClass  = {cs.LG},
  url           = {https://arxiv.org/abs/2409.17808}
}

@article{kaelbling1998planning,
  title         = {Planning and acting in partially observable stochastic domains},
  author        = {Kaelbling, Leslie Pack and Littman, Michael L. and Cassandra, Anthony R.},
  journal       = {Artificial Intelligence},
  volume        = {101},
  number        = {1--2},
  pages         = {99--134},
  year          = {1998},
  publisher     = {Elsevier},
  url           = {https://people.csail.mit.edu/lpk/papers/aij98-pomdp.pdf}
}

@misc{sun2026vlajepa,
  title         = {VLA-JEPA: Enhancing Vision-Language-Action Model with Latent World Model},
  author        = {Sun, Jingwen and Zhang, Wenyao and Qi, Zekun and Ren, Shaojie and Liu, Zezhi and Zhu, Hanxin and Sun, Guangzhong and Jin, Xin and Chen, Zhibo},
  year          = {2026},
  eprint        = {2602.10098},
  archivePrefix = {arXiv},
  primaryClass  = {cs.RO},
  url           = {https://arxiv.org/abs/2602.10098}
}

@misc{zhou2025socialworldmodels,
  title         = {Social World Models},
  author        = {Zhou, Xuhui and Liu, Jiarui and Yerukola, Akhila and Kim, Hyunwoo and Sap, Maarten},
  year          = {2025},
  eprint        = {2509.00559},
  archivePrefix = {arXiv},
  primaryClass  = {cs.AI},
  url           = {https://arxiv.org/abs/2509.00559}
}

@inproceedings{perdomo2020performative,
  title         = {Performative Prediction},
  author        = {Perdomo, Juan C. and Zrnic, Tijana and Mendler-D{\"u}nner, Celestine and Hardt, Moritz},
  booktitle     = {Proceedings of the 37th International Conference on Machine Learning},
  pages         = {7599--7609},
  year          = {2020},
  series        = {Proceedings of Machine Learning Research},
  volume        = {119},
  publisher     = {PMLR},
  url           = {https://proceedings.mlr.press/v119/perdomo20a.html}
}

@inproceedings{xie2024physgaussian,
  author    = {Xie, Tianyi and Zong, Zeshun and Qiu, Yuxing and Li, Xuan and Feng, Yutao and Yang, Yin and Jiang, Chenfanfu},
  title     = {{PhysGaussian}: Physics-Integrated 3{D} {Gaussians} for Generative Dynamics},
  booktitle = {Proceedings of the IEEE/CVF Conference on Computer Vision and Pattern Recognition (CVPR)},
  year      = {2024}
}

@article{qin2024worldsimbench,
  title={Worldsimbench: Towards video generation models as world simulators},
  author={Qin, Yiran and Shi, Zhelun and Yu, Jiwen and Wang, Xijun and Zhou, Enshen and Li, Lijun and Yin, Zhenfei and Liu, Xihui and Sheng, Lu and Shao, Jing and others},
  journal={arXiv preprint arXiv:2410.18072},
  year={2024}
}

@article{yue2025ewmbench,
  title={Ewmbench: Evaluating scene, motion, and semantic quality in embodied world models},
  author={Yue, Hu and Huang, Siyuan and Liao, Yue and Chen, Shengcong and Zhou, Pengfei and Chen, Liliang and Yao, Maoqing and Ren, Guanghui},
  journal={arXiv preprint arXiv:2505.09694},
  year={2025}
}

@article{xu2026worldmark,
  title={WorldMark: A Unified Benchmark Suite for Interactive Video World Models},
  author={Xu, Xiaojie and Lin, Zhengyuan and He, Kang and Feng, Yukang and Mao, Xiaofeng and Yin, Yuanyang and Zhang, Kaipeng and Ge, Yongtao},
  journal={arXiv preprint arXiv:2604.21686},
  year={2026}
}

@article{ying2026wbench,
  title={WBench: A Comprehensive Multi-turn Benchmark for Interactive Video World Model Evaluation},
  author={Ying, Kaining and Hu, Hengrui and Ren, Siyu and Li, Jiamu and Chen, Fengjiao and Wang, Ziwen and Cao, Xuezhi and Cai, Xunliang and Ding, Henghui},
  journal={arXiv preprint arXiv:2605.25874},
  year={2026}
}

@article{zhang2026mbench,
  title={MBench: A Comprehensive Benchmark on Memory Capability for Video World Models},
  author={Zhang, Shengjun and Zhang, Zhang and Huang, Simin and Tang, Zhenyu and Wang, Hanyang and Dai, Chensheng and Chen, Min and Li, Yifan and Li, Yuxin and Chen, Yingjie and others},
  journal={arXiv preprint arXiv:2606.00793},
  year={2026}
}

@article{chen2025worldprediction,
  title={WorldPrediction: A Benchmark for High-level World Modeling and Long-horizon Procedural Planning},
  author={Chen, Delong and Chung, Willy and Bang, Yejin and Ji, Ziwei and Fung, Pascale},
  journal={arXiv preprint arXiv:2506.04363},
  year={2025}
}

@article{wu2026worldreasonbench,
  title={WorldReasonBench: Human-Aligned Stress Testing of Video Generators as Future World-State Predictors},
  author={Wu, Keming and Cui, Yijing and Xue, Wenhan and Wang, Qijie and Luo, Xuan and Feng, Zhiyuan and Yang, Zuhao and Wang, Sudong and Jiang, Sicong and Zhu, Haowei and others},
  journal={arXiv preprint arXiv:2605.10434},
  year={2026}
}

@inproceedings{duan2025worldscore,
  title={Worldscore: A unified evaluation benchmark for world generation},
  author={Duan, Haoyi and Yu, Hong-Xing and Chen, Sirui and Fei-Fei, Li and Wu, Jiajun},
  booktitle={Proceedings of the IEEE/CVF International Conference on Computer Vision},
  pages={27713--27724},
  year={2025}
}

@inproceedings{lu20264dworldbench,
  title={4dworldbench: A comprehensive evaluation framework for 3d/4d world generation models},
  author={Lu, Yiting and Luo, Wei and Tu, Peiyan and Li, Haoran and Zhu, Hanxin and Yu, Zihao and Wang, Xingrui and Chen, Xinyi and Peng, Xinge and Li, Xin and others},
  booktitle={Proceedings of the IEEE/CVF Conference on Computer Vision and Pattern Recognition},
  pages={34322--34332},
  year={2026}
}
\end{document}